\documentclass[11pt]{article}

\usepackage[margin=1in]{geometry}

\usepackage{microtype}
\usepackage{graphicx}
\usepackage{subcaption}
\usepackage{booktabs} % for professional tables
\usepackage{float}
\usepackage{algorithm}
\usepackage{algorithmic}

\usepackage{hyperref}
\usepackage[round,authoryear]{natbib}

\usepackage{amsmath}
\usepackage{amssymb}
\usepackage{mathtools}
\usepackage{amsthm}
\usepackage{bm}

\usepackage[capitalize,noabbrev]{cleveref}

\theoremstyle{plain}

\theoremstyle{definition}

\theoremstyle{remark}

\usepackage[textsize=tiny]{todonotes}

\title{GEM: Guided Expectation-Maximization for Behavior-Normalized Candidate Action Selection in Offline RL}
\author{
Haoyu Wang\textsuperscript{1},
Jingcheng Wang\textsuperscript{1}\thanks{Corresponding author: \texttt{jcwang@sjtu.edu.cn}},
Shunyu Wu\textsuperscript{1},
Xinwei Xiao\textsuperscript{1} \\[0.5em]
\textsuperscript{1}School of Automation and Intelligent Sensing, Shanghai Jiao Tong University, Shanghai, China\\[0.25em]
\textsuperscript{1}\texttt{\{create\_arc, jcwang, shunyuwu, xxw971205\}@sjtu.edu.cn}
}
\date{}

\begin{document}
\maketitle

\begin{abstract}
Offline reinforcement learning (RL) can fit strong value functions from fixed datasets, yet reliable deployment still hinges on the action-selection interface used to query them. When the dataset induces a branched or multimodal action landscape, unimodal policy extraction can blur competing hypotheses and yield “in-between” actions that are weakly supported by data, making decisions brittle even with a strong critic. We introduce GEM (Guided Expectation–Maximization), an analytical framework that makes action selection both multimodal and explicitly controllable. GEM trains a Gaussian Mixture Model (GMM) actor via critic-guided, advantage-weighted EM-style updates that preserve distinct components while shifting probability mass toward high-value regions, and learns a tractable GMM behavior model to quantify support. During inference, GEM performs candidate-based selection: it generates a parallel candidate set and reranks actions using a conservative ensemble lower-confidence bound together with behavior-normalized support, where the behavior log-likelihood is standardized within each state’s candidate set to yield stable, comparable control across states and candidate budgets. Empirically, GEM is competitive across D4RL benchmarks, and offers a simple inference-time budget knob (candidate count) that trades compute for decision quality without retraining.
\end{abstract}

\begin{figure}[t]
  \centering
  \includegraphics[width=\linewidth]{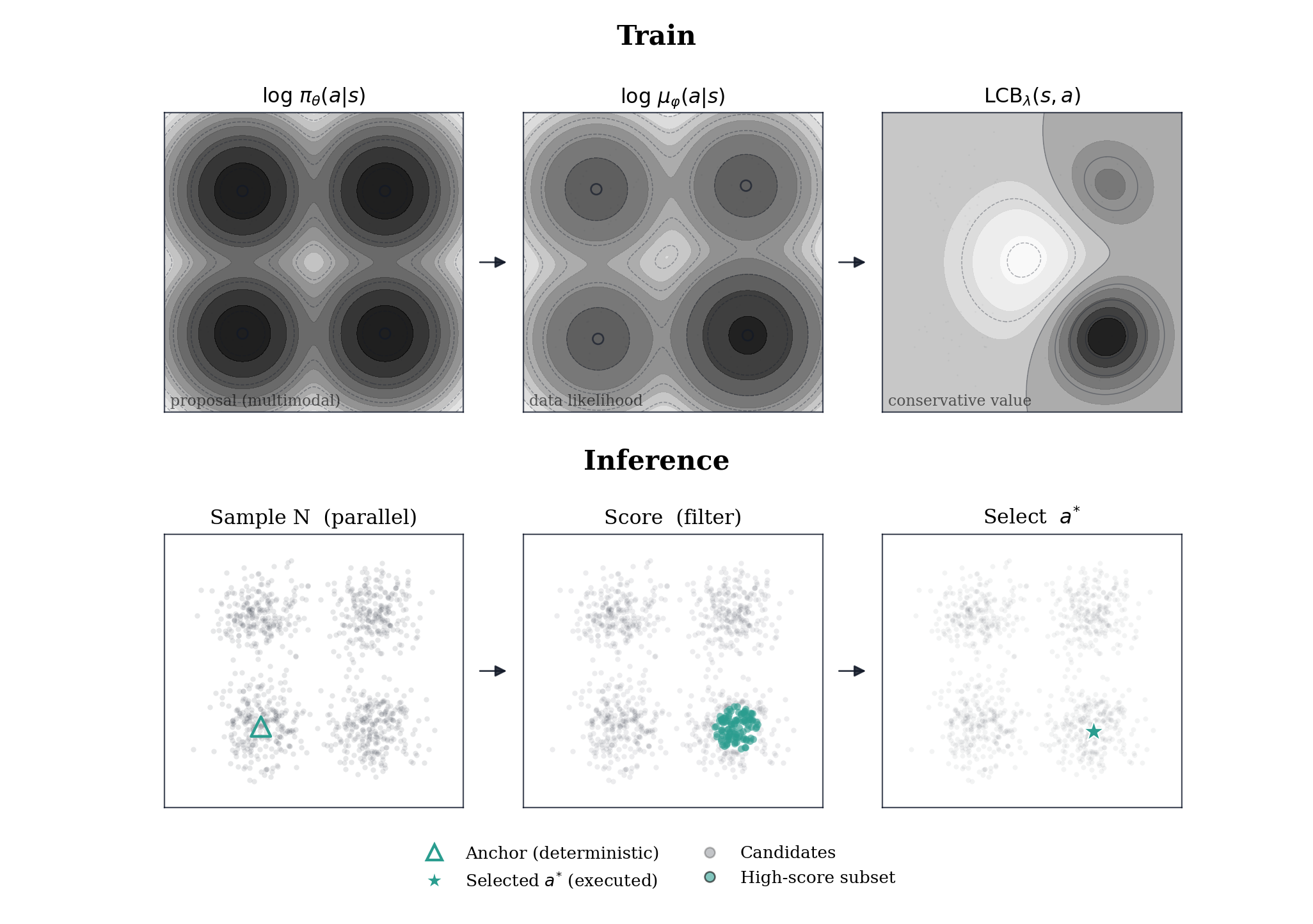}
  \caption{\textbf{GEM mechanism (schematic at a fixed state in a 2-D action space).}
  \textbf{Training (top row):} GEM learns three ingredients that will be combined only at test time:
  (i) a multimodal actor $\pi_\theta(a\mid s)$ (left; shown as $\log\pi_\theta(a\mid s)$) that preserves multiple plausible action hypotheses,
  (ii) an independent behavior density $\mu_\varphi(a\mid s)$ (middle; shown as $\log\mu_\varphi(a\mid s)$) used to quantify dataset support,
  and (iii) a conservative value statistic $\mathrm{LCB}_\lambda(s,a)$ from a critic ensemble (right).
  \textbf{Inference (bottom row):} given a queried state $s$, GEM samples $N$ candidates in parallel and adds a deterministic anchor (triangle),
  scores each candidate using Eq.~\ref{eq:score} (conservative value plus behavior-normalized support), and executes the top-ranked action $a^\star$ (star).
  The green cluster highlights the small subset of candidates that remain competitive after scoring, illustrating how the interface filters many proposals
  down to a support-aligned, high-value choice under maximization.}
  \label{fig:gem_mechanism}
\end{figure}

\section{Introduction}

Offline reinforcement learning (RL) learns from a fixed dataset without further environment interaction. A central difficulty is distributional shift: once deployed decisions drift away from the dataset support, value estimates can become unreliable precisely on actions that the data do not substantiate. This failure mode has been extensively analyzed through extrapolation and approximation error, motivating a family of offline RL algorithms that constrain action selection or regularize value learning to avoid out-of-distribution (OOD) queries \citep{levine2020offline,10078377,fujimoto2019offpolicydeepreinforcementlearning,kumar2019stabilizingoffpolicyqlearningbootstrapping,wu2019behaviorregularizedofflinereinforcement,kumar2020conservativeqlearningofflinereinforcement,kidambi2021morelmodelbasedoffline,luo2025learningtrustbellmanupdates}. As a result, modern offline RL can often fit strong critics under fixed data.

Deployment, however, is not only a question of how a critic is trained---it is also a question of how that critic is queried. At test time, an agent must map a state to an executable action, and this action-selection interface determines whether decisions remain near the dataset support or are pulled toward OOD regions. The most common interface in continuous control is single-shot execution of a parametric actor (often with behavioral regularization or conservative values) \citep{fujimoto2021minimalistapproachofflinereinforcement}. This interface is simple and efficient, but it hard-codes a single decision rule: it provides no principled way to trade additional test-time computation for improved decisions, and it offers no structured mechanism to represent multiple competing action hypotheses.

Candidate-based action selection provides a practical alternative interface. Given a state, one generates a set of plausible actions and selects the best one under a scoring rule. The candidate budget becomes a clean deployment knob: it can be increased when compute is available and decreased under latency constraints, without retraining. Candidate selection is not new in offline RL---support-aware methods already generate multiple candidates and choose using a critic \citep{zhou2020plaslatentactionspace}---but promoting deployment to an $\arg\max$ over candidates changes the failure mode in a way that is easy to understate and costly to ignore offline.

Naive candidate maximization can systematically amplify OOD risk. Increasing the candidate budget strengthens extreme-value effects: if the critic exhibits local overestimation or uncertainty, the maximum over a larger set is more likely to be attained by a spuriously high-valued action that is weakly supported by the dataset. In other words, the very mechanism that makes candidates attractive as a scalable interface (searching over more hypotheses) can increase off-support querying unless the selection rule explicitly controls both support and uncertainty. This motivates the central question of this paper: how can offline RL provide a candidate-based deployment interface where increasing test-time compute improves decisions rather than exacerbating OOD errors?

A second, orthogonal challenge is representational mismatch in continuous actions. Many actor-extraction objectives can be interpreted as advantage-weighted regression toward a target action distribution \citep{peng2019advantageweightedregressionsimplescalable,nair2021awacacceleratingonlinereinforcement,kostrikov2021offlinereinforcementlearningimplicit}. When the conditional action landscape is branched or multimodal, restricting the actor to a unimodal family can blur separated hypotheses and yield `in-between'' actions that are weakly supported by data. This mismatch is not universal, but it becomes consequential in a candidate-based interface: proposal quality determines which hypotheses are even available to be compared, while selection quality determines whether expanding the hypothesis set remains safe.

We propose \textbf{Guided Expectation--Maximization (GEM)}, an analytical action-selection framework built around Gaussian mixture models (GMMs) that makes candidate-based deployment controllable. GEM couples (i) a critic-guided, advantage-weighted EM-style training view that preserves distinct mixture components while shifting probability mass toward high-value regions, with (ii) an inference rule that reranks a per-state candidate set using a conservative ensemble lower-confidence bound together with an explicit behavior-support signal. The key design is interface-centric: the support term is behavior-normalized within each state's candidate set, so a single support weight has a stable meaning across states and candidate budgets. This perspective complements policy-constraint and conservative-value approaches \citep{an2021uncertaintybasedofflinereinforcementlearning} and is structurally different from expressive generative policies (e.g., sequence models and diffusion models) that improve expressivity but may trade away cheap deployment-time likelihood control and incur iterative inference costs \citep{chen2021decisiontransformerreinforcementlearning,wang2023diffusionpoliciesexpressivepolicy,janner2022planningdiffusionflexiblebehavior,hansenestruch2023idqlimplicitqlearningactorcritic}. We select GMMs not as a claim about expressivity ceilings, but as a claim about interfaces: GMMs provide parallelizable candidate proposal, a tractable behavior likelihood for explicit support control, and interpretable components that enable mechanism-level auditing of proposal and selection.

Our contributions are:
We formulate offline RL deployment as an \emph{action-selection interface} problem: candidate-based selection is a practical compute knob, but naive maximization can amplify OOD errors as candidate budgets scale.
We introduce GEM, which trains a multimodal GMM actor via critic-guided, advantage-weighted EM-style updates that preserve mixture components while shifting probability mass toward high-value regions.
We design an inference-time scoring rule that jointly controls uncertainty and support by combining a conservative ensemble lower-confidence bound with an explicit behavior-support term that is normalized within each state's candidate set for stable deployment control.
We provide a pure test-time scaling knob via candidate search, enabling compute--quality tradeoffs without retraining while keeping both policy and behavior likelihood analytically tractable.
\paragraph{Roadmap.}
Section~2 presents preliminaries and motivates candidate-based deployment interfaces in offline RL.
Section~3 describes GEM, including GMM actor/behavior modeling and inference-time candidate reranking.
Section~4 evaluates GEM with ablations that isolate proposal design, support control, and conservatism.
Section~5 discusses limitations and implications for deployment-time interfaces.

\section{Preliminaries and Motivation}
\label{sec:prelim_motivation}

\subsection{Offline RL deployment as a query interface}

We consider a continuous-action Markov decision process with states $s\in\mathcal{S}$ and actions $a\in\mathcal{A}$.
Offline RL learns from a fixed dataset $\mathcal{D}=\{(s,a,r,s')\}$ collected by an unknown behavior policy with conditional density $\mu(a\mid s)$.
Deployment is operational: for a given state $s$, which action $a$ should be executed?

A central offline failure mode is distributional shift at deployment, which can be viewed as a mismatch between what is evaluated and what is supported.
When the deployed decision rule selects actions that are weakly supported under $\mu(\cdot\mid s)$, value estimation is forced to extrapolate, and the resulting policy can be brittle even if the critic is accurate on-support.
This motivates a \emph{query-interface} view: action selection determines which $(s,a)$ pairs are actually queried at test time, and therefore directly controls OOD exposure.

\subsection{Candidate-based selection as a pure deployment compute knob}
Candidate-based selection replaces single-shot execution with a propose--select interface.
For each state $s$, one constructs a candidate set
\begin{equation}
\label{eq:candset}
\mathcal{C}(s)=\{a_0,a_1,\ldots,a_N\},
\end{equation}
where $a_0$ is an anchor action (deterministic proposal) and $a_{1:N}$ are sampled candidates.
A scoring rule $\mathrm{Score}(s,a)$ ranks candidates, and the executed action is typically
\[
a^\star \in \arg\max_{a\in\mathcal{C}(s)} \mathrm{Score}(s,a).
\]
This interface introduces a clean deployment knob: the candidate budget $N$ scales test-time computation without retraining.

\subsection{Extreme-value amplification as a deployment constraint}
Candidate maximization is not benign in offline RL.
Critics can exhibit local overestimation and uncertainty, particularly off-support.
As $N$ increases, maximization strengthens selection pressure toward outliers: with more hypotheses, it becomes more likely that at least one candidate attains a spuriously high estimated score, and the selector will systematically choose it.
The interface-level implication is a hard constraint: \emph{scaling compute by increasing $N$ must not systematically increase OOD querying}.

\subsection{Two complementary controls: explicit support and conservative value}
A scalable selection interface requires two complementary controls: an explicit measure of behavioral support, and a conservative statistic that counters uncertainty-driven outliers.

\paragraph{Independent support model.}
We represent support using a separate behavior model $\mu_\varphi(a\mid s)$ fit by behavior cloning to approximate the dataset density.
Using $\mu_\varphi$ rather than the learned actor avoids self-certification: the actor cannot be used to certify its own support without collapsing the meaning of ``on-support'' into what it already proposes.

\paragraph{Calibration via per-state candidate-set normalization.}
Raw $\log\mu_\varphi(a\mid s)$ varies in scale across states and datasets, so an additive term $w_p \log\mu_\varphi(a\mid s)$ is not a stable knob.
We instead standardize support within each state's candidate set:
\begin{equation}
\label{eq:zscore}
\begin{aligned}
\mathrm{zscore}_s(x(a))
&=
\frac{x(a)-\bar x_s}{\max(\sigma_s,\epsilon)},\\
\bar x_s
&=
\frac{1}{|\mathcal{C}(s)|}\sum_{a'\in\mathcal{C}(s)} x(a'),\\
\sigma_s
&=
\sqrt{\frac{1}{|\mathcal{C}(s)|}\sum_{a'\in\mathcal{C}(s)}
\big(x(a')-\bar x_s\big)^2},\\
x(a)
&=
\log\mu_\varphi(a\mid s).
\end{aligned}
\end{equation}
so that $w_p$ admits a consistent interpretation: it trades conservative value against \emph{relative support measured in candidate-set standard deviations}.

\paragraph{Conservative value via ensembles.}
Support control alone does not eliminate all failures because uncertainty can persist even on-support under function approximation.
Given an ensemble $\{Q_i(s,a)\}_{i=1}^M$, define
\begin{equation}
\label{eq:lcb}
\mathrm{LCB}_\lambda(s,a)=\bar Q(s,a)-\lambda\,\mathrm{Std}\big(Q_i(s,a)\big),
\end{equation}
which targets overestimation and uncertainty-driven outliers.
This term complements support: $\mu_\varphi$ controls dataset density, while $\mathrm{LCB}_\lambda$ controls estimator risk.
\begin{equation}
\label{eq:score}
\mathrm{Score}(s,a)
=
\mathrm{LCB}_\lambda(s,a)
+
w_p\cdot \mathrm{zscore}_s\!\big(\log\mu_\varphi(a\mid s)\big).
\end{equation}

\subsection{Analytical multimodality and an audit via NLL gap}
Candidate selection can only compare proposed hypotheses.
When the conditional action landscape is branched or multimodal, restricting the proposal family to unimodal forms can blur separated hypotheses into weakly supported ``in-between'' actions.
This motivates analytical multimodality in the proposal family, together with an explicit diagnostic that audits whether mixture structure yields tangible benefit.

\paragraph{Multimodality diagnostic via NLL gap.}
We compute an NLL-gap diagnostic on dataset actions under a fitted $K$-component GMM
$\pi_K(a\mid s)=\sum_{k=1}^K w_k(s)\,\mathcal{N}\!\big(a;\mu_k(s),\Sigma_k(s)\big)$.
Define
\[
\mathrm{NLL}_{\mathrm{gmm}}
=
\mathbb{E}_{(s,a)\sim\mathcal{D}}
\Big[
-\log \sum_{k=1}^K w_k(s)\,\mathcal{N}\!\big(a;\mu_k(s),\Sigma_k(s)\big)
\Big],
\]
and define the ``top-1'' proxy by selecting $k^\star(s)=\arg\max_k w_k(s)$ and evaluating the corresponding Gaussian NLL:
\[
\mathrm{NLL}_{\mathrm{top1}}
=
\mathbb{E}_{(s,a)\sim\mathcal{D}}
\Big[
-\log \mathcal{N}\!\big(a;\mu_{k^\star(s)}(s),\Sigma_{k^\star(s)}(s)\big)
\Big].
\]
We report this gap in Appendix~\ref{app:nll_gap} as a dataset-level audit of conditional action multimodality.

The discussion above yields three requirements for a scalable offline deployment interface:
(i) a proposal mechanism that can represent multiple action hypotheses and generate candidates efficiently;
(ii) an explicit, independent support signal that can be calibrated so a single weight has stable meaning across states and candidate budgets;
(iii) a conservative selection rule that controls uncertainty and overestimation risk under function approximation.
GEM is designed to satisfy all three simultaneously.

\section{Method}
\label{sec:method}

\subsection{Overview}
GEM is an action-selection interface for querying offline critics under a fixed dataset.
Phase~I learns (i) an ensemble critic used through Eq.~\ref{eq:lcb}, (ii) a $K$-component GMM actor trained by critic-\emph{guided} EM-style updates, and (iii) an independent GMM behavior density used only for support.
Phase~II is the \emph{inference-time} action-selection interface: given a queried state $s$, it constructs an anchor+$N$ candidate set (Eq.~\ref{eq:candset}) and returns the executed action by reranking candidates with the fixed-form conservative-support score in Eq.~\ref{eq:score}.

\subsection{Phase I: three learned models}
\label{sec:phase1}

\paragraph{Critic ensemble.}
We learn an ensemble $\{Q_i\}_{i=1}^M$ and a value baseline $V$ with standard offline regression.
At deployment the ensemble is used only via $\mathrm{LCB}_\lambda$ in Eq.~\ref{eq:lcb}.

\paragraph{Actor GMM and guided EM-style update.}
The actor $\pi_\theta(a\mid s)$ is a diagonal-covariance $K$-component GMM with gating weights $w_k(s)$ and component parameters $(\mu_k(s),\Sigma_k(s))$.
For a dataset pair $(s,a)$, define the component log-joint
\[
u_k(s,a)=\log w_k(s)+\log \mathcal{N}\!\big(a;\mu_k(s),\Sigma_k(s)\big),
\]
\[
\qquad
\gamma_k(s,a)=\mathrm{softmax}_k\!\big(u_k(s,a)\big).
\]
We optimize the loose EM-style surrogate
\[
\mathrm{ELBO}_{\mathrm{loose}}(s,a)=\sum_{k=1}^K \gamma_k(s,a)\,u_k(s,a),
\]
optionally treating $\gamma$ as fixed when taking gradients.
Guidance is implemented as a per-sample weight $w(s,a)\propto \exp(\beta A(s,a))$ (clamped for stability), where $A(s,a)$ is constructed from the learned critic and value baseline; guidance affects only this weighting, while responsibilities remain defined purely by $\{u_k\}$.
We also include a lightweight entropy bonus on the gating distribution to discourage premature collapse.

\paragraph{Behavior GMM $\mu_\varphi(a\mid s)$.}
We fit an independent $K$-component GMM behavior density $\mu_\varphi(a\mid s)$ by maximum likelihood on dataset actions.
It is used only to evaluate $\log\mu_\varphi(a\mid s)$ for support scoring; the support term never uses the actor density.

\subsection{Phase II: candidate construction and selection}
\label{sec:phase2}

\paragraph{Candidates (anchor + single source).}
At inference time, for each queried state $s$, we form the candidate set $\mathcal{C}(s)$ in Eq.~\ref{eq:candset}.
The anchor is deterministic: $k^\star(s)=\arg\max_k w_k(s)$ and $a_0=\mu_{k^\star(s)}(s)$.
The remaining $N$ candidates are sampled from \emph{exactly one} source (actor \emph{or} behavior), with no mixing and no fallback.

\paragraph{Inference-time scoring rule.}
At inference, for each $a\in\mathcal{C}(s)$ We score candidates using Eq.~\ref{eq:score} and execute the top-ranked action.

The inference interface outputs the executed action as the top-scoring candidate under Eq.~\ref{eq:score}.
Optionally, as a variance-reduction output operator, we return the mean of the top-$k_{\mathrm{smooth}}$ actions ranked by Eq.~\ref{eq:score}$,$ (default $k_{\mathrm{smooth}}=1$).

\subsection{Mechanism statements (interface-level semantics)}
\label{sec:mechanism}

\paragraph{T1 (Unimodal projection $\Rightarrow$ ``in-between'' actions).}
\textbf{Statement.} If the target conditional action distribution is multimodal with separated modes, projecting onto a unimodal Gaussian family can produce a central tendency lying between modes, yielding weakly supported actions.
\textbf{Sketch.} Reverse-KL / I-projection views align moment structure dominated by the mean; for separated mixtures the mean can fall in a low-density region, motivating multimodal proposals and the NLL-gap audit in Section~\ref{sec:prelim_motivation}.

\paragraph{T2 (``Guided'' EM-style update shifts mass toward high-advantage data).}
\textbf{Statement.} The actor update is EM-style via soft responsibilities and a responsibility-weighted surrogate, and \emph{guided} because dataset actions are reweighted by a bounded exponential of a critic-derived advantage.
\textbf{Sketch.} Responsibilities are determined by mixture log-joints; advantage weighting concentrates fitting pressure on higher-value data, while a gating-entropy bonus discourages premature collapse, retaining multi-hypothesis proposals.

\paragraph{T3 (Candidate-set z-scoring stabilizes $w_p$).}
\textbf{Statement.} Standardizing $\log\mu_\varphi(a\mid s)$ within $\mathcal{C}(s)$ (Eq.~\ref{eq:zscore}) makes $w_p$ comparable across states and budgets as an exchange rate between conservative value and relative support.
\textbf{Sketch.} Z-scoring is invariant to affine rescaling within a state, stabilizing scale and interpretation under heterogeneous likelihood magnitudes and changing $|\mathcal{C}(s)|$.

\paragraph{T4 (LCB counters budget-driven outliers).}
\textbf{Statement.} Increasing the candidate budget strengthens extreme-value selection under estimator noise; pessimism via Eq.~\ref{eq:lcb} reduces the chance that larger $N$ systematically selects high-uncertainty outliers.
\textbf{Sketch.} Maximization over more candidates increases the probability of spuriously high estimates; LCB penalizes ensemble disagreement, complementing behavior-normalized support in Eq.~\ref{eq:score}.
\begin{algorithm}[t]
\caption{\textsc{GEM} Inference Interface (Phase II, primary)}
\label{alg:gem_phase2_primary}
\footnotesize
\begin{algorithmic}[1]
\REQUIRE Learned critics $\{Q_i\}_{i=1}^M$; behavior density $\mu_\varphi$; actor GMM gating $w_k(s)$ and component means $\mu_k(s)$.
\REQUIRE Inference hyperparams: budget $N$, pessimism $\lambda$, support weight $w_p$, smoothing $k_{\text{smooth}}$; candidate source \texttt{src}$\in\{\pi,\mu\}$.

\STATE \textbf{Input:} queried state $s$.
\STATE \textbf{Anchor (mode-based deterministic proposal).}
\STATE $k^\star \leftarrow \arg\max_k w_k(s)$; \ $a_0 \leftarrow \mu_{k^\star}(s)$.
\STATE \textbf{Candidate set (anchor + single source, no mixing).}
\STATE Sample $a_{1:N}\sim \texttt{src}(\cdot\mid s)$; \ $\mathcal C(s)\leftarrow\{a_0,a_1,\dots,a_N\}$.

\STATE \textbf{Conservative value term (budget-robust).}
\FORALL{$a\in\mathcal C(s)$}
  \STATE Compute $\mathrm{LCB}_\lambda(s,a)$ using critic ensemble (Eq.~\ref{eq:lcb}).
\ENDFOR

\STATE \textbf{Support term (relative, candidate-normalized).}
\FORALL{$a\in\mathcal C(s)$}
  \STATE $\ell(a)\leftarrow \log\mu_\varphi(a\mid s)$.
\ENDFOR
\STATE $\tilde\ell(a)\leftarrow \mathrm{zscore}_s(\ell(a))$ (Eq.~\ref{eq:zscore}).

\STATE \textbf{Fixed-form conservative-support score \& selection.}
\STATE $\mathrm{Score}(s,a)\leftarrow \mathrm{LCB}_\lambda(s,a)+w_p\,\tilde\ell(a)$ (Eq.~\ref{eq:score}).
\STATE Let $k \leftarrow \min(k_{\text{smooth}},|\mathcal C(s)|)$.
\STATE Return mean of top-$k$ actions ranked by $\mathrm{Score}(s,\cdot)$ (default $k=1$).
\end{algorithmic}
\end{algorithm}
\noindent\textbf{Training pseudocode.} For completeness, we provide Phase~I training pseudocode in Appendix~A (Algorithm~\ref{alg:gem_phase1_supporting}).

% =========================
\section{Experiments: Auditing Test-time Inference Interfaces}
\label{sec:results}
% =========================

\graphicspath{{main_figs/}{sub_figs/}{group_figs/}{figures/}}

\subsection{Evaluation protocol and reading guide}
\label{sec:results_protocol}

We evaluate on D4RL continuous-control benchmarks and report normalized score (higher is better).
Unless stated otherwise, all numbers are the mean over three random seeds after $10^6$ gradient updates with no early stopping.
At test time, GEM uses the propose--select interface from Section~3: an anchor plus $N$ sampled candidates are scored by Eq.~\ref{eq:score} and the top-ranked action is executed.
Crucially, $N$, $\lambda$, and the support-weight schedule $w_p$ are test-time inference knobs: we sweep them without retraining, which is the concrete meaning of the abstract claim that GEM trades compute for decision quality without retraining.

We additionally report two evaluator-side audits (not training losses).
\textbf{SupportViolationRate} (\textsc{Violation}) is the fraction of decision steps where the selected action is low-support under the behavior model after candidate-set standardization, $z(\log\mu_\varphi(a^\star\!\mid s))<-2.0$.
\textbf{CollapseDist} is the per-step Euclidean distance from the selected action (top-1 under Eq.~\ref{eq:score}, before smoothing) to the nearest component mean of the behavior GMM.
Suite-aggregated plots in the main paper omit error bars because cross-environment score scales differ substantially and can visually swamp the scaling trends; environment-level breakdown plots are provided in Appendix~\ref{app:env_breakdowns}.

Reading guide: Section~\ref{sec:results_main} anchors standard benchmark performance and specifies the test-time inference sweep used to select default knobs per suite.
Section~\ref{sec:results_scaling} motivates $N{=}1024$ via candidate-budget scaling and saturation.
Section~\ref{sec:results_tradeoff} measures the compute shape of this knob and contrasts it with sequential-step inference knobs.
Sections~\ref{sec:results_stress}--\ref{sec:results_support_knob} then fix the defaults and audit mechanism claims (support/pessimism complementarity and calibrated support normalization); the corresponding figures are collected in Appendix~\ref{app:additional}.

% -------------------------------------------------
\subsection{Benchmark anchoring with full training and inference-time knob selection}
\label{sec:results_main}

\begin{table*}[t]
\centering
\caption{Unified D4RL benchmark comparison (normalized score, higher is better).
Suites: Locomotion (HalfCheetah/Hopper/Walker2d), AntMaze, and Maze2D.}
\label{tab:unified_d4rl_all}
% ============================================================
% Unified D4RL table body (verbatim numbers).
% NOTE:
% - This file is meant to be \input{} inside an outer table/table* float.
% - No \begin{table} ... \end{table} in this file.
% - Uses only \hline (no booktabs dependency).
% - If you keep \resizebox, ensure \usepackage{graphicx} in the preamble.
% ============================================================

\resizebox{\textwidth}{!}{%
\begin{tabular}{lccccccccccccc}
\hline
Task & TD3+BC & AWAC & CQL & IQL & ReBRAC & SAC-N & EDAC & DT & IDQL & Diffusion-QL & FQL & SSAR & GEM(OURS) \\
\hline

\multicolumn{14}{l}{\textbf{Locomotion}} \\
\hline
HC-M    & 48.10 & 50.02 & 47.04 & 48.31 & 64.04 & 68.20 & 67.70 & 42.20 & 51.10 & 51.10 & \texttt{nan} & 60.00 & 54.42 \\
HC-MR   & 44.84 & 45.13 & 45.04 & 44.46 & 51.18 & 60.70 & 62.06 & 38.91 & 45.90 & 47.80 & \texttt{nan} & 51.70 & 49.60 \\
HC-ME   & 90.78 & 95.00 & 95.63 & 94.74 & 103.80 & 98.96 & 104.76 & 91.55 & 95.90 & 96.80 & \texttt{nan} & 98.50 & 100.95 \\
H-M     & 60.30 & 63.02 & 59.08 & 67.53 & 102.29 & 40.82 & 101.70 & 65.10 & 65.40 & 90.50 & \texttt{nan} & 95.40 & 100.08 \\
H-MR    & 64.42 & 98.88 & 95.11 & 97.43 & 94.98 & 100.33 & 99.66 & 81.77 & 92.10 & 101.30 & \texttt{nan} & 101.50 & 101.23 \\
H-ME    & 101.17 & 101.90 & 99.26 & 107.42 & 109.45 & 101.31 & 105.19 & 110.44 & 108.60 & 111.10 & \texttt{nan} & 106.70 & 110.33 \\
W-M     & 82.71 & 68.52 & 80.75 & 80.91 & 85.82 & 87.47 & 93.36 & 67.63 & 82.50 & 87.00 & \texttt{nan} & 86.40 & 83.51 \\
W-MR    & 85.62 & 80.62 & 73.09 & 82.15 & 84.25 & 78.99 & 87.10 & 59.80 & 85.10 & 95.50 & \texttt{nan} & 94.10 & 97.02 \\
W-ME    & 110.03 & 111.44 & 109.56 & 111.72 & 111.86 & 114.93 & 114.75 & 107.11 & 112.70 & 110.10 & \texttt{nan} & 112.40 & 109.74 \\
Sum     & 687.97 & 714.53 & 704.56 & 734.67 & 807.67 & 751.71 & 836.28 & 664.51 & 739.30 & 791.20 & \texttt{nan} & 806.70 & 806.88 \\
Avg     & 76.45 & 79.39 & 78.28 & 81.63 & 89.74 & 83.52 & 92.92 & 73.84 & 82.14 & 87.91 & \texttt{nan} & 89.63 & 89.65 \\
\hline

\multicolumn{14}{l}{\textbf{AntMaze}} \\
\hline
AM-U     & 70.75 & 56.75 & 92.75 & 77.00 & 97.75 & 0.00 & 0.00 & 57.00 & 94.00 & 93.40 & 96.00 & 94.70 & 83.00 \\
AM-U-D   & 44.75 & 54.75 & 37.25 & 54.25 & 83.50 & 0.00 & 0.00 & 51.75 & 80.20 & 66.20 & 89.00 & 65.10 & 79.33 \\
AM-M-P   & 0.25  & 0.00  & 65.75 & 65.75 & 89.50 & 0.00 & 0.00 & 0.00  & 84.50 & 76.60 & 78.00 & 59.80 & 78.33 \\
AM-M-D   & 0.25  & 0.00  & 67.25 & 73.75 & 83.50 & 0.00 & 0.00 & 0.00  & 84.80 & 78.60 & 71.00 & 59.60 & 82.67 \\
AM-L-P   & 0.00  & 0.00  & 20.75 & 42.00 & 52.25 & 0.00 & 0.00 & 0.00  & 63.50 & 46.40 & 84.00 & 35.50 & 37.33 \\
AM-L-D   & 0.00  & 0.00  & 20.50 & 30.25 & 64.00 & 0.00 & 0.00 & 0.00  & 67.90 & 56.60 & 83.00 & 26.70 & 37.00 \\
Sum      & 116.00 & 111.50 & 304.25 & 343.00 & 470.50 & 0.00 & 0.00 & 108.75 & 474.90 & 417.80 & 501.00 & 341.40 & 397.66 \\
Avg      & 19.33 & 18.58 & 50.71 & 57.17 & 78.42 & 0.00 & 0.00 & 18.12 & 79.15 & 69.63 & 83.50 & 56.90 & 66.28 \\
\hline

\multicolumn{14}{l}{\textbf{Maze2D}} \\
\hline
MZ-U   & 29.41 & 65.65 & -8.90 & 42.11 & 106.87 & 130.59 & 95.26 & 18.08 & \texttt{nan} & \texttt{nan} & \texttt{nan} & \texttt{nan} & 59.03 \\
MZ-M   & 59.45 & 84.63 & 86.11 & 34.85 & 105.11 & 88.61 & 57.04 & 31.71 & \texttt{nan} & \texttt{nan} & \texttt{nan} & \texttt{nan} & 117.61 \\
MZ-L   & 97.10 & 215.50 & 23.75 & 61.72 & 78.33 & 204.76 & 95.60 & 35.66 & \texttt{nan} & \texttt{nan} & \texttt{nan} & \texttt{nan} & 90.16 \\
Sum    & 185.96 & 365.78 & 100.96 & 138.68 & 290.31 & 423.96 & 247.90 & 85.45 & \texttt{nan} & \texttt{nan} & \texttt{nan} & \texttt{nan} & 266.80 \\
Avg    & 61.99 & 121.92 & 33.65 & 46.23 & 96.77 & 141.32 & 82.64 & 28.48 & \texttt{nan} & \texttt{nan} & \texttt{nan} & \texttt{nan} & 88.93 \\
\hline
\end{tabular}%
}

\end{table*}

Table~\ref{tab:unified_d4rl_all} anchors end-task credibility under standard reporting while keeping the proposal and support models analytically tractable (GMM actor and GMM behavior density).
All entries in this table correspond to full training runs ($10^6$ gradient updates, no early stopping, three-seed mean); after training, we select deployment knobs by sweeping inference-time parameters on a held-out validation protocol, then report the corresponding test performance.
This separation matters operationally: the score function in Eq.~\ref{eq:score} contains inference-only controls ($N$, $\lambda$, and $w_p$), so performance can often be improved by deployment-time tuning without retraining, aligning with the abstract claim.

\paragraph{Inference-time sweep used for Table~\ref{tab:unified_d4rl_all}.}
During benchmark evaluation, we sweep the pessimism coefficient $\lambda$ and the terminal support weight $w_p^{\mathrm{end}}$ while keeping trained networks fixed.
We use a cosine schedule for the support weight,
$w_p(t)=w_p^{\mathrm{end}}+\tfrac{1}{2}(1-w_p^{\mathrm{end}})\bigl(1+\cos(\pi t/T)\bigr)$,
so $w_p$ starts at $1$ and anneals to $w_p^{\mathrm{end}}$ over an episode horizon of length $T$.
The best settings selected by this sweep are:
for Locomotion and Maze2D, $\lambda{=}1$ and $w_p^{\mathrm{end}}{=}0.4$;
for AntMaze, $\lambda{=}3.5$ and $w_p^{\mathrm{end}}{=}0.2$.
These suite-specific defaults are then held fixed for all ablations in Sections~\ref{sec:results_scaling}--\ref{sec:results_support_knob}, where each ablation changes exactly one factor at a time.

\paragraph{What the test-time inference interface adds (anchored to IQL).}
GEM uses an IQL-style training backbone, so IQL in Table~\ref{tab:unified_d4rl_all} is the closest reference point for isolating what the test-time inference interface buys.
On suite averages, GEM improves over IQL by $+8.02$ on Locomotion (89.65 vs 81.63), $+9.11$ on AntMaze (66.28 vs 57.17), and $+42.70$ on Maze2D (88.93 vs 46.23).
The largest per-task gains concentrate where candidate selection can exploit multimodal hypotheses instead of collapsing to ``in-between'' actions, e.g., Hopper-medium (H-M) $+32.55$ (100.08 vs 67.53) and Maze2D-medium (MZ-M) $+82.76$ (117.61 vs 34.85).
The gains are not uniform: Walker2d-medium-expert (W-ME) is slightly below IQL ($-1.98$), and AntMaze-large-play (AM-L-P) is below IQL ($-4.67$), consistent with GEM targeting test-time selection risk rather than long-horizon stitching or credit assignment.

Relative to representative generative / sequence-style policies reported in the same table, GEM is competitive on Locomotion (89.65 vs Diffusion-QL 87.91, $+1.74$; vs Decision Transformer 73.84, $+15.81$) and trails Diffusion-QL on AntMaze (66.28 vs 69.63, $-3.35$) while remaining far above Decision Transformer (18.12, $+48.16$).
After anchoring standard return, the remainder of this section tests the paper’s central interface claim: GEM exposes retrain-free test-time inference knobs (candidate budget $N$ and the scoring weights in Eq.~\ref{eq:score}) and keeps large-$N$ maximization auditable and controllable via explicit support and pessimism instead of drifting toward maximizer outliers.

\subsection{Candidate-budget scaling}
\label{sec:results_scaling}

\begin{figure}[t]
  \centering
  \includegraphics[width=\linewidth]{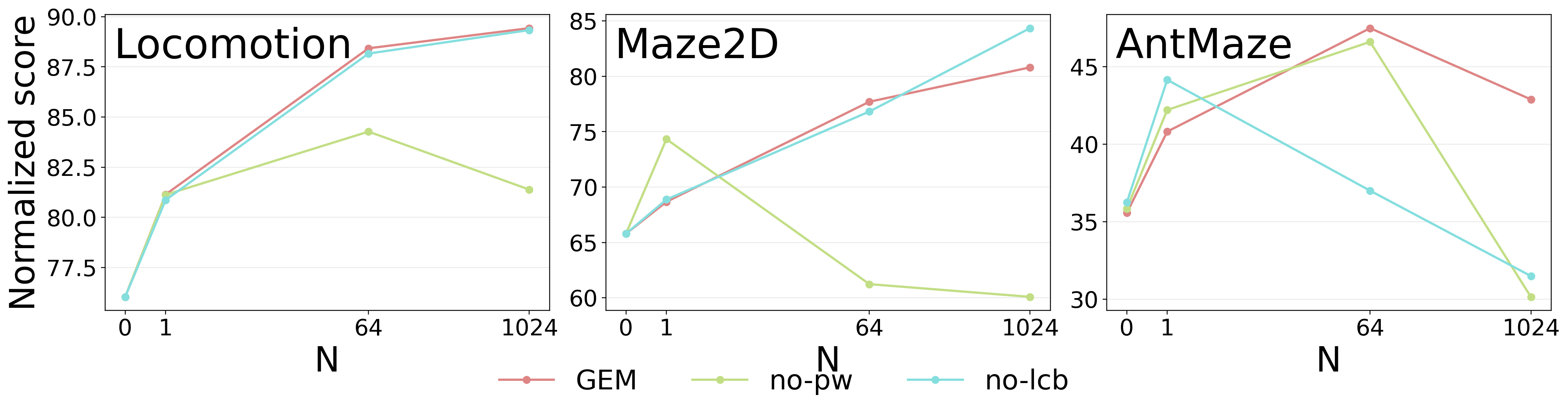}\\
  \includegraphics[width=\linewidth]{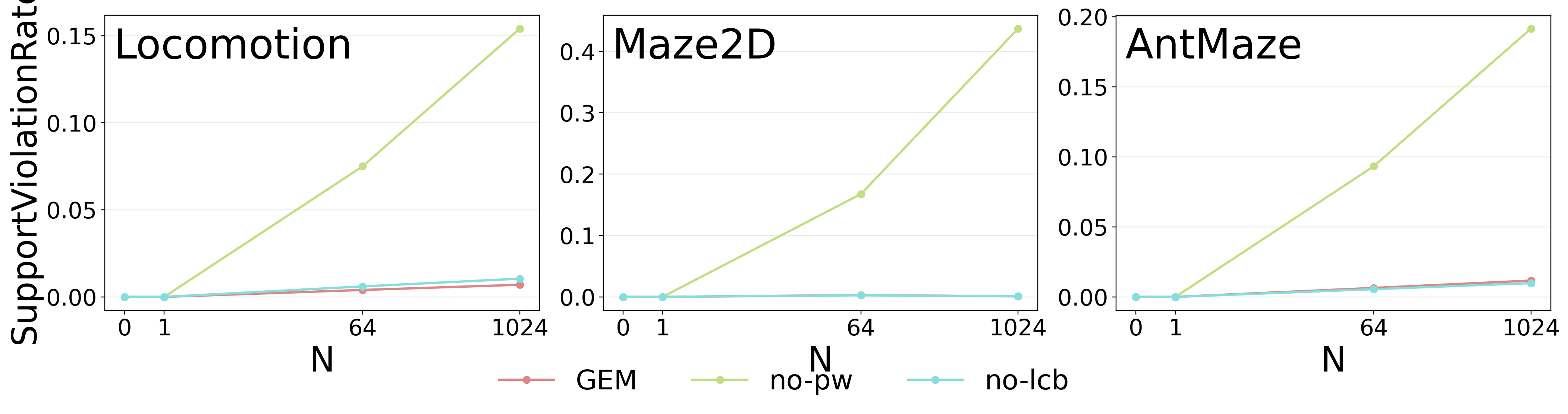}\\
  \includegraphics[width=\linewidth]{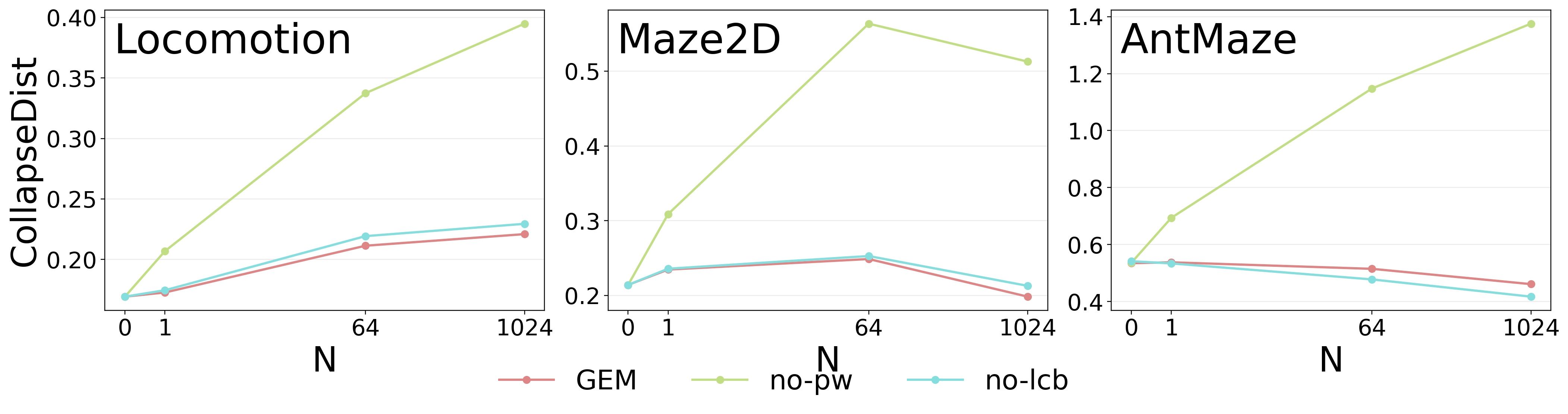}
  \caption{Suite-level scaling with candidate budget $N$ (Score on top). We report suite means for normalized score and the two deployment audits as functions of $N$. Environment-level breakdowns are in Appendix~\ref{app:env_breakdowns}.}
  \label{fig:suite_scaling_stack}
\end{figure}

Figure~\ref{fig:suite_scaling_stack} evaluates the central interface constraint from Section~\ref{sec:prelim_motivation}: increasing $N$ strengthens maximization pressure, so a viable deployment interface should improve decision quality without systematically drifting toward weakly supported actions.
Across the tested budgets, increasing $N$ yields consistent gains in suite-mean score on the stable suites (Locomotion and Maze2D), while the two audits remain controlled under GEM.
The improvement from small budgets to large budgets is most pronounced up to $N{=}1024$, after which additional gains empirically diminish in our sweeps; we therefore use $N{=}1024$ as a principled default that is both a strong setting and a stress point for maximization.
Full environment-level scaling plots, which reveal where the suite mean is dominated by a subset of tasks, are provided in Appendix~\ref{app:env_breakdowns}.

This scaling study also clarifies what is meant by ``trades compute for decision quality without retraining'' in practice.
Changing $N$ is purely a deployment-time compute knob, and it composes with the test-time inference sweep in Section~\ref{sec:results_main}:
 both $N$ and the $(\lambda,w_p)$ terms in Eq.~\ref{eq:score} can be adjusted after training to improve outcomes or reduce risk, and the effect can be audited immediately through \textsc{Violation} and \textsc{CollapseDist}.

\subsection{Deployment compute profile: parallel candidates versus iterative inference}
\label{sec:results_tradeoff}

\begin{figure}[t]
  \centering
  \includegraphics[width=\linewidth]{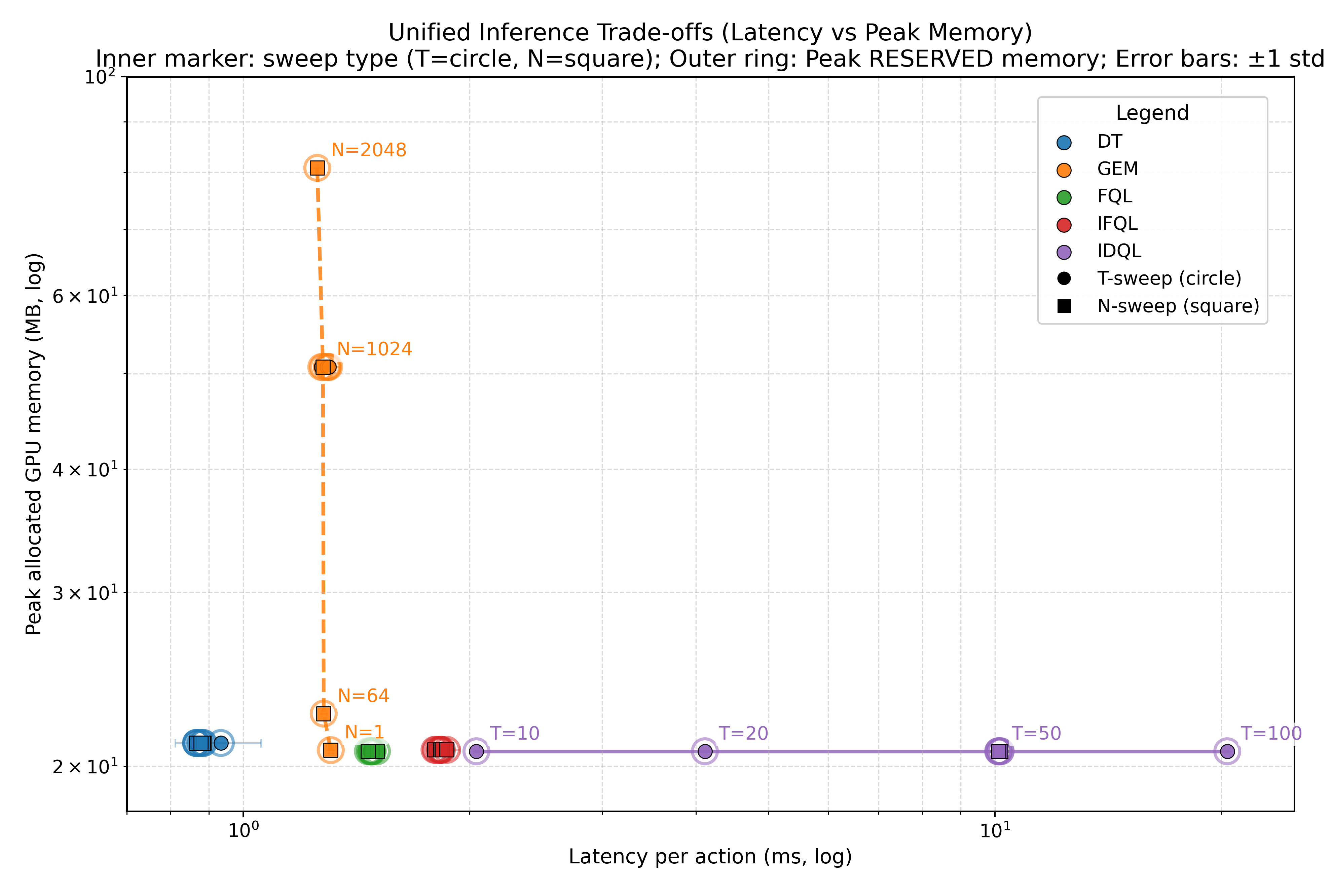}
  \caption{Deployment compute profile under the shared measurement harness (latency and peak memory) across all logged methods and sweep settings.}
  \label{fig:master_tradeoff}
\end{figure}

Figure~\ref{fig:master_tradeoff} characterizes the practical cost shape of GEM’s deployment knobs and contrasts it with iterative generative inference.
Two measured facts matter for the mechanism story.
First, for GEM the latency is roughly unchanged as $N$ increases across the measured settings ($1.25$--$1.33$ ms from $N{=}1$ to $N{=}2048$), while peak allocated memory increases substantially (from $20.76$ MB at $N{=}1$ to $80.83$ MB at $N{=}2048$; full numbers in Table~\ref{tab:tradeoff_full} in Appendix~\ref{app:tradeoff_full}).
This indicates that GEM’s scaling largely manifests as a parallel scoring workload that saturates GPU throughput while trading additional memory for a larger candidate tensor, rather than increasing sequential latency.

Second, methods with an explicit iterative-step inference knob exhibit a different cost shape: in our harness, IDQL shows a near $10\times$ latency increase when increasing the step count ($2.04$ ms at $T{=}10$ versus $20.37$ ms at $T{=}100$; Table~\ref{tab:tradeoff_full} in Appendix~\ref{app:tradeoff_full}).
This is the practical diffusion-style inference tax: sequential refinement increases wall-clock latency step-by-step, while GEM expands candidates in parallel and mainly pays in GPU memory rather than time.

This contrast reflects a mechanism-level knob difference: GEM scales mainly via parallel batch size (GPU throughput/memory), whereas diffusion-style methods scale via sequential depth (latency).
At the same time, we do not claim universality across implementations; we report the complete measured table and encourage treating it as an engineering profile rather than a theoretical guarantee.

\subsection{Stress test at $N{=}1024$: isolating interface controls}
\label{sec:results_stress}

We now fix $N{=}1024$ and ablate interface components to isolate which controls prevent large-$N$ maximization from selecting unsupported or unreliable actions.
We use \textsc{Violation} and \textsc{CollapseDist} to decompose deployment risk into two distinguishable signals—support drift and mode deviation—so we can validate that support and pessimism play complementary roles under candidate maximization.
Full suite-level ablation delta plots (Score/\textsc{Violation}/\textsc{CollapseDist}) are in Appendix~\ref{app:ablations} (Figure~\ref{fig:app_ablation_suite}).

Removing behavior-normalized support (\texttt{no\_pw}) is the clearest maximization failure mode: with many candidates, the selector becomes more likely to pick weak-support outliers, which shows up directly as higher \textsc{Violation}.
Removing pessimism (\texttt{no\_lcb}) exposes a different axis: score can degrade and \textsc{CollapseDist} can increase even when \textsc{Violation} does not rise as sharply, consistent with uncertainty-driven misranking that can persist on-support under function approximation.
These two signatures are precisely why Eq.~\ref{eq:score} keeps both terms: support constrains dataset density and mitigates OOD querying, while pessimism reduces budget-driven extreme-value amplification under critic noise.

Proposal structure remains first-class because candidate maximization can only choose among proposed hypotheses.
Ablations that restrict candidate sources or remove anchoring change the hypothesis set and increase reliance on sampling luck, which can shift both score and audits under large $N$.
This reinforces the proposal/selection decomposition emphasized in Section~3: selection controls cannot recover value that proposals never include, and deterministic anchoring reduces tail-risk from unlucky candidate draws.

\subsection{Support normalization and knob semantics}
\label{sec:results_support_knob}

This subsection tests whether the support weight behaves as an interpretable deployment knob across states and candidate budgets.
The goal is calibration: validate that candidate-set standardization makes $w_p$ comparable across states, not to claim that z-score normalization strictly dominates raw likelihood in every environment.
The corresponding suite plots are moved to Appendix~\ref{app:support_knob_plots}: normalization-mode deltas are in Figure~\ref{fig:app_mode_delta_stack}, and the $w_p$ sweep frontier is in Figure~\ref{fig:app_pw_frontier_suite}.
Changing only the normalization mode tests scale sensitivity.
Because raw $\log\mu_\varphi(a\mid s)$ can vary substantially across states, an unnormalized support term can make $w_p$ behave like a dataset- and state-dependent knob.
Candidate-set z-scoring removes this affine scale variability within each queried state, making $w_p$ closer to an exchange rate between conservative value and relative support measured in candidate-set standard deviations.
The $w_p$ frontier plot visualizes this semantics directly: increasing $w_p$ pushes the selector toward relatively higher-support candidates within the current candidate set, typically reducing \textsc{Violation} at the cost of some return, without retraining.
This is the intended deployment contract: the interface exposes controllable risk-return tradeoffs that remain inspectable under maximization.

% =========================
\section{Discussion: Test-time Inference Interfaces in Offline RL}
\label{sec:discussion}
% =========================
Offline RL is often framed as learning under distribution shift, but our results point to an equally decisive factor: the \emph{test-time inference interface} decides which $(s,a)$ pairs the critic is actually queried on. Under candidate maximization, the interface is not an implementation detail—it is part of the algorithm’s contract, governing how $\arg\max$ amplifies error, uncertainty, and support.

\subsection{What GEM changes at test time}
GEM turns inference into an explicit, tunable interface: it proposes hypotheses in parallel (with a deterministic anchor) and selects using a fixed-form score that couples conservative value (ensemble LCB) with an \emph{independent} support signal normalized per-state over the candidate set. This yields retrain-free knobs—$N$, $\lambda$, and the $w_p$ schedule—that trade compute and risk against return, with effects directly visible in both performance and audits.

\subsection{Mechanistic lessons from the audits}
Candidate maximization creates an extreme-value hazard: as $N$ grows, the chance of selecting a spurious maximizer rises. The ablations isolate two controls with non-overlapping roles: support primarily prevents coverage failures (low-support selections), while pessimism mitigates estimation risk (high-disagreement outliers that can remain problematic even on-support). Reporting \textsc{Violation} and \textsc{CollapseDist} separately therefore serves a purpose: it decomposes failure modes so post-training tuning can target the right risk.

A second lesson is knob semantics. Raw $\log\mu_\varphi(a\mid s)$ has heterogeneous scale across states and tasks; candidate-set standardization makes $w_p$ behave closer to a cross-state exchange rate between conservative value and \emph{relative} support within the current candidate set. This is a calibration claim—stable meaning—not a claim of universal dominance.

\subsection{Compute shape as a deployment consideration}
Compute tradeoffs are resource-shaped. In our harness, scaling $N$ primarily increases memory with little latency change, consistent with parallel scoring; iterative refinement, in contrast, increases sequential latency with steps. These knobs therefore target different deployment regimes, and reporting the cost “shape” makes compute--quality tradeoffs interpretable beyond headline scores.

\subsection{Limits and implications}
The interface view also makes limits concrete. AntMaze often benefits from stitching or waypoint-level planning; we intentionally use a no-stitching protocol to isolate interface effects. Support audits inherit the quality of $\mu_\varphi$, and memory grows with $N$, which can constrain high-dimensional action spaces. Finally, conservative support-aligned selection can cap return when high-reward actions are rare or poorly modeled by $\mu_\varphi$, motivating better support modeling and state-dependent budgeting.

Two natural directions follow: (i) adaptive inference that allocates $N$ and $(\lambda,w_p)$ from uncertainty/support cues, and (ii) richer proposal families that retain an independent, calibrated support interface. The broader takeaway is simple: deployable offline RL is not just a learned critic and actor—it is the inference interface that queries them, and making that interface controllable and auditable is a direct lever for reliability without retraining.

\section*{Impact Statement}

Our work aims to improve the reliability of offline reinforcement learning (RL) when the deployed environment differs from the offline data distribution, by better handling multi-modal action distributions and reducing pathological action selection under distribution shift. If successful, such methods could benefit real-world applications where online data collection is costly or risky (e.g., robotics, industrial control), potentially improving data efficiency and reducing the need for unsafe exploration.

However, offline RL methods can also amplify harms if deployed in safety-critical settings without sufficient safeguards. In particular, failures under distribution shift may lead to unsafe actions, and performance can be sensitive to the quality, coverage, and biases of the offline dataset. More expressive policies that capture multiple behavioral modes may additionally be misapplied to select actions that appear plausible under the model but are undesirable in practice. These risks are not unique to our approach but are relevant to its intended use.

To mitigate potential negative impacts, we emphasize that the proposed method should be used with (i) strong offline evaluation across diverse conditions and stress tests for distribution shift, (ii) conservative deployment practices such as action constraints, safety filters, or human oversight when applicable, and (iii) careful dataset documentation and auditing to understand coverage gaps and bias. We also encourage reporting failure cases and ablations that clarify when the method degrades, to reduce the likelihood of overconfident real-world use.

\bibliography{cites}
\bibliographystyle{plainnat}

\newpage
\appendix
\onecolumn

% Keep the same graphic search paths as main text
\graphicspath{{main_figs/}{sub_figs/}{group_figs/}{figures/}}
\section{Additional Diagnostics and Full Results}
\label{app:additional}

% -------------------------------------------------
\subsection{NLL-gap diagnostic: multimodality is widespread}
\label{app:nll_gap}

We use the NLL-gap diagnostic to audit whether multimodal conditional action structure is present broadly in the offline datasets, rather than being a peculiarity of a few benchmarks.
For each environment we compute
$\mathrm{gap}=\mathrm{NLL}_{\mathrm{top1}}-\mathrm{NLL}_{\mathrm{gmm}}$,
where $\mathrm{NLL}_{\mathrm{gmm}}$ evaluates the fitted $K$-component GMM likelihood and $\mathrm{NLL}_{\mathrm{top1}}$ evaluates only the most-weighted component (top-1 proxy).
A positive gap indicates that mixture structure provides a strictly better likelihood fit than any single dominant mode, consistent with multimodal conditional action structure.

Across the 18 evaluated D4RL environments, the gap is consistently positive (Table~\ref{tab:nll_gap_all_env}), supporting the paper’s motivation that multimodality is a common dataset property that a unimodal projection can blur into weak-support ``in-between'' actions.

\begin{figure}[t]
  \centering
  \includegraphics[width=\linewidth]{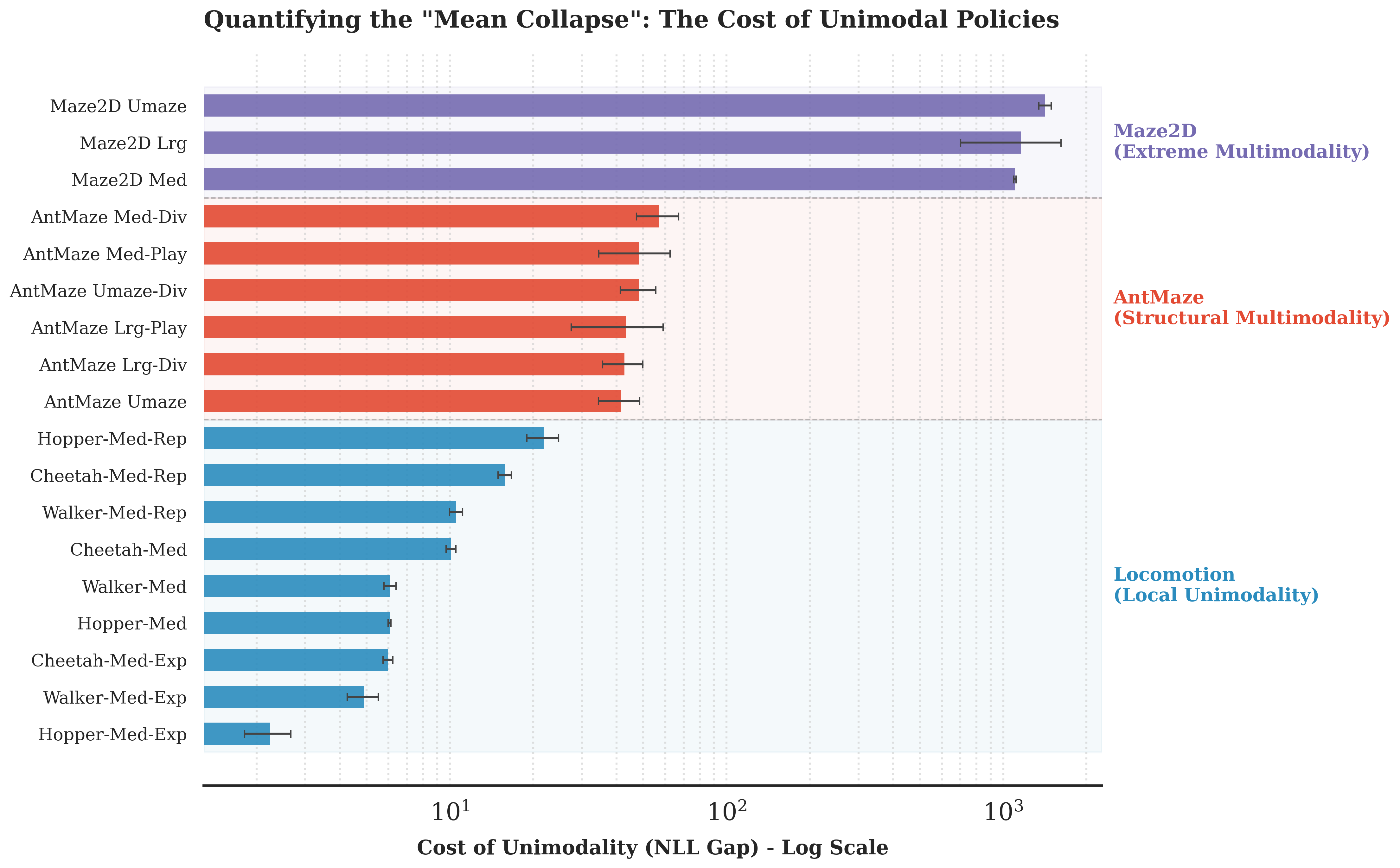}
  \caption{NLL-gap diagnostic summary. Larger positive gaps indicate stronger mixture benefits over a top-1 unimodal proxy.}
  \label{fig:nll_gap_summary}
\end{figure}

\begin{table}[t]
\centering
\caption{Per-environment NLL-gap statistics ($\mathrm{gap}=\mathrm{NLL}_{\mathrm{top1}}-\mathrm{NLL}_{\mathrm{gmm}}$; mean$\pm$std). All entries are positive in the provided results, indicating consistent mixture benefits.}
\label{tab:nll_gap_all_env}
\footnotesize
\begin{tabular}{l r}
\hline
Environment & Mean gap $\pm$ std \\
\hline
Maze2D-umaze & 168.29$\pm$7.71 \\
Maze2D-medium & 147.30$\pm$6.85 \\
Maze2D-large & 142.64$\pm$7.34 \\
Antmaze-umaze-diverse & 38.80$\pm$3.46 \\
Antmaze-umaze & 31.53$\pm$2.83 \\
Antmaze-medium-diverse & 23.73$\pm$2.68 \\
Antmaze-medium-play & 22.04$\pm$2.44 \\
Antmaze-large-diverse & 19.02$\pm$2.46 \\
Antmaze-large-play & 15.89$\pm$2.23 \\
Walker2d-medium & 10.83$\pm$1.52 \\
Walker2d-medium-expert & 9.83$\pm$1.37 \\
Walker2d-expert & 8.76$\pm$1.31 \\
Hopper-medium & 8.63$\pm$1.10 \\
Hopper-medium-expert & 8.21$\pm$1.05 \\
Hopper-expert & 7.48$\pm$1.01 \\
HalfCheetah-medium & 6.65$\pm$0.83 \\
HalfCheetah-medium-expert & 6.28$\pm$0.79 \\
HalfCheetah-expert & 5.72$\pm$0.76 \\
\hline
\end{tabular}
\end{table}

% -------------------------------------------------
\subsection{Phase I training pseudocode (supporting)}
\label{app:phase1_pseudocode}

\begin{algorithm}[t]
\caption{\textsc{GEM} Training (Phase I, supporting)}
\label{alg:gem_phase1_supporting}
\footnotesize
\begin{algorithmic}[1]
\REQUIRE Offline dataset $\mathcal D$; critics $\{Q_i,Q_i^{\text{tgt}}\}_{i=1}^M$, value $V$; actor GMM $\pi_\theta$; behavior GMM $\mu_\varphi$.
\REQUIRE Hyperparams: steps $T$, discount $\gamma$, soft-update $\rho$, expectile $\tau$; guidance temp $\beta$; gate-entropy weight $\alpha$.

\STATE \textbf{Fit behavior density (support-only model).}
\STATE Train $\mu_\varphi$ on $(s,a)\sim\mathcal D$ by MLE: minimize $-\log\mu_\varphi(a|s)$.

\FOR{$t=1$ \textbf{to} $T$}
  \STATE Sample $(s,a,r,s',d)\sim\mathcal D$.
  \STATE Update $V$ by expectile regression with $\mathrm{adv}\leftarrow Q^{\text{tgt}}(s,a)-V(s)$.
  \STATE Update each $Q_i$ toward $y\leftarrow r+\gamma(1-d)V(s')$; soft-update targets with rate $\rho$.
  \STATE \parbox[t]{\linewidth}{Actor update (guided EM-style): $u_k(s,a)=\log w_k(s)+\log\mathcal N_k(a|s)$,\;
$\gamma_k(s,a)=\mathrm{softmax}_k(u_k(s,a))$,\;
$\mathrm{ELBO}_{\text{loose}}(s,a)=\sum_k \gamma_k(s,a)\,u_k(s,a)$ (optionally stopgrad on $\gamma$).}

  \STATE Guidance weight $w(s,a)\propto \exp(\beta\,\mathrm{adv})$ (clamp); guidance reweights samples only.
  \STATE \parbox[t]{\linewidth}{Update $\theta$ by minimizing $\mathbb E_{(s,a)\sim\mathcal D}[\,w(s,a)\cdot(-\mathrm{ELBO}_{\text{loose}}(s,a))\,]-\alpha\,\mathbb E_{s\sim\mathcal D}[H(w(\cdot|s))]$.}
\ENDFOR
\STATE \textbf{Output:} $\{Q_i\}$ for $\mathrm{LCB}_\lambda$; actor gating/means for anchor+proposals; behavior $\mu_\varphi$ for support.
\end{algorithmic}
\end{algorithm}

% -------------------------------------------------
\subsection{Environment-level breakdowns for candidate-budget scaling}
\label{app:env_breakdowns}

This appendix provides the detailed scaling plots referenced in Section~\ref{sec:results_scaling}.
For each suite we provide three separate figures that expand the suite mean into its constituent environments and report Score, \textsc{Violation}, and \textsc{CollapseDist} versus candidate budget $N$.

\begin{figure}[H]
  \centering
  \includegraphics[width=\linewidth]{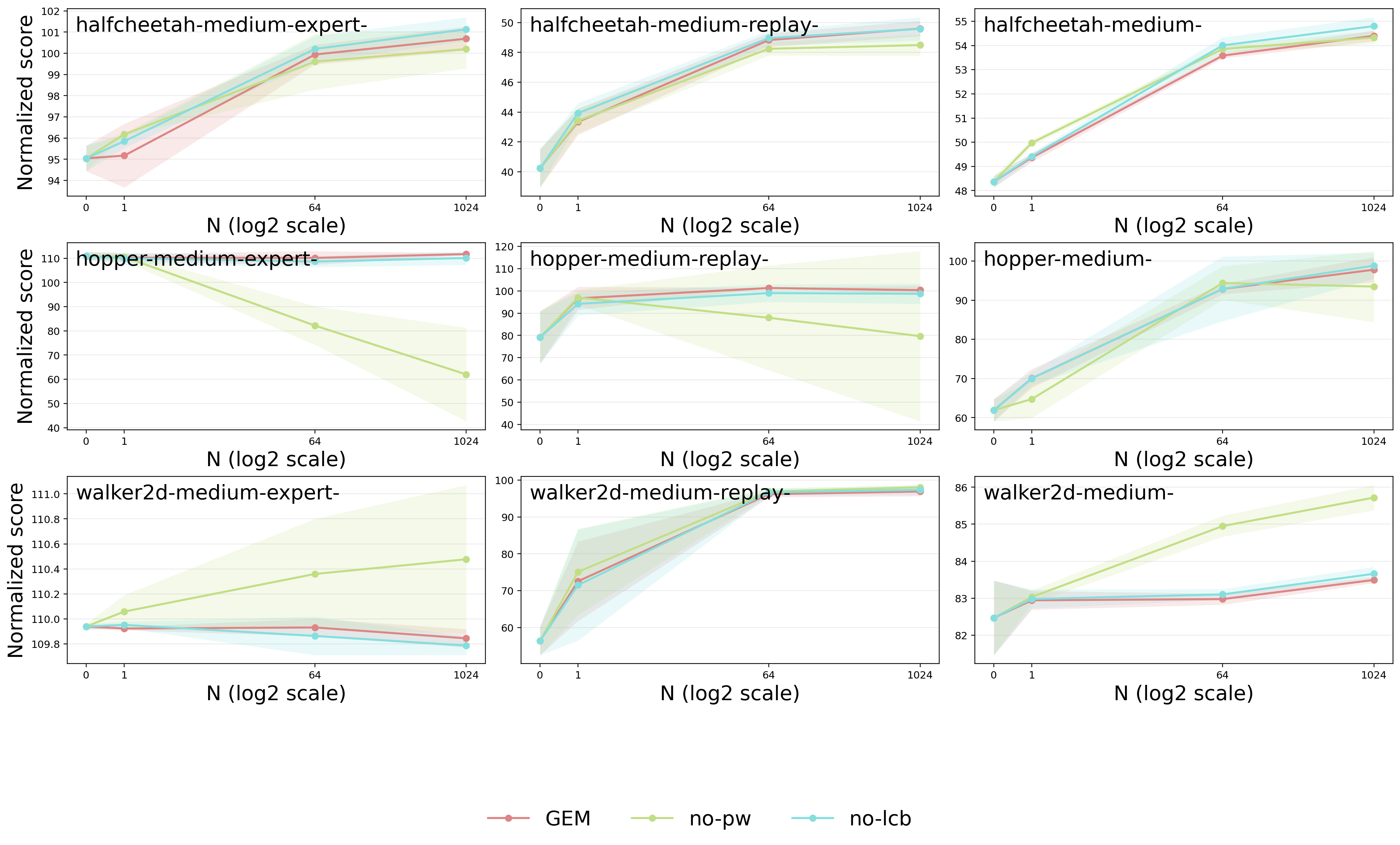}
  \caption{Locomotion suite: per-environment candidate-budget scaling for Score.}
  \label{fig:app_env_scaling_locomotion_score}
\end{figure}

\begin{figure}[H]
  \centering
  \includegraphics[width=\linewidth]{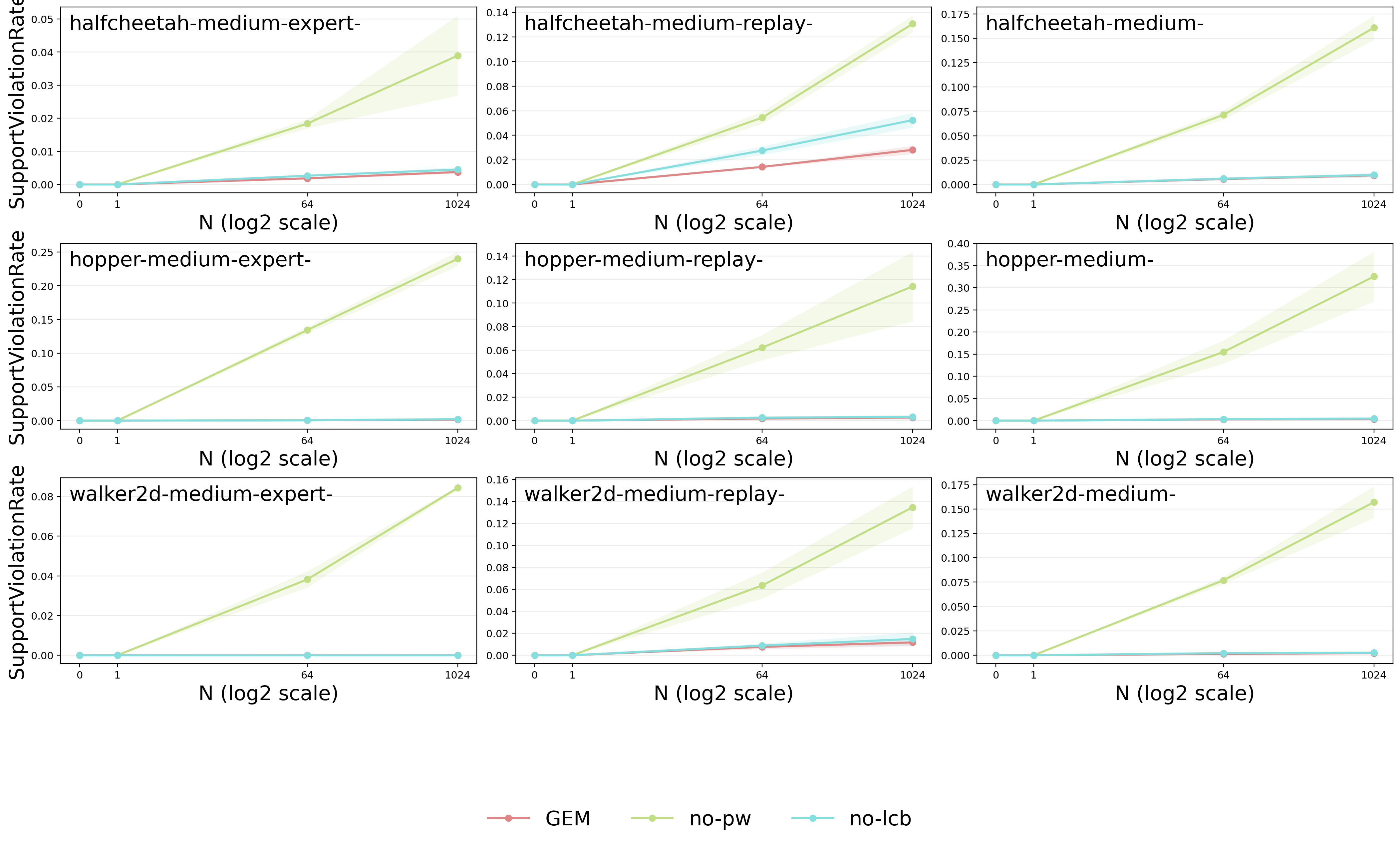}
  \caption{Locomotion suite: per-environment candidate-budget scaling for \textsc{Violation}.}
  \label{fig:app_env_scaling_locomotion_violation}
\end{figure}

\begin{figure}[H]
  \centering
  \includegraphics[width=\linewidth]{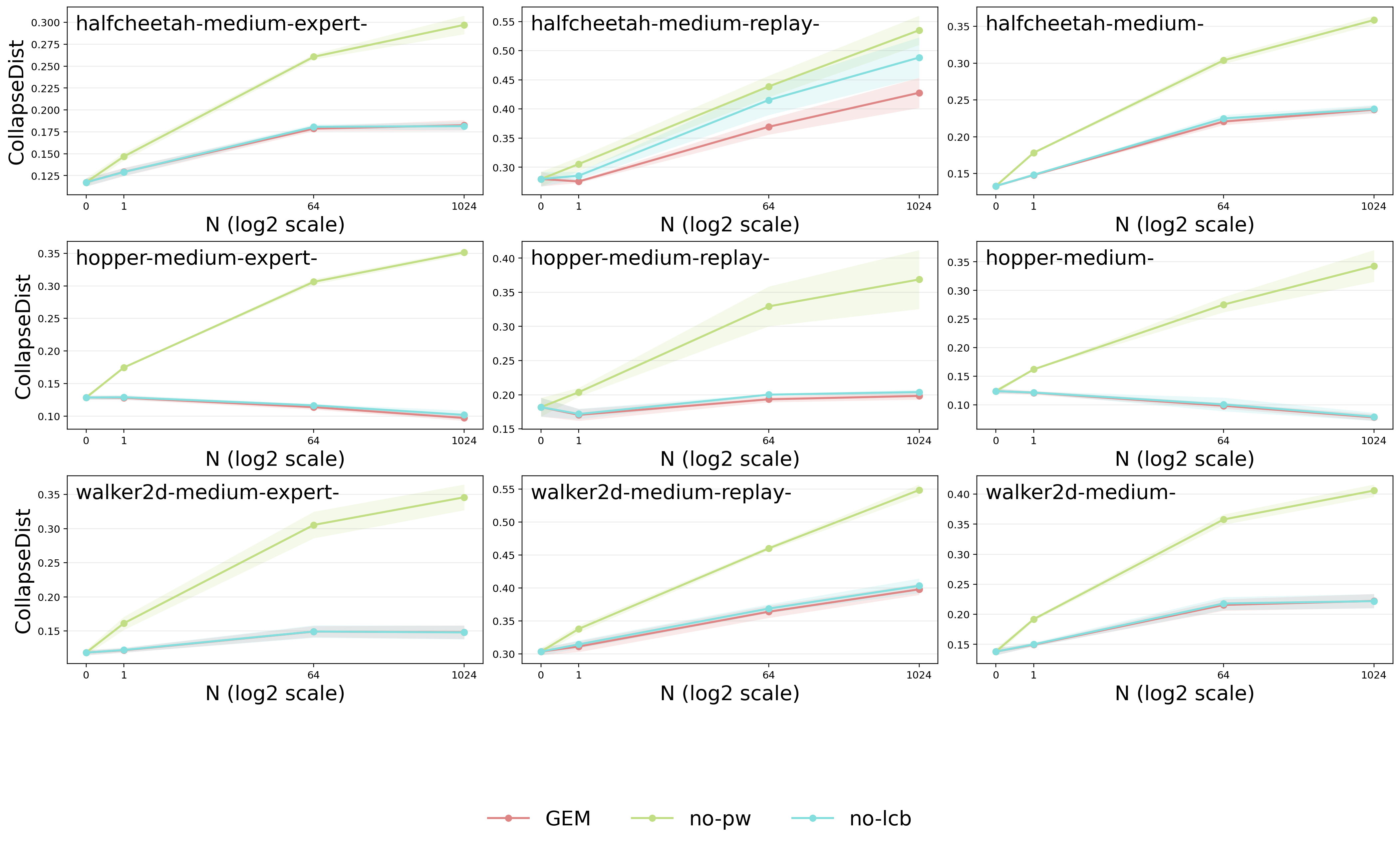}
  \caption{Locomotion suite: per-environment candidate-budget scaling for \textsc{CollapseDist}.}
  \label{fig:app_env_scaling_locomotion_collapsedist}
\end{figure}

\begin{figure}[H]
  \centering
  \includegraphics[width=\linewidth]{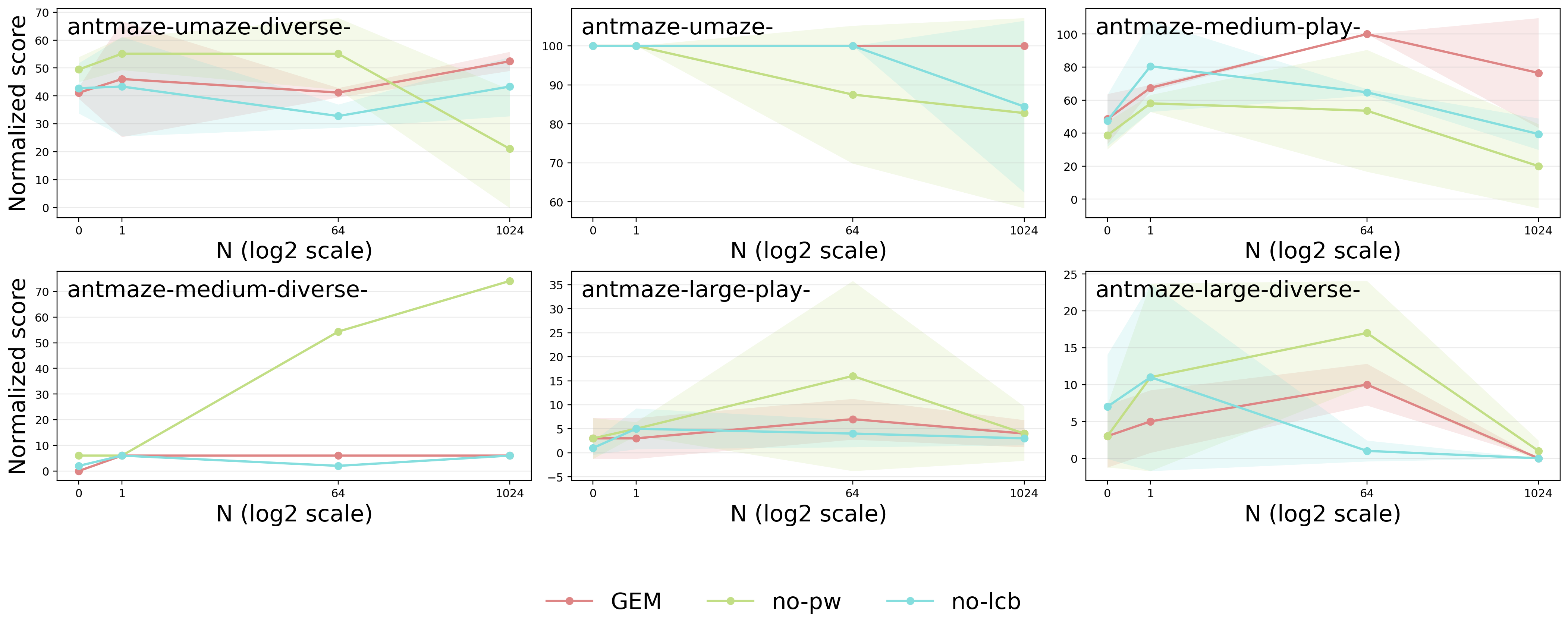}
  \caption{AntMaze suite: per-environment candidate-budget scaling for Score.}
  \label{fig:app_env_scaling_antmaze_score}
\end{figure}

\begin{figure}[H]
  \centering
  \includegraphics[width=\linewidth]{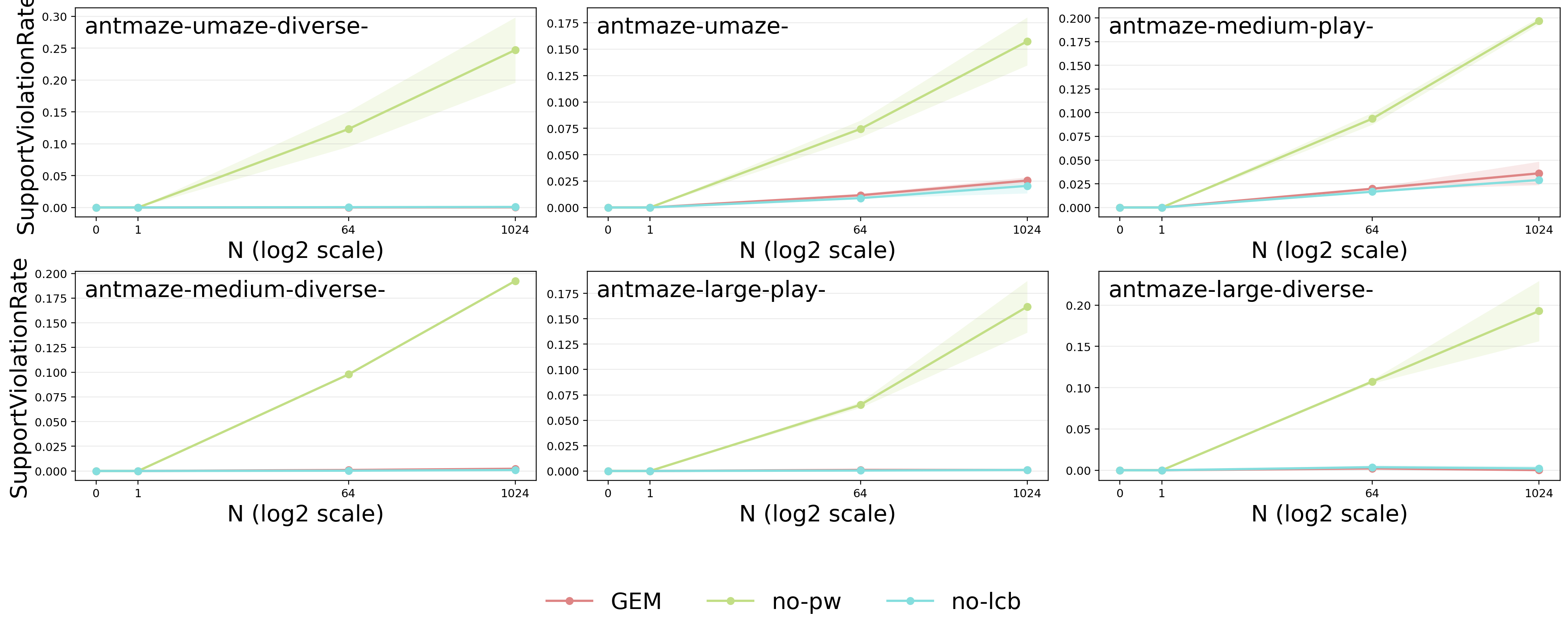}
  \caption{AntMaze suite: per-environment candidate-budget scaling for \textsc{Violation}.}
  \label{fig:app_env_scaling_antmaze_violation}
\end{figure}

\begin{figure}[H]
  \centering
  \includegraphics[width=\linewidth]{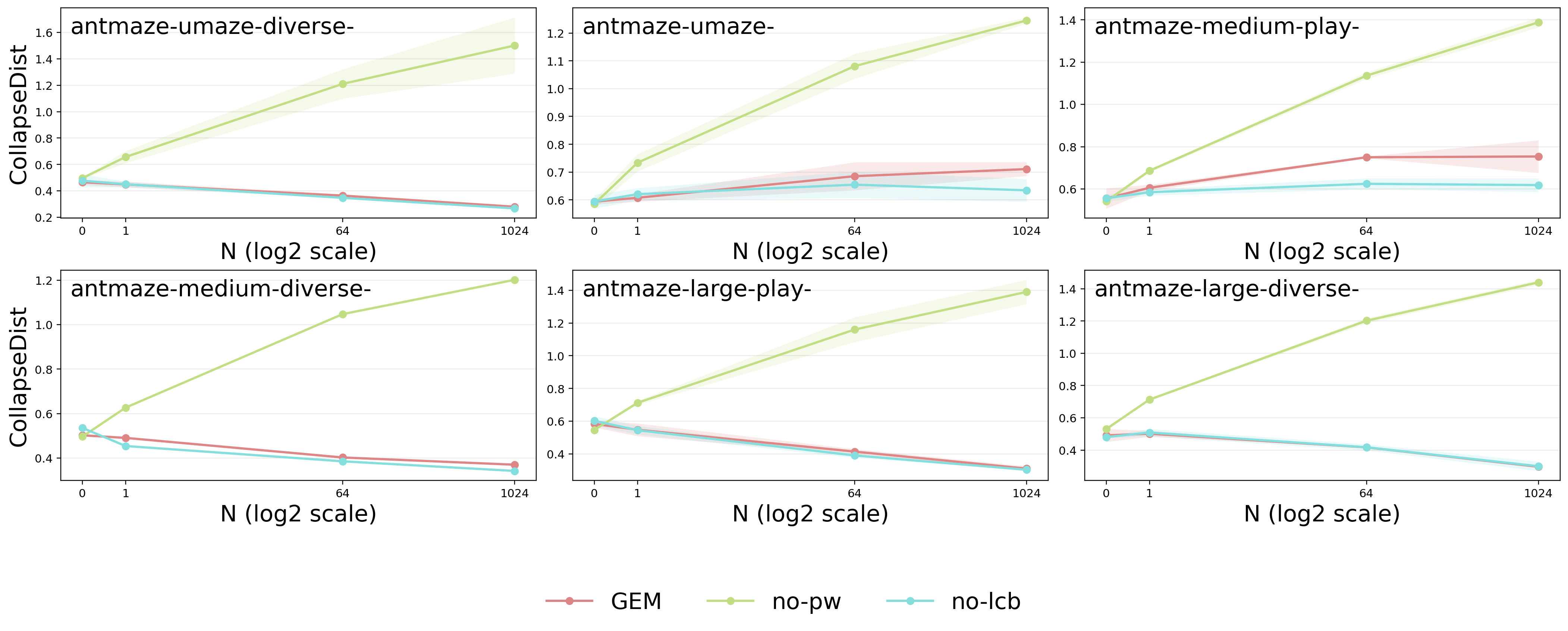}
  \caption{AntMaze suite: per-environment candidate-budget scaling for \textsc{CollapseDist}.}
  \label{fig:app_env_scaling_antmaze_collapsedist}
\end{figure}

\begin{figure}[H]
  \centering
  \includegraphics[width=\linewidth]{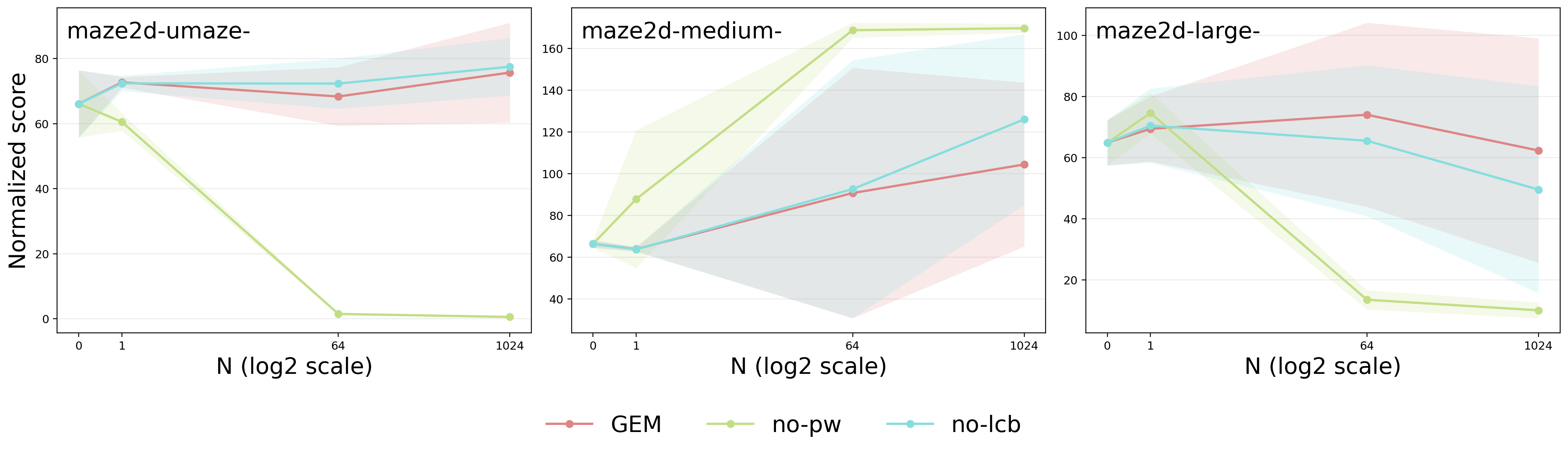}
  \caption{Maze2D suite: per-environment candidate-budget scaling for Score.}
  \label{fig:app_env_scaling_maze2d_score}
\end{figure}

\begin{figure}[H]
  \centering
  \includegraphics[width=\linewidth]{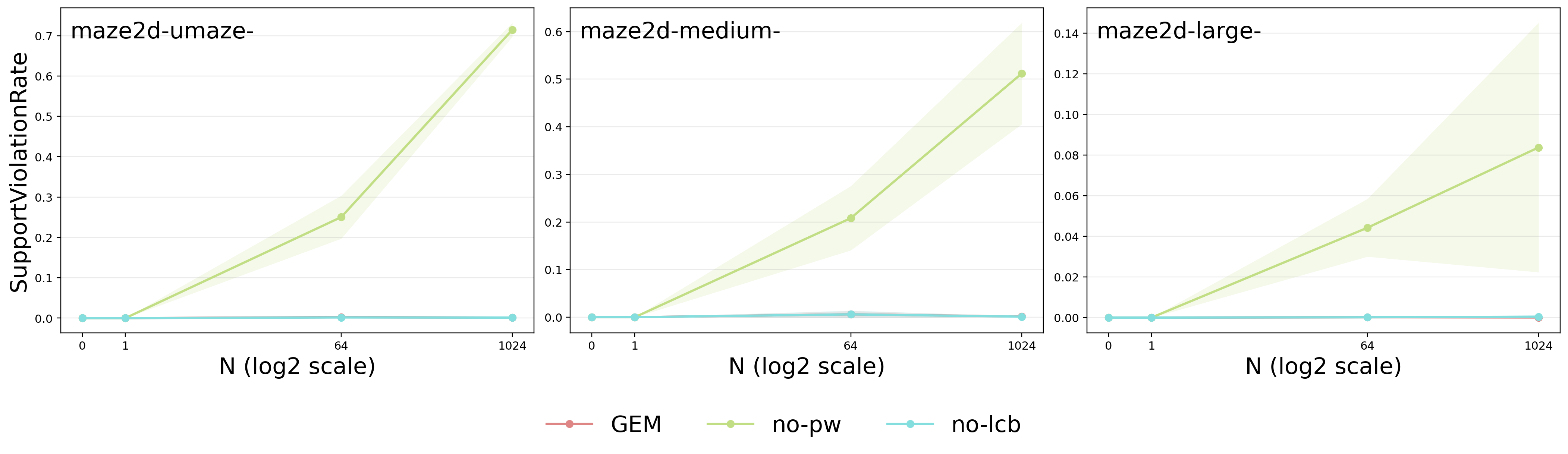}
  \caption{Maze2D suite: per-environment candidate-budget scaling for \textsc{Violation}.}
  \label{fig:app_env_scaling_maze2d_violation}
\end{figure}

\begin{figure}[H]
  \centering
  \includegraphics[width=\linewidth]{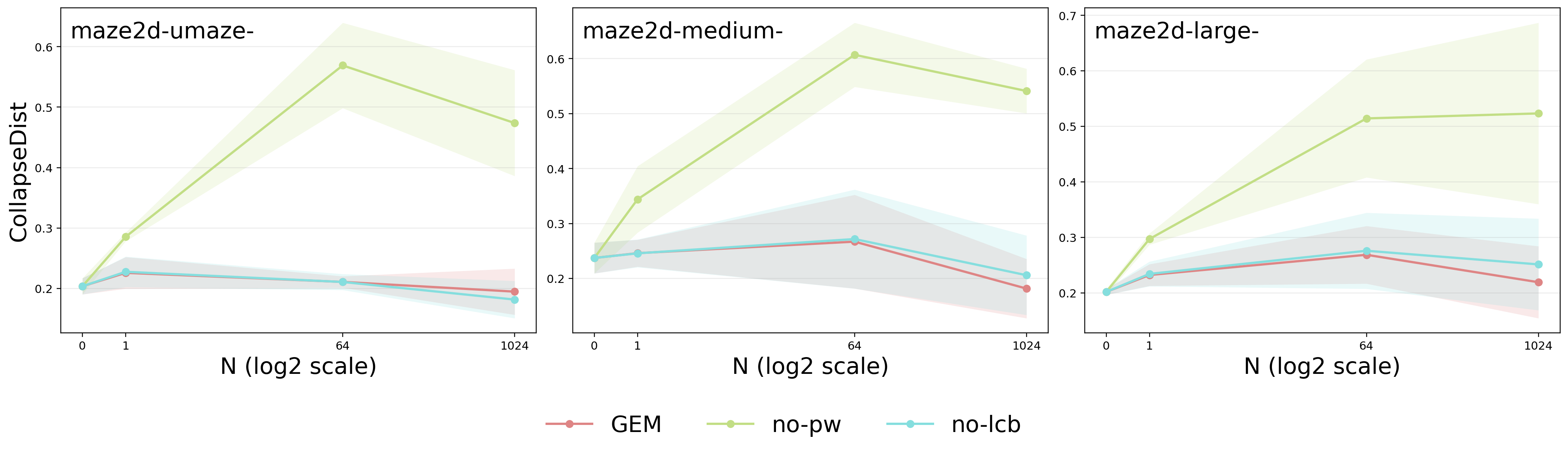}
  \caption{Maze2D suite: per-environment candidate-budget scaling for \textsc{CollapseDist}.}
  \label{fig:app_env_scaling_maze2d_collapsedist}
\end{figure}

% -------------------------------------------------
\subsection{Robustness to dynamics perturbations (MassScale)}
\label{app:robustness_mass}

We evaluate deployment-time robustness to dynamics shift by scaling MuJoCo body masses at test time (MassScale; dashed vertical line at 1.0 indicates the nominal setting), with all learned networks fixed and no retraining.
We report aggregated normalized score trends for GEM and a behavior cloning (BC) reference; shaded regions indicate variability across tasks within each suite.
\textbf{Caveat (AntMaze):} under this MassScale robustness protocol, our AntMaze evaluation collapses to near-zero normalized scores for all methods (including BC), indicating that the perturbation+scoring pipeline is not informative for AntMaze’s sparse-reward goal-reaching; we therefore report the panel for completeness but do not interpret it as algorithmic failure.

\begin{figure}[t]
  \centering
  \includegraphics[width=\linewidth]{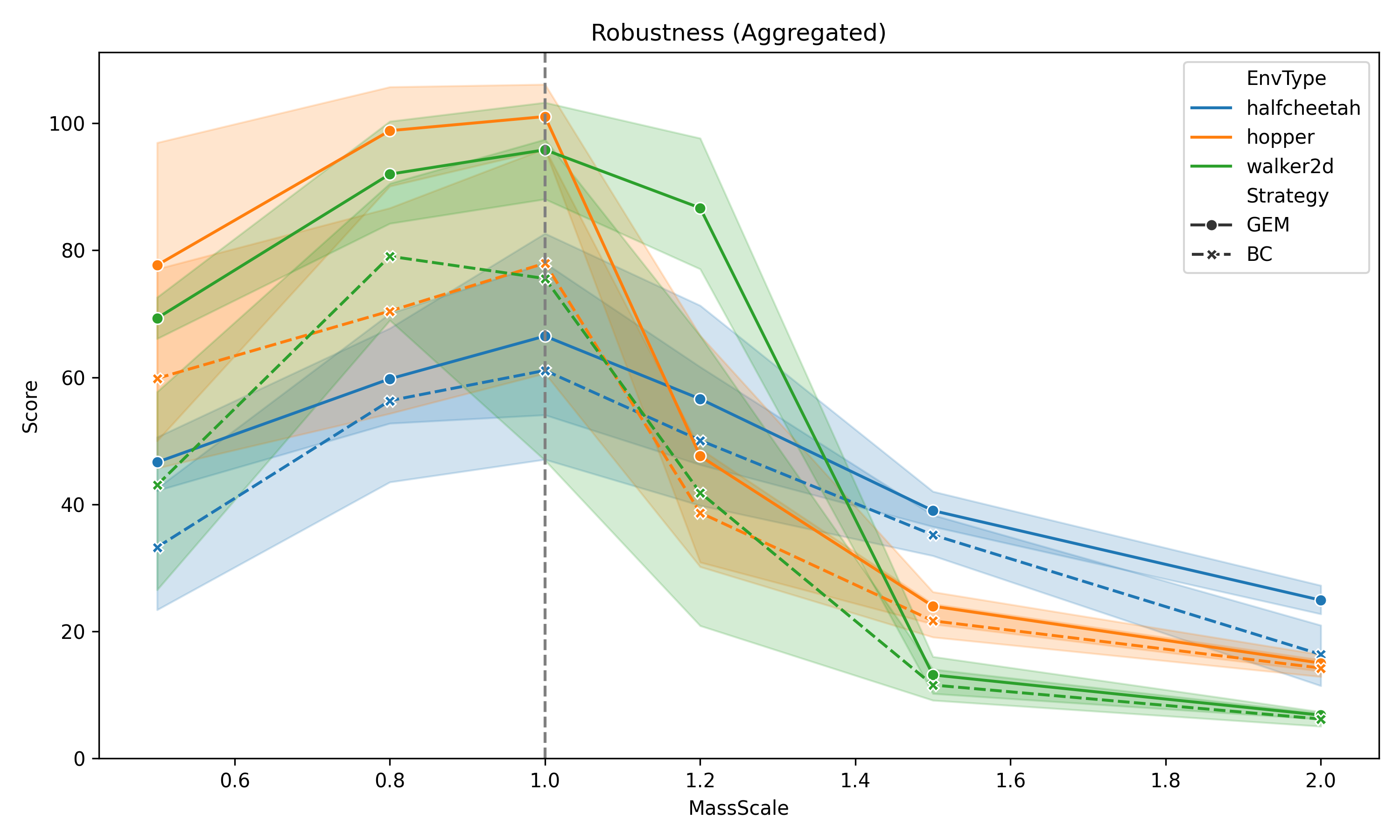}
  \caption{Robustness (aggregated) under mass-scaling perturbations on the Locomotion suite.}
  \label{fig:app_massscale_locomotion}
\end{figure}

\begin{figure}[t]
  \centering
  \includegraphics[width=\linewidth]{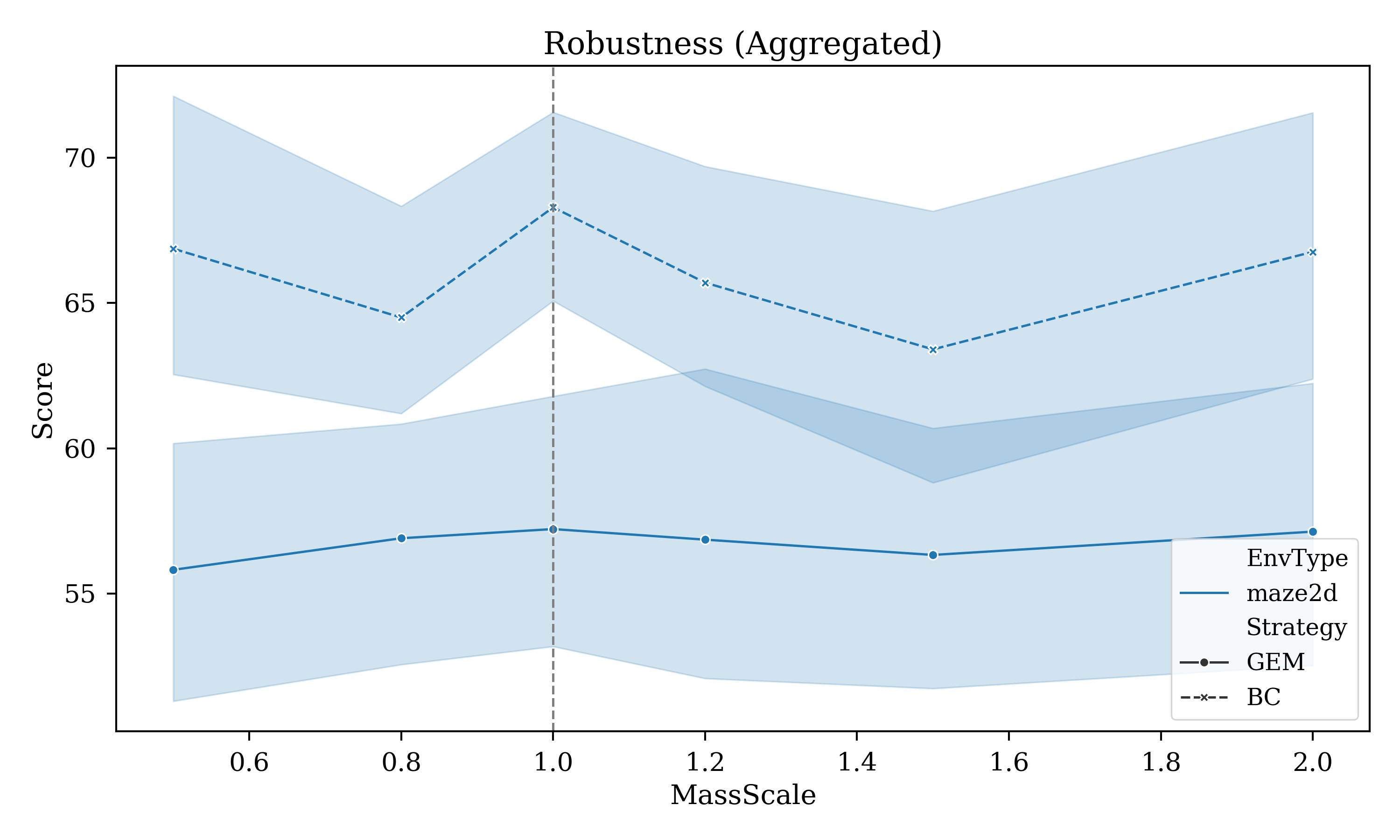}
  \caption{Robustness (aggregated) under mass-scaling perturbations on Maze2D.}
  \label{fig:app_massscale_maze2d}
\end{figure}

\begin{figure}[t]
  \centering
  \includegraphics[width=\linewidth]{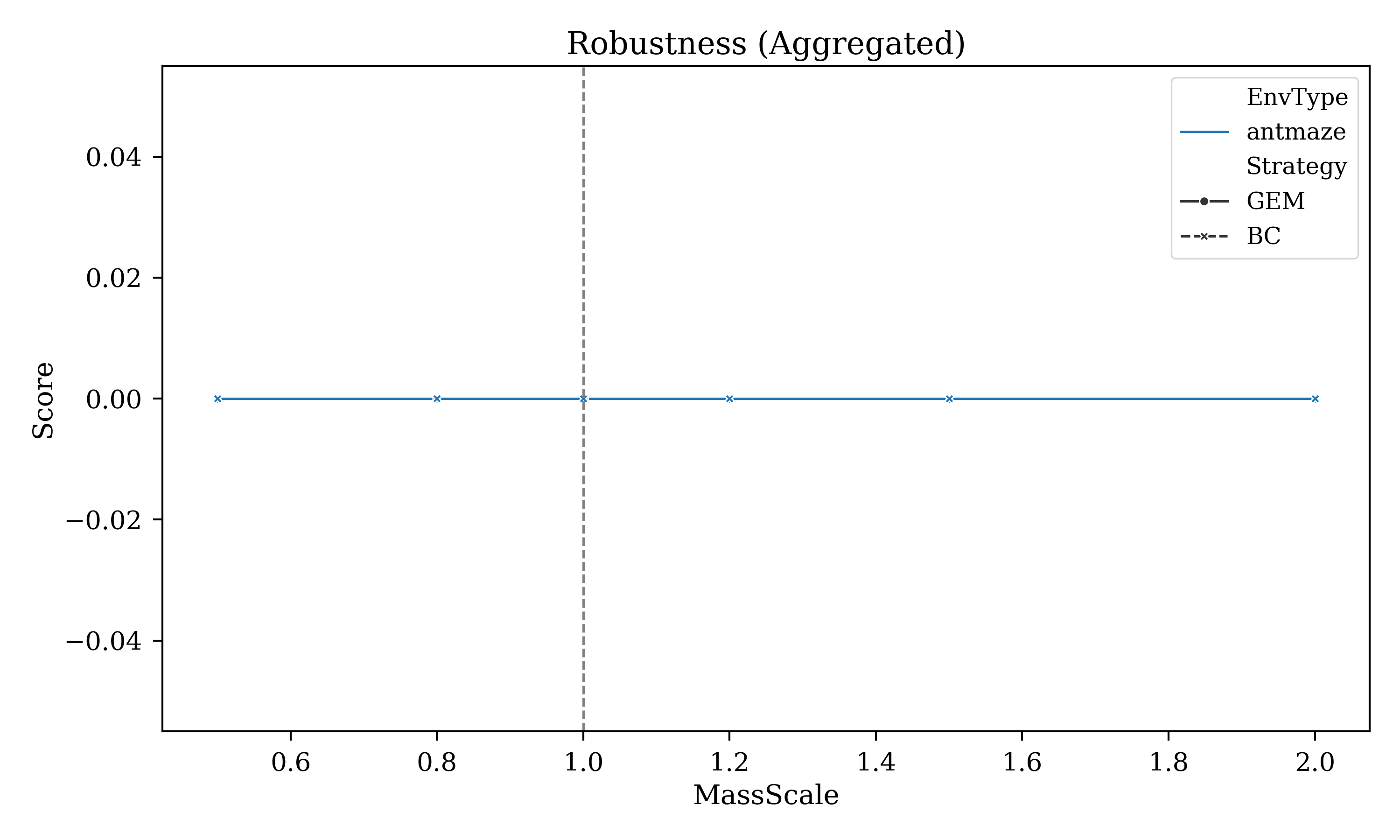}
  \caption{Robustness (aggregated) under mass-scaling perturbations on AntMaze. Near-zero scores across MassScale indicate widespread failure under this dynamics shift under our evaluation protocol.}
  \label{fig:app_massscale_antmaze}
\end{figure}

% -------------------------------------------------
\subsection{Stress-test ablations at $N{=}1024$ (suite deltas)}
\label{app:ablations}

This appendix collects the full suite-level delta plots referenced in Section~\ref{sec:results_stress}.
All variants change exactly one interface factor while holding the remaining test-time inference knobs fixed to the suite defaults reported in Section~\ref{sec:results_main}.

\begin{figure}[t]
  \centering
  \includegraphics[width=\linewidth]{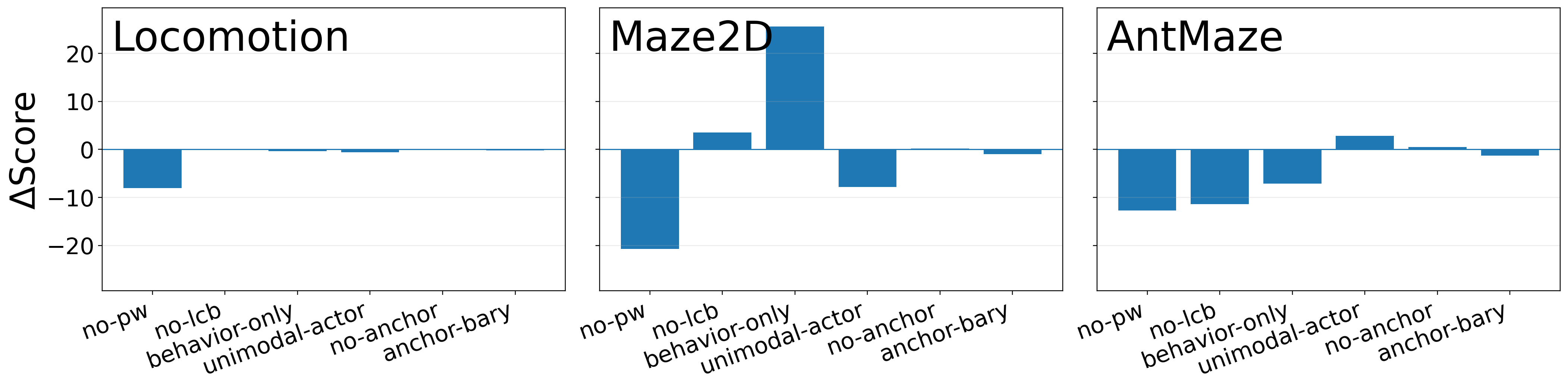}\\
  \includegraphics[width=\linewidth]{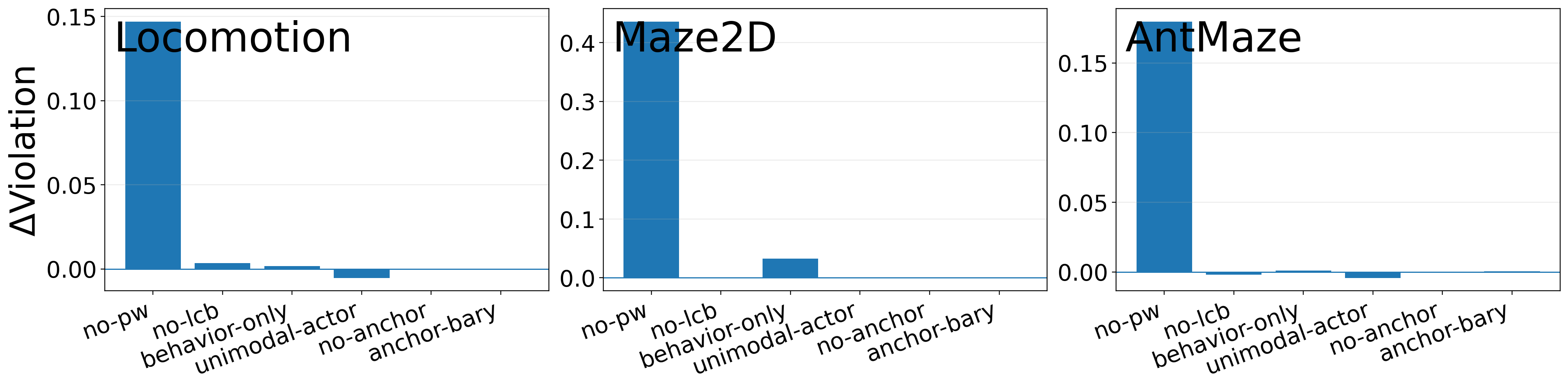}\\
  \includegraphics[width=\linewidth]{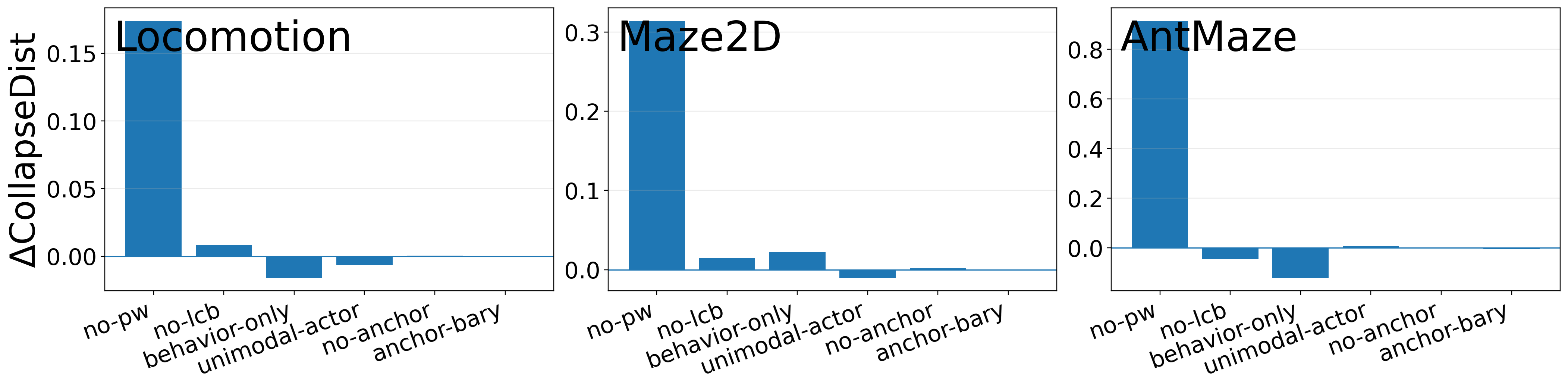}
  \caption{Suite-level interface ablations at $N{=}1024$ (deltas relative to GEM). Top: $\Delta$Score. Middle: $\Delta$\textsc{Violation}. Bottom: $\Delta$\textsc{CollapseDist}.}
  \label{fig:app_ablation_suite}
\end{figure}

% -------------------------------------------------
\subsection{Support normalization and $w_p$ knob semantics}
\label{app:support_knob_plots}

This appendix provides the calibration evidence referenced in Section~\ref{sec:results_support_knob}:
(i) a normalization-mode comparison that changes \emph{only} the support normalization used inside the inference score,
and (ii) a post-training $w_p$ sweep that visualizes the intended knob semantics.

\paragraph{What is being tested.}
Recall the inference score (Eq.~\ref{eq:score}):
\[
\mathrm{Score}(s,a)=\mathrm{LCB}_\lambda(s,a)+w_p\cdot \mathrm{Support}(s,a),
\]
where GEM’s default uses candidate-set standardized support
$\mathrm{Support}(s,a)=\mathrm{zscore}_s(\log\mu_\varphi(a\mid s))$ (Eq.~\ref{eq:zscore}).
The goal here is \emph{calibration}, i.e., to validate that $w_p$ behaves like a state-wise exchange rate between conservative value and \emph{relative} support within the current candidate set.
Importantly, this is \textbf{not} a universality claim that z-score normalization strictly dominates raw likelihood in every environment.

\paragraph{Normalization-mode delta (z-score vs raw).}
We compare two inference-time variants that share the same trained networks and all other inference knobs:
\begin{itemize}
\item \textbf{Mode=\texttt{zscore} (default):} $\mathrm{Support}(s,a)=\mathrm{zscore}_s(\log\mu_\varphi(a\mid s))$.
\item \textbf{Mode=\texttt{raw}:} $\mathrm{Support}(s,a)=\log\mu_\varphi(a\mid s)$ (no per-state candidate-set standardization).
\end{itemize}
In both cases, we keep the candidate budget fixed at $N{=}1024$ and use the suite-specific $\lambda$ and the same $w_p$ schedule as in Section~\ref{sec:results_main},
so the only changed factor is the normalization used inside the score.
We report suite-mean deltas (Mode=\texttt{zscore} minus Mode=\texttt{raw}) for normalized return and the two deployment audits.

\begin{figure}[t]
  \centering
  \includegraphics[width=0.3\linewidth]{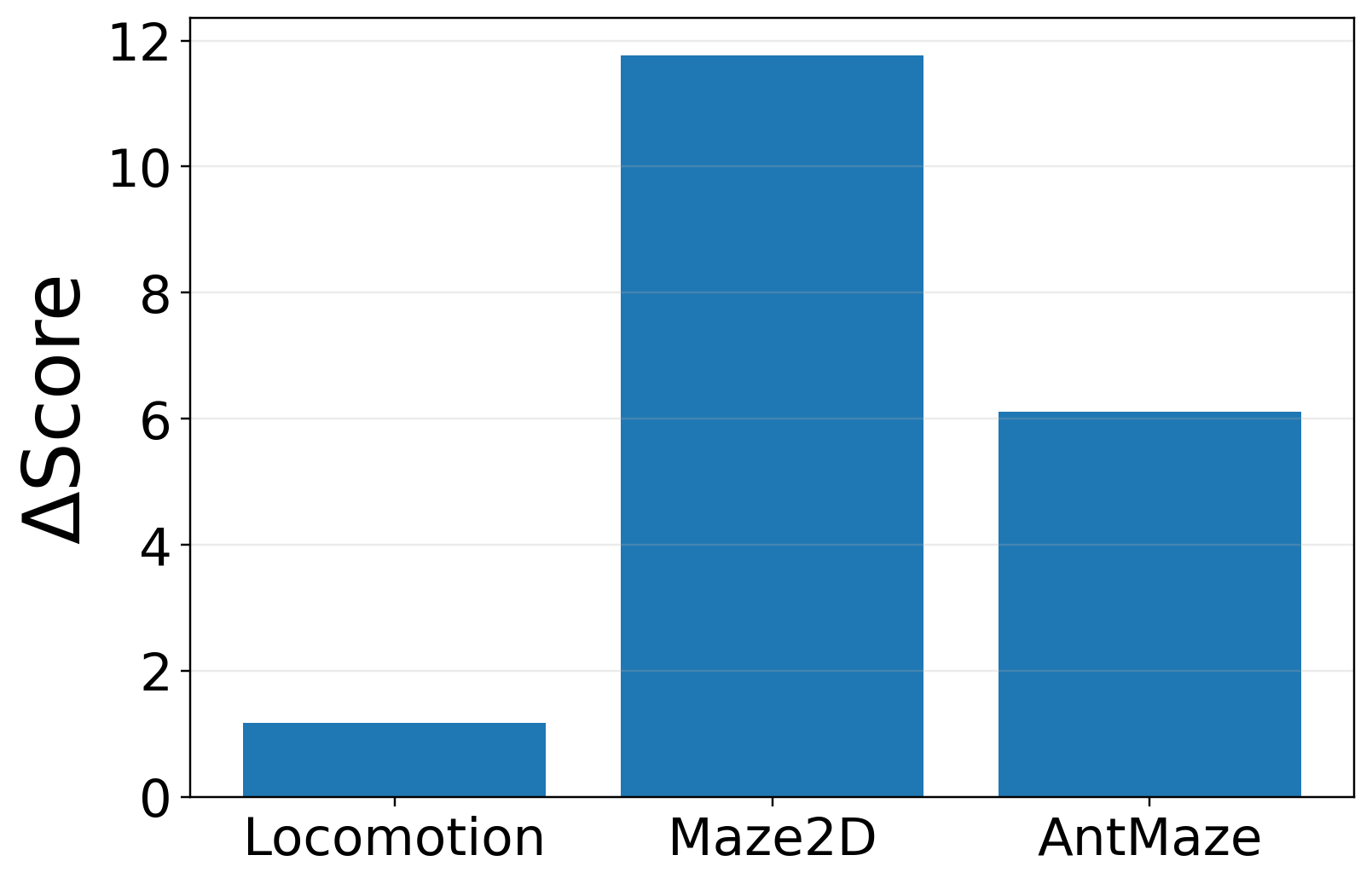}\\
  \includegraphics[width=0.3\linewidth]{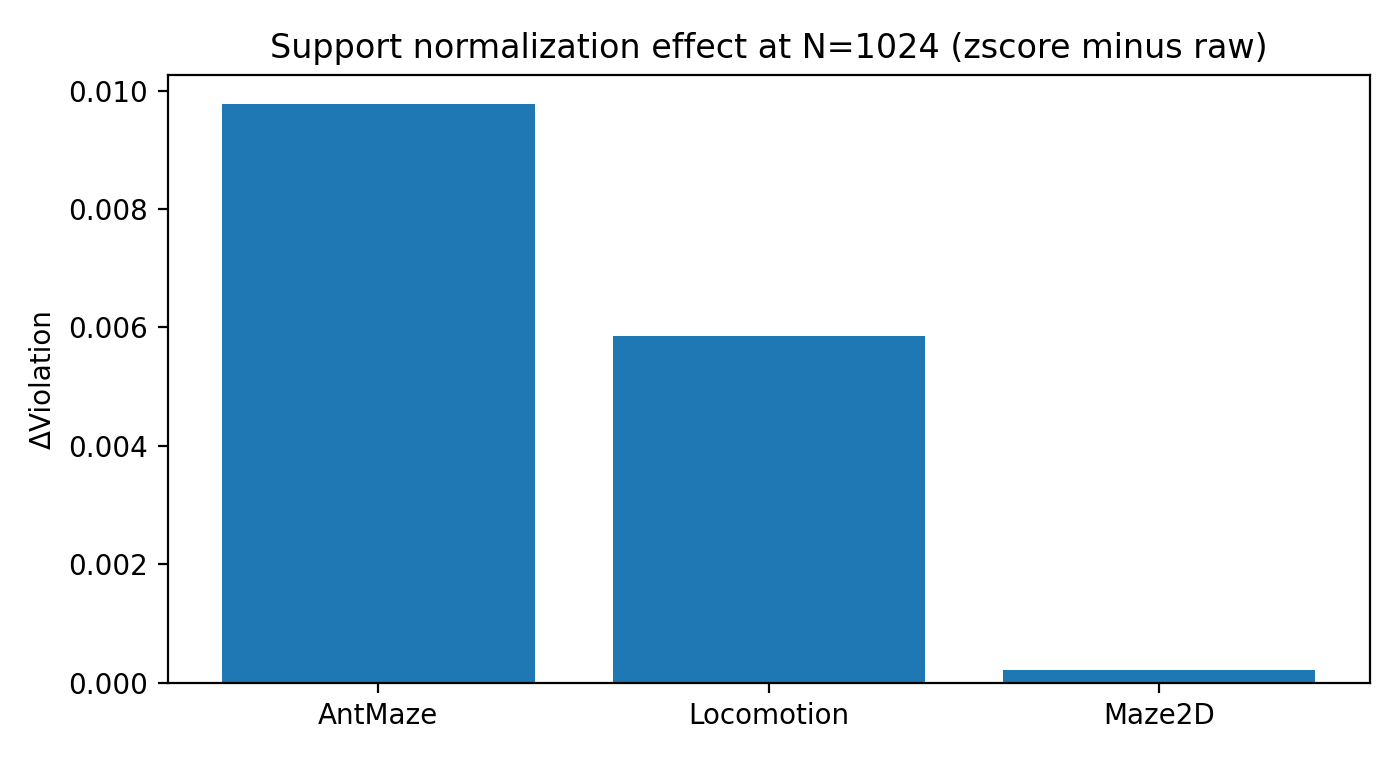}\\
  \includegraphics[width=0.3\linewidth]{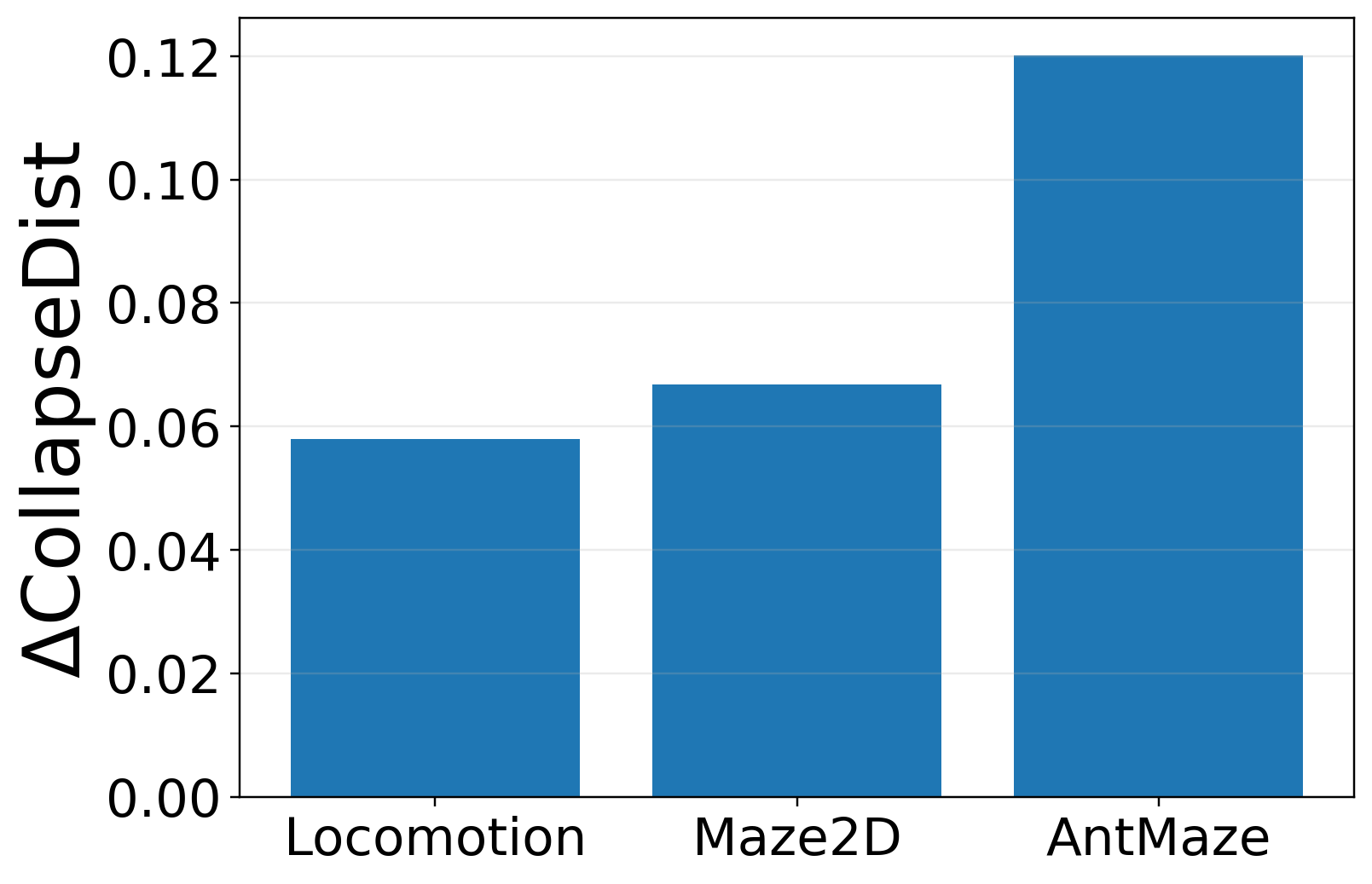}
  \caption{\textbf{Effect of support normalization at $N{=}1024$ (Mode=\texttt{zscore} minus Mode=\texttt{raw}).}
  Top: $\Delta$Score (positive is better). Middle: $\Delta$\textsc{Violation}. Bottom: $\Delta$\textsc{CollapseDist}.
  This comparison isolates scale sensitivity: raw $\log\mu_\varphi(a\mid s)$ can vary substantially across states, so its additive influence can make a single global $w_p$ behave inconsistently across the state space.}
  \label{fig:app_mode_delta_stack}
\end{figure}

\paragraph{$w_p$ sweep frontier (risk--return knob).}
To visualize knob semantics directly, we sweep $w_p$ \emph{post-training} with all networks fixed.
For each suite we fix $N{=}1024$ and the suite-default pessimism $\lambda$ (Section~\ref{sec:results_main}),
then vary the terminal support weight $w_p^{\mathrm{end}}$ in the episode schedule while keeping the schedule form unchanged.
For each setting we report suite-mean normalized score and \textsc{Violation}.
The expected pattern under calibrated support is a monotone risk--return tradeoff:
increasing $w_p$ pushes selection toward relatively higher-support candidates within $\mathcal{C}(s)$, typically lowering \textsc{Violation} at some cost in score,
without retraining.

\begin{figure}[t]
  \centering
  \includegraphics[width=\linewidth]{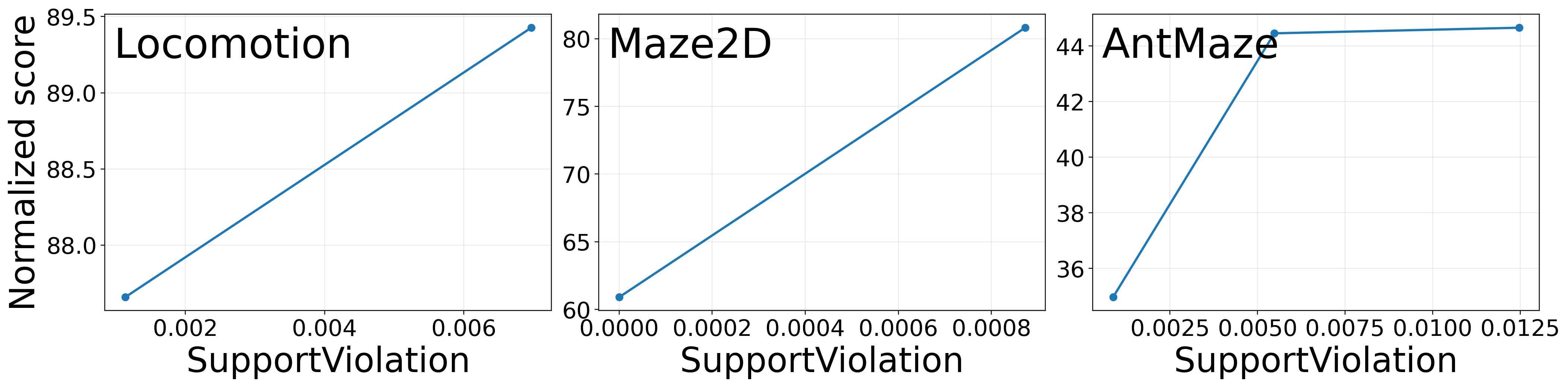}
  \caption{\textbf{$w_p$ sweep frontier at $N{=}1024$ (suite level): Score vs \textsc{Violation}.}
  Each point corresponds to a post-training inference-only choice of $w_p$ (with fixed trained networks).
  The curve visualizes the intended deployment contract: $w_p$ acts as an interpretable exchange rate between conservative value and \emph{relative} behavior support within the current candidate set, enabling retrain-free risk control under candidate maximization.}
  \label{fig:app_pw_frontier_suite}
\end{figure}

\subsection{Full deployment compute measurements}
\label{app:tradeoff_full}

Table~\ref{tab:tradeoff_full} reports the complete latency and memory measurements for all logged methods and sweep settings under the shared harness used for Figure~\ref{fig:master_tradeoff}.
We report mean$\pm$std across repeated measurements.

\begin{table*}[t]
\centering
\caption{Full deployment compute measurements (mean$\pm$std). Latency is in milliseconds; memory is peak allocated/reserved MB; Params is model parameter count (millions).}
\label{tab:tradeoff_full}
\scriptsize
\setlength{\tabcolsep}{4pt}
\begin{tabular}{l l c c c c}
\hline
Method & Setting & Latency (ms) & Peak alloc (MB) & Peak reserved (MB) & Params (M) \\
\hline
Decision Transformer & N1 & 0.88$\pm$0.05 & 20.71$\pm$0.00 & 25.00$\pm$0.00 & 5.27 \\
Decision Transformer & N64 & 0.88$\pm$0.05 & 20.71$\pm$0.00 & 25.00$\pm$0.00 & 5.27 \\
Decision Transformer & N1024 & 0.88$\pm$0.05 & 20.71$\pm$0.00 & 25.00$\pm$0.00 & 5.27 \\
Decision Transformer & N2048 & 0.88$\pm$0.05 & 20.71$\pm$0.00 & 25.00$\pm$0.00 & 5.27 \\
Decision Transformer & T10 & 0.88$\pm$0.05 & 20.71$\pm$0.00 & 25.00$\pm$0.00 & 5.27 \\
Decision Transformer & T20 & 0.88$\pm$0.05 & 20.71$\pm$0.00 & 25.00$\pm$0.00 & 5.27 \\
Decision Transformer & T50 & 0.88$\pm$0.05 & 20.71$\pm$0.00 & 25.00$\pm$0.00 & 5.27 \\
Decision Transformer & T100 & 0.88$\pm$0.05 & 20.71$\pm$0.00 & 25.00$\pm$0.00 & 5.27 \\
Diffusion-QL & N1 & 1.84$\pm$0.08 & 20.71$\pm$0.00 & 25.00$\pm$0.00 & 2.16 \\
Diffusion-QL & N64 & 1.84$\pm$0.08 & 20.71$\pm$0.00 & 25.00$\pm$0.00 & 2.16 \\
Diffusion-QL & N1024 & 1.84$\pm$0.08 & 20.71$\pm$0.00 & 25.00$\pm$0.00 & 2.16 \\
Diffusion-QL & N2048 & 1.84$\pm$0.08 & 20.71$\pm$0.00 & 25.00$\pm$0.00 & 2.16 \\
Diffusion-QL & T10 & 1.92$\pm$0.11 & 20.71$\pm$0.00 & 25.00$\pm$0.00 & 2.16 \\
Diffusion-QL & T20 & 3.70$\pm$0.18 & 20.71$\pm$0.00 & 25.00$\pm$0.00 & 2.16 \\
Diffusion-QL & T50 & 9.17$\pm$0.33 & 20.71$\pm$0.00 & 25.00$\pm$0.00 & 2.16 \\
Diffusion-QL & T100 & 18.33$\pm$0.55 & 20.71$\pm$0.00 & 25.00$\pm$0.00 & 2.16 \\
GEM & N1 & 1.25$\pm$0.05 & 20.76$\pm$0.04 & 25.00$\pm$0.00 & 2.38 \\
GEM & N64 & 1.26$\pm$0.05 & 21.91$\pm$0.05 & 25.00$\pm$0.00 & 2.38 \\
GEM & N1024 & 1.27$\pm$0.05 & 42.35$\pm$0.12 & 47.00$\pm$0.00 & 2.38 \\
GEM & N2048 & 1.33$\pm$0.06 & 80.83$\pm$0.22 & 89.00$\pm$0.00 & 2.38 \\
GEM & T10 & 1.25$\pm$0.05 & 20.76$\pm$0.04 & 25.00$\pm$0.00 & 2.38 \\
GEM & T20 & 1.25$\pm$0.05 & 20.76$\pm$0.04 & 25.00$\pm$0.00 & 2.38 \\
GEM & T50 & 1.25$\pm$0.05 & 20.76$\pm$0.04 & 25.00$\pm$0.00 & 2.38 \\
GEM & T100 & 1.25$\pm$0.05 & 20.76$\pm$0.04 & 25.00$\pm$0.00 & 2.38 \\
IDQL & N1 & 10.17$\pm$0.21 & 20.71$\pm$0.00 & 25.00$\pm$0.00 & 2.16 \\
IDQL & N64 & 10.11$\pm$0.20 & 20.71$\pm$0.00 & 25.00$\pm$0.00 & 2.16 \\
IDQL & N1024 & 10.18$\pm$0.20 & 20.71$\pm$0.00 & 25.00$\pm$0.00 & 2.16 \\
IDQL & N2048 & 10.12$\pm$0.19 & 20.71$\pm$0.00 & 25.00$\pm$0.00 & 2.16 \\
IDQL & T10 & 2.04$\pm$0.09 & 20.71$\pm$0.00 & 25.00$\pm$0.00 & 2.16 \\
IDQL & T20 & 4.11$\pm$0.14 & 20.71$\pm$0.00 & 25.00$\pm$0.00 & 2.16 \\
IDQL & T50 & 10.10$\pm$0.23 & 20.71$\pm$0.00 & 25.00$\pm$0.00 & 2.16 \\
IDQL & T100 & 20.37$\pm$0.34 & 20.71$\pm$0.00 & 25.00$\pm$0.00 & 2.16 \\
IQL & N1 & 0.88$\pm$0.05 & 20.71$\pm$0.00 & 25.00$\pm$0.00 & 2.16 \\
IQL & N64 & 0.88$\pm$0.05 & 20.71$\pm$0.00 & 25.00$\pm$0.00 & 2.16 \\
IQL & N1024 & 0.88$\pm$0.05 & 20.71$\pm$0.00 & 25.00$\pm$0.00 & 2.16 \\
IQL & N2048 & 0.88$\pm$0.05 & 20.71$\pm$0.00 & 25.00$\pm$0.00 & 2.16 \\
IQL & T10 & 0.88$\pm$0.05 & 20.71$\pm$0.00 & 25.00$\pm$0.00 & 2.16 \\
IQL & T20 & 0.88$\pm$0.05 & 20.71$\pm$0.00 & 25.00$\pm$0.00 & 2.16 \\
IQL & T50 & 0.88$\pm$0.05 & 20.71$\pm$0.00 & 25.00$\pm$0.00 & 2.16 \\
IQL & T100 & 0.88$\pm$0.05 & 20.71$\pm$0.00 & 25.00$\pm$0.00 & 2.16 \\
\hline
\end{tabular}
\end{table*}

% ============================================================
% Appendix B
% ============================================================
\section{Inference-Time Scoring: Candidate-Set Normalization and Conservative Selection}
\label{app:score}

This appendix makes the inference-time interface fully explicit and justifies the two terms in
Eq.~\eqref{eq:score} through (i) scale-stable support control and (ii) budget-robust conservatism.

\subsection{Exact candidate construction used at deployment}
\label{app:score:candidates}

For a queried state $s$, GEM constructs a candidate set of size $|\mathcal{C}(s)| = N+1$:
\begin{equation}
\label{eq:appB_candset}
\mathcal{C}(s) = \{a_0, a_1, \ldots, a_N\}.
\end{equation}

\paragraph{Anchor action.}
The anchor is deterministic and extracted from the actor mixture weights:
\begin{equation}
\label{eq:appB_anchor}
k^\star(s) \leftarrow \arg\max_{k\in\{1,\dots,K\}} w_k(s),
\qquad
a_0 \leftarrow \mu_{k^\star}(s).
\end{equation}
This matches the implementation that selects the mean of the most probable component under the actor's gating distribution (deterministic mode).

\paragraph{Single-source sampling (no mixing, no fallback).}
The remaining $N$ candidates are sampled from \emph{exactly one} source:
\begin{equation}
\label{eq:appB_source}
a_{1:N} \sim
\begin{cases}
\pi_\theta(\cdot\mid s), & \text{if \texttt{use\_gem\_candidates} is True},\\
\mu_\varphi(\cdot\mid s), & \text{otherwise}.
\end{cases}
\end{equation}
There is no mixture of sources and no auxiliary fallback gate. Importantly, regardless of the sampling source,
\emph{support is always evaluated only under} the behavior model $\mu_\varphi$.

\subsection{Conservative value via an ensemble lower confidence bound}
\label{app:score:lcb}

Let $\{Q_i(s,a)\}_{i=1}^M$ be an ensemble of critics. Define the sample mean and (population) standard deviation:
\begin{equation}
\label{eq:appB_meanstd}
\bar Q(s,a) \triangleq \frac{1}{M}\sum_{i=1}^M Q_i(s,a),
\qquad
\mathrm{Std}(Q_i(s,a)) \triangleq \sqrt{\frac{1}{M}\sum_{i=1}^M (Q_i(s,a)-\bar Q(s,a))^2 }.
\end{equation}
The pessimistic statistic used by GEM is
\begin{equation}
\label{eq:appB_lcb}
\mathrm{LCB}_\lambda(s,a) \triangleq \bar Q(s,a) - \lambda\, \mathrm{Std}(Q_i(s,a)),
\end{equation}
matching Eq.~\eqref{eq:lcb}.

\paragraph{Why LCB is the right \emph{interface-level} control.}
Candidate-based deployment executes
\(
\arg\max_{a\in\mathcal{C}(s)} \mathrm{Score}(s,a).
\)
Any estimation noise in scores is \emph{magnified} by maximization, and this amplification grows with $|\mathcal{C}(s)|$.
LCB explicitly downweights candidates with high ensemble disagreement, reducing the chance that a larger candidate budget
systematically selects high-uncertainty outliers.

\subsection{Behavior support and the exact candidate-set standardization}
\label{app:score:zscore}

\paragraph{Raw support.}
For each candidate $a\in\mathcal{C}(s)$, compute
\begin{equation}
\label{eq:appB_rawsupport}
\ell(a) \triangleq \log \mu_\varphi(a\mid s).
\end{equation}

\paragraph{Candidate-set normalization (exactly as implemented).}
Let $\bm{\ell} \in \mathbb{R}^{|\mathcal{C}(s)|}$ be the vector of log-likelihoods over the current candidate set:
\(
\bm{\ell} = [\ell(a_0), \ell(a_1), \dots, \ell(a_N)]^\top.
\)
Define
\begin{equation}
\label{eq:appB_transform_pw}
\mu_\ell \triangleq \frac{1}{|\mathcal{C}(s)|}\sum_{a\in\mathcal{C}(s)} \ell(a),
\qquad
\sigma_\ell \triangleq \sqrt{\frac{1}{|\mathcal{C}(s)|}\sum_{a\in\mathcal{C}(s)} (\ell(a)-\mu_\ell)^2},
\qquad
\tilde \ell(a) \triangleq \frac{\ell(a)-\mu_\ell}{\max(\sigma_\ell,\,10^{-6})}.
\end{equation}
This is the exact normalization performed by \texttt{transform\_pw\_term} in the code: mean over the candidate vector,
standard deviation with \texttt{unbiased=False}, and \texttt{clamp\_min(1e-6)}.

\paragraph{Stable semantics of $w_p$.}
After normalization, the support term is measured in ``candidate-set standard deviations'':
\(
\tilde\ell(a) = +1
\)
means ``one standard deviation above the candidate-set mean support'' for this state.
Thus the scalar $w_p$ becomes a state-robust exchange rate between conservative value units and standardized support units.

\subsubsection{Invariance properties}
\label{app:score:invariance}

Candidate-set standardization yields two invariances that are essential for interface control.

\paragraph{Shift invariance.}
For any constant $c\in\mathbb{R}$, define $\ell'(a)=\ell(a)+c$.
Then $\mu_{\ell'}=\mu_\ell+c$ and $\sigma_{\ell'}=\sigma_\ell$, so
\begin{equation}
\label{eq:appB_shiftinv}
\tilde\ell'(a)=\frac{\ell(a)+c-(\mu_\ell+c)}{\sigma_\ell}=\tilde\ell(a).
\end{equation}

\paragraph{Scale invariance.}
For any $\alpha>0$, define $\ell''(a)=\alpha \ell(a)$.
Then $\mu_{\ell''}=\alpha\mu_\ell$ and $\sigma_{\ell''}=\alpha\sigma_\ell$, so
\begin{equation}
\label{eq:appB_scaleinv}
\tilde\ell''(a)=\frac{\alpha\ell(a)-\alpha\mu_\ell}{\alpha\sigma_\ell}=\tilde\ell(a).
\end{equation}

\paragraph{Implication.}
Even if $\log\mu_\varphi(a\mid s)$ varies dramatically in absolute scale across states, datasets, or candidate budgets,
the normalized support $\tilde\ell(a)$ preserves relative support ordering within $\mathcal{C}(s)$ and makes $w_p$
comparably meaningful across states.

\subsection{The fixed-form (``locked'') inference score}
\label{app:score:locked}

Given the two quantities above, GEM evaluates each candidate via a fixed-form score:
\begin{equation}
\label{eq:appB_score}
\mathrm{Score}(s,a)
=
\mathrm{LCB}_\lambda(s,a)
+
w_p \cdot \tilde\ell(a),
\qquad
\tilde\ell(a)=\mathrm{zscore}_{\mathcal{C}(s)}\big(\log\mu_\varphi(a\mid s)\big),
\end{equation}
which is exactly Eq.~\eqref{eq:score}.

\paragraph{Meaning of ``locked.''}
We use ``locked'' to mean: once Phase~I finishes, the \emph{functional form} of the inference rule is fixed and does not depend
on additional learned selection networks or state-dependent mixing heuristics. Deployment-time control is entirely through
a small set of explicit scalar knobs: $N$ (candidate budget), $\lambda$ (LCB pessimism), $w_p$ (support weight),
and the optional $k_{\mathrm{smooth}}$ (output smoothing).

\paragraph{Top-$k$ smoothing operator.}
After computing scores over $\mathcal{C}(s)$, the interface outputs
\begin{equation}
\label{eq:appB_topk}
\hat a(s)
=
\frac{1}{k}\sum_{j\in \mathrm{TopK}(\mathrm{Score})} a_j,
\qquad
k = \min(k_{\mathrm{smooth}}, |\mathcal{C}(s)|),
\end{equation}
where $k_{\mathrm{smooth}}=1$ recovers the top-1 action.

\subsection{Why candidate maximization imposes an additional constraint (extreme-value amplification)}
\label{app:score:extremevalue}

This subsection formalizes the interface-level issue that motivates combining support control with conservatism.

\paragraph{Setup.}
Fix a state $s$ and write a generic score estimate as
\begin{equation}
\label{eq:appB_noise}
\widehat{\mathrm{Score}}(s,a) = \mathrm{Score}^\star(s,a) + \varepsilon(s,a),
\end{equation}
where $\varepsilon$ captures approximation error, extrapolation error, and/or stochastic estimation noise.

Candidate-based deployment selects
\begin{equation}
\label{eq:appB_argmax}
\hat a \in \arg\max_{a\in\mathcal{C}(s)} \widehat{\mathrm{Score}}(s,a).
\end{equation}
Even if $\varepsilon$ is mean-zero, the maximization operator introduces a positive bias toward extreme noise values,
and this bias typically grows with $|\mathcal{C}(s)|$.

\paragraph{A standard bound for sub-Gaussian noise.}
Assume for simplicity that $\{\varepsilon(s,a_j)\}_{j=0}^N$ are independent and $\sigma$-sub-Gaussian:
\(
\mathbb{E}[\exp(t\varepsilon)] \le \exp(\sigma^2 t^2/2)
\)
for all $t\in\mathbb{R}$.
Then
\begin{align}
\mathbb{E}\Big[\max_{0\le j\le N} \varepsilon(s,a_j)\Big]
&=
\frac{1}{t}\,
\mathbb{E}\Big[\log \exp\big(t\max_j \varepsilon_j\big)\Big] \\
&\le
\frac{1}{t}\,
\mathbb{E}\Big[\log \sum_{j=0}^N \exp(t \varepsilon_j)\Big] \\
&\le
\frac{1}{t}\,
\log \sum_{j=0}^N \mathbb{E}[\exp(t\varepsilon_j)] \\
&\le
\frac{1}{t}\,
\log\Big((N+1)\exp(\sigma^2 t^2/2)\Big) \\
&=
\frac{\log(N+1)}{t} + \frac{\sigma^2 t}{2}.
\end{align}
Optimizing over $t>0$ gives $t^\star = \sqrt{2\log(N+1)}/\sigma$ and thus
\begin{equation}
\label{eq:appB_maxnoise}
\mathbb{E}\Big[\max_{0\le j\le N} \varepsilon(s,a_j)\Big]
\le
\sigma \sqrt{2\log(N+1)}.
\end{equation}

\paragraph{Interface implication.}
Increasing $N$ increases the expected maximum noise term at least on the order of $\sqrt{\log N}$ under broad conditions.
In offline RL, candidates with unusually high estimated value often coincide with weak support or high uncertainty.
Therefore, a scalable candidate interface must include explicit mechanisms that counteract this amplification.
In GEM, $\tilde\ell(a)$ penalizes weak support in a scale-stable way, while $\mathrm{LCB}_\lambda$ penalizes disagreement-driven outliers.

\begin{table}[t]
\centering
\caption{GEM primary performance (Mode=zscore, $N=1024$) aggregated by D4RL task. Columns report normalized score (mean $\pm$ std across training seeds), average support z-score, support violation rate, collapse distance, and evaluation throughput.}
\label{tab:gem-main-by-env}
{\small
\begin{tabular}{lcccccc}
\toprule
TestEnv & GEM & n & SupportZ & Viol & Collapse & SPS \\
\midrule
antmaze-large-diverse-v2 & 1.00 $\pm$ 1.41 & 3 & 1.99 & 0.0019 & 0.308 & 2742 \\
antmaze-large-play-v2 & 5.00 $\pm$ 1.41 & 3 & 2.02 & 0.0024 & 0.318 & 2760 \\
antmaze-medium-diverse-v2 & 11.76 $\pm$ 10.54 & 3 & 1.86 & 0.0031 & 0.357 & 2913 \\
antmaze-medium-play-v2 & 76.36 $\pm$ 33.43 & 3 & 1.05 & 0.0356 & 0.748 & 2962 \\
antmaze-umaze-diverse-v2 & 57.37 $\pm$ 5.55 & 3 & 2.03 & 0.0002 & 0.278 & 3173 \\
antmaze-umaze-v2 & 75.71 $\pm$ 20.55 & 3 & 0.81 & 0.0189 & 0.309 & 3195 \\
halfcheetah-medium-expert-v2 & 100.69 $\pm$ 0.57 & 3 & 1.19 & 0.0038 & 0.183 & 3282 \\
halfcheetah-medium-replay-v2 & 50.03 $\pm$ 0.90 & 3 & 1.31 & 0.0000 & 0.264 & 3365 \\
halfcheetah-medium-v2 & 42.67 $\pm$ 0.36 & 3 & 1.68 & 0.0000 & 0.173 & 3328 \\
hopper-medium-expert-v2 & 111.73 $\pm$ 0.45 & 3 & 1.30 & 0.0015 & 0.097 & 3182 \\
hopper-medium-replay-v2 & 100.36 $\pm$ 1.97 & 3 & 0.87 & 0.0027 & 0.198 & 3147 \\
hopper-medium-v2 & 74.89 $\pm$ 0.54 & 3 & 1.81 & 0.0000 & 0.168 & 3126 \\
maze2d-large-v1 & 68.66 $\pm$ 45.58 & 3 & 1.47 & 0.0048 & 0.201 & 3294 \\
maze2d-medium-v1 & 104.44 $\pm$ 39.24 & 3 & 1.47 & 0.0014 & 0.182 & 3299 \\
maze2d-umaze-v1 & 154.13 $\pm$ 21.42 & 3 & 0.67 & 0.0000 & 0.260 & 3218 \\
walker2d-medium-expert-v2 & 109.84 $\pm$ 0.07 & 3 & 1.54 & 0.0000 & 0.148 & 3062 \\
walker2d-medium-replay-v2 & 87.01 $\pm$ 2.61 & 3 & 1.09 & 0.0055 & 0.197 & 3120 \\
walker2d-medium-v2 & 78.53 $\pm$ 0.45 & 3 & 1.39 & 0.0000 & 0.153 & 3035 \\
\bottomrule
\end{tabular}
}
\end{table}

\begin{table}[t]
\centering
\caption{Mechanism ablations at $N=1024$ (Mode=zscore), aggregated over all tasks and training seeds. $\Delta$ is the per-task score difference relative to GEM (mean $\pm$ std across tasks).}
\label{tab:ablation-summary}
{\small
\begin{tabular}{lccccccc}
\toprule
Variant & Mean & Delta & SupportZ & Viol & Collapse & SPS & n \\
\midrule
GEM & 76.88 & 0.00 $\pm$ 0.00 & 1.25 & 0.0069 & 0.273 & 3064 & 47 \\
anchor\_bary & 76.29 & -0.60 $\pm$ 3.87 & 1.24 & 0.0070 & 0.271 & 3043 & 47 \\
behavior\_only\_cands & 79.92 & 3.03 $\pm$ 20.58 & 0.93 & 0.0144 & 0.240 & 3077 & 47 \\
no\_anchor & 77.04 & 0.15 $\pm$ 2.32 & 1.24 & 0.0069 & 0.273 & 3109 & 47 \\
no\_lcb & 74.84 & -2.04 $\pm$ 16.02 & 1.27 & 0.0085 & 0.270 & 3093 & 47 \\
no\_pw & 65.31 & -11.57 $\pm$ 37.56 & -0.90 & 0.2169 & 0.647 & 3082 & 47 \\
unimodal\_actor\_cands & 75.70 & -1.19 $\pm$ 10.30 & 1.64 & 0.0027 & 0.269 & 3055 & 47 \\
\bottomrule
\end{tabular}
}
\end{table}

\begin{table}[t]
\centering
\caption{Per-task ablation deltas at $N=1024$ (Mode=zscore). Columns report mean normalized score of GEM and mean score change ($\Delta$) of each variant relative to GEM (positive is better).}
\label{tab:ablation-by-env-delta}
{\small
\begin{tabular}{lccccc}
\toprule
TestEnv & GEM & no-pw & no-lcb & unimodal-actor & behavior-only \\
\midrule
antmaze-large-diverse-v2 & 0.00 & +1.00 & +0.00 & +8.00 & +0.00 \\
antmaze-large-play-v2 & 4.00 & +0.00 & -1.00 & -2.00 & -4.00 \\
antmaze-medium-diverse-v2 & 6.00 & +68.14 & +0.00 & -4.00 & +2.00 \\
antmaze-medium-play-v2 & 76.47 & -56.47 & -37.05 & +23.53 & -19.80 \\
antmaze-umaze-diverse-v2 & 52.42 & -31.42 & -9.06 & -11.93 & -16.23 \\
antmaze-umaze-v2 & 71.58 & -68.58 & -5.00 & +0.00 & -6.00 \\
halfcheetah-medium-expert-v2 & 100.55 & +0.22 & -0.33 & -0.77 & +0.32 \\
halfcheetah-medium-replay-v2 & 50.11 & -0.11 & -0.29 & +0.09 & -0.05 \\
halfcheetah-medium-v2 & 42.67 & +0.33 & +0.42 & +0.58 & +0.59 \\
hopper-medium-expert-v2 & 111.73 & -0.39 & -0.86 & -0.50 & -0.42 \\
hopper-medium-replay-v2 & 100.36 & -0.55 & -2.14 & -1.92 & -0.61 \\
hopper-medium-v2 & 74.89 & +0.22 & +0.31 & -0.34 & +0.16 \\
maze2d-large-v1 & 68.66 & -68.66 & -8.69 & -0.02 & +38.54 \\
maze2d-medium-v1 & 104.44 & -98.08 & -3.97 & -4.04 & +6.81 \\
maze2d-umaze-v1 & 154.13 & -154.13 & -3.47 & -8.71 & +4.74 \\
walker2d-medium-expert-v2 & 109.84 & -0.46 & -0.25 & -0.59 & -0.54 \\
walker2d-medium-replay-v2 & 87.01 & -1.95 & -1.68 & -1.56 & +0.19 \\
walker2d-medium-v2 & 78.53 & -0.75 & -1.03 & -1.08 & -0.12 \\
\bottomrule
\end{tabular}
}
\end{table}

\begin{table}[t]
\centering
\caption{Test-time scaling with candidate budget $N$ for GEM (Mode=zscore), aggregated over all sweep runs. Larger $N$ improves average score while reducing throughput.}
\label{tab:N-scaling-summary}
{\small
\begin{tabular}{rccccccc}
\toprule
N & Score & Std & SupportZ & Viol & Collapse & Steps/s & n \\
\midrule
0 & 64.24 & 30.99 & 0.00 & 0.0000 & 0.265 & 5625 & 846 \\
1 & 65.77 & 30.53 & 0.92 & 0.0000 & 0.271 & 5254 & 846 \\
64 & 72.32 & 34.25 & 1.26 & 0.0027 & 0.257 & 5383 & 846 \\
1024 & 72.79 & 34.46 & 1.33 & 0.0048 & 0.237 & 3105 & 846 \\
2048 & 73.41 & 35.00 & 1.34 & 0.0055 & 0.233 & 1958 & 846 \\
\bottomrule
\end{tabular}
}
\end{table}

\begin{table}[t]
\centering
\caption{Effect of support normalization: baseline GEM difference between Mode=zscore and Mode=raw (reported as $\mathrm{Score}_{z}-\mathrm{Score}_{raw}$) aggregated by task.}
\label{tab:mode-delta-by-env}
{\small
\begin{tabular}{lcc}
\toprule
TestEnv & Delta & n \\
\midrule
antmaze-large-diverse-v2 & 0.00 $\pm$ 0.00 & 2 \\
antmaze-large-play-v2 & 5.00 $\pm$ 1.41 & 2 \\
antmaze-medium-diverse-v2 & 9.76 & 1 \\
antmaze-medium-play-v2 & 19.89 $\pm$ 25.51 & 2 \\
antmaze-umaze-diverse-v2 & 3.81 $\pm$ 19.59 & 2 \\
antmaze-umaze-v2 & -1.12 $\pm$ 6.91 & 2 \\
halfcheetah-medium-expert-v2 & 0.35 $\pm$ 0.42 & 3 \\
halfcheetah-medium-replay-v2 & 0.08 $\pm$ 0.12 & 3 \\
halfcheetah-medium-v2 & -0.14 $\pm$ 0.09 & 3 \\
hopper-medium-expert-v2 & 0.09 $\pm$ 0.06 & 3 \\
hopper-medium-replay-v2 & 1.45 $\pm$ 0.59 & 3 \\
hopper-medium-v2 & 0.02 $\pm$ 0.19 & 3 \\
maze2d-large-v1 & 10.56 $\pm$ 6.06 & 3 \\
maze2d-medium-v1 & 47.69 $\pm$ 38.87 & 3 \\
maze2d-umaze-v1 & 24.60 $\pm$ 17.96 & 3 \\
walker2d-medium-expert-v2 & 0.03 $\pm$ 0.08 & 3 \\
walker2d-medium-replay-v2 & 0.06 $\pm$ 0.10 & 3 \\
walker2d-medium-v2 & 0.01 $\pm$ 0.05 & 3 \\
\bottomrule
\end{tabular}
}
\end{table}

\begin{table}[t]
\centering
\caption{Multimodality diagnostic (dataset NLL gap). Larger gap indicates higher cost of a unimodal projection, motivating multimodal proposals.}
\label{tab:nll-gap}
{\small
\begin{tabular}{lcc}
\toprule
Environment & Gap & Seeds \\
\midrule
antmaze-large-diverse-v2 & 42.71 $\pm$ 7.04 & 3 \\
antmaze-large-play-v2 & 43.22 $\pm$ 15.75 & 3 \\
antmaze-medium-diverse-v2 & 57.16 $\pm$ 9.94 & 3 \\
antmaze-medium-play-v2 & 48.45 $\pm$ 13.97 & 3 \\
antmaze-umaze-diverse-v2 & 48.42 $\pm$ 7.08 & 3 \\
antmaze-umaze-v2 & 48.86 $\pm$ 7.26 & 3 \\
halfcheetah-medium-expert-v2 & 5.97 $\pm$ 0.24 & 3 \\
halfcheetah-medium-replay-v2 & 1.37 $\pm$ 0.02 & 3 \\
halfcheetah-medium-v2 & 1.74 $\pm$ 0.13 & 3 \\
hopper-medium-expert-v2 & 2.24 $\pm$ 0.43 & 3 \\
hopper-medium-replay-v2 & 1.43 $\pm$ 0.26 & 3 \\
hopper-medium-v2 & 6.05 $\pm$ 0.07 & 3 \\
maze2d-large-v1 & 6.46 $\pm$ 0.05 & 3 \\
maze2d-medium-v1 & 1.53 $\pm$ 0.00 & 3 \\
maze2d-umaze-v1 & 2.04 $\pm$ 0.01 & 3 \\
walker2d-medium-expert-v2 & 4.88 $\pm$ 0.63 & 3 \\
walker2d-medium-replay-v2 & 2.40 $\pm$ 0.29 & 3 \\
walker2d-medium-v2 & 6.08 $\pm$ 0.30 & 3 \\
\bottomrule
\end{tabular}
}
\end{table}

\begin{table}[t]
\centering
\caption{Inference latency (ms) under diffusion-step sweep $T$ (mean $\pm$ std).}
\label{tab:latency-T}
{\small
\begin{tabular}{lcccc}
\toprule
Method & T10 & T100 & T20 & T50 \\
\midrule
DT & 0.934 $\pm$ 0.122 & 0.866 $\pm$ 0.008 & 0.874 $\pm$ 0.013 & 0.866 $\pm$ 0.010 \\
FQL & 1.479 $\pm$ 0.014 & 1.487 $\pm$ 0.018 & 1.480 $\pm$ 0.012 & 1.474 $\pm$ 0.024 \\
GEM & 1.296 $\pm$ 0.016 & 1.301 $\pm$ 0.043 & 1.270 $\pm$ 0.006 & 1.292 $\pm$ 0.052 \\
IDQL & 2.042 $\pm$ 0.017 & 20.367 $\pm$ 0.163 & 4.112 $\pm$ 0.014 & 10.096 $\pm$ 0.050 \\
IFQL & 1.831 $\pm$ 0.017 & 1.827 $\pm$ 0.010 & 1.821 $\pm$ 0.019 & 1.804 $\pm$ 0.026 \\
\bottomrule
\end{tabular}
}
\end{table}

\begin{table}[t]
\centering
\caption{Inference latency (ms) under candidate-budget sweep $N$ (mean $\pm$ std).}
\label{tab:latency-N}
{\small
\begin{tabular}{lcccc}
\toprule
Method & N1 & N1024 & N2048 & N64 \\
\midrule
DT & 0.886 $\pm$ 0.006 & 0.884 $\pm$ 0.007 & 0.878 $\pm$ 0.006 & 0.865 $\pm$ 0.008 \\
FQL & 1.483 $\pm$ 0.015 & 1.506 $\pm$ 0.014 & 1.465 $\pm$ 0.038 & 1.482 $\pm$ 0.010 \\
GEM & 1.307 $\pm$ 0.007 & 1.276 $\pm$ 0.016 & 1.254 $\pm$ 0.002 & 1.279 $\pm$ 0.010 \\
IDQL & 10.168 $\pm$ 0.046 & 10.175 $\pm$ 0.051 & 10.115 $\pm$ 0.108 & 10.107 $\pm$ 0.072 \\
IFQL & 1.796 $\pm$ 0.008 & 1.837 $\pm$ 0.035 & 1.865 $\pm$ 0.076 & 1.830 $\pm$ 0.012 \\
\bottomrule
\end{tabular}
}
\end{table}

\begin{table}[t]
\centering
\caption{Peak GPU allocated memory (MB) under diffusion-step sweep $T$ (mean $\pm$ std).}
\label{tab:mem-T}
{\small
\begin{tabular}{lcccc}
\toprule
Method & T10 & T100 & T20 & T50 \\
\midrule
DT & 21.1 $\pm$ 0.0 & 21.1 $\pm$ 0.0 & 21.1 $\pm$ 0.0 & 21.1 $\pm$ 0.0 \\
FQL & 20.7 $\pm$ 0.0 & 20.7 $\pm$ 0.0 & 20.7 $\pm$ 0.0 & 20.7 $\pm$ 0.0 \\
GEM & 50.8 $\pm$ 0.0 & 50.8 $\pm$ 0.0 & 50.8 $\pm$ 0.0 & 50.8 $\pm$ 0.0 \\
IDQL & 20.7 $\pm$ 0.0 & 20.7 $\pm$ 0.0 & 20.7 $\pm$ 0.0 & 20.7 $\pm$ 0.0 \\
IFQL & 20.8 $\pm$ 0.0 & 20.8 $\pm$ 0.0 & 20.8 $\pm$ 0.0 & 20.8 $\pm$ 0.0 \\
\bottomrule
\end{tabular}
}
\end{table}

\begin{table}[t]
\centering
\caption{Peak GPU allocated memory (MB) under candidate-budget sweep $N$ (mean $\pm$ std).}
\label{tab:mem-N}
{\small
\begin{tabular}{lcccc}
\toprule
Method & N1 & N1024 & N2048 & N64 \\
\midrule
DT & 21.1 $\pm$ 0.0 & 21.1 $\pm$ 0.0 & 21.1 $\pm$ 0.0 & 21.1 $\pm$ 0.0 \\
FQL & 20.7 $\pm$ 0.0 & 20.7 $\pm$ 0.0 & 20.7 $\pm$ 0.0 & 20.7 $\pm$ 0.0 \\
GEM & 50.8 $\pm$ 0.0 & 50.8 $\pm$ 0.0 & 50.8 $\pm$ 0.0 & 50.8 $\pm$ 0.0 \\
IDQL & 20.7 $\pm$ 0.0 & 20.7 $\pm$ 0.0 & 20.7 $\pm$ 0.0 & 20.7 $\pm$ 0.0 \\
IFQL & 20.8 $\pm$ 0.0 & 20.8 $\pm$ 0.0 & 20.8 $\pm$ 0.0 & 20.8 $\pm$ 0.0 \\
\bottomrule
\end{tabular}
}
\end{table}

% ============================================================
% Appendix C
% ============================================================
\section{Guided EM-Style Actor Update: Variational Derivation and Exact Surrogate}
\label{app:em}

This appendix derives the EM-style surrogate optimized by the actor and clarifies in what precise sense the update is
(i) \emph{EM-style} (infer responsibilities then update parameters) and (ii) \emph{guided} (dataset actions are reweighted by a critic-derived advantage).
We keep the derivation aligned with the implementation: responsibilities are computed by a softmax over component log-joints and
are \emph{detached} by default when taking gradients.

\subsection{Actor as a latent-variable conditional mixture}
\label{app:em:model}

For each state $s$, the actor defines a $K$-component Gaussian mixture:
\begin{equation}
\label{eq:appC_gmm}
\pi_\theta(a\mid s)
=
\sum_{k=1}^K w_{\theta,k}(s)\,
\mathcal{N}\!\big(a;\mu_{\theta,k}(s), \Sigma_{\theta,k}\big),
\qquad
\sum_{k=1}^K w_{\theta,k}(s)=1,
\quad
w_{\theta,k}(s)>0.
\end{equation}
The implementation uses a diagonal covariance with per-component log-std parameters:
\(
\Sigma_{\theta,k} = \mathrm{diag}(\sigma_{\theta,k}^2)
\)
where $\sigma_{\theta,k}$ does not depend on $s$ (a learned parameter per component and action dimension).

Introduce a discrete latent variable $z\in\{1,\dots,K\}$:
\begin{equation}
\label{eq:appC_latent}
p_\theta(z=k\mid s) = w_{\theta,k}(s),
\qquad
p_\theta(a\mid s,z=k)=\mathcal{N}\!\big(a;\mu_{\theta,k}(s),\Sigma_{\theta,k}\big).
\end{equation}
Then
\(
\pi_\theta(a\mid s)=\sum_k p_\theta(z=k\mid s)\,p_\theta(a\mid s,z=k).
\)

\subsection{Exact component log-joints and responsibilities}
\label{app:em:C_resp}

Define the per-component log-joint (up to the conditioning on $s$):
\begin{equation}
\label{eq:appC_logjoint}
u_{\theta,k}(s,a)
\triangleq
\log p_\theta(z=k\mid s) + \log p_\theta(a\mid s,z=k)
=
\log w_{\theta,k}(s) + \log \mathcal{N}\!\big(a;\mu_{\theta,k}(s),\Sigma_{\theta,k}\big).
\end{equation}
Then the exact mixture log-likelihood is
\begin{equation}
\label{eq:appC_loglik}
\log \pi_\theta(a\mid s)
=
\log\sum_{k=1}^K \exp\big(u_{\theta,k}(s,a)\big).
\end{equation}

\paragraph{Responsibilities (posterior over components).}
The exact posterior is a softmax over log-joints:
\begin{equation}
\label{eq:appC_resp_softmax}
\gamma_{\theta,k}(s,a)
\triangleq
p_\theta(z=k\mid s,a)
=
\frac{\exp(u_{\theta,k}(s,a))}{\sum_{j=1}^K \exp(u_{\theta,j}(s,a))}
=
\mathrm{softmax}_k\big(u_{\theta,k}(s,a)\big).
\end{equation}
This matches the implementation:
\(
\log w_{\theta,k}(s)
\)
is obtained via \texttt{log\_softmax} of logits, and
\(
\log \mathcal{N}(\cdot)
\)
is computed in closed form (diagonal Gaussian).

\subsection{Variational view: Jensen lower bound and its tightness}
\label{app:em:variational}

The ``EM'' interpretation can be made fully explicit via a standard variational bound.

Let $q(z\mid s,a)$ be any categorical distribution over $\{1,\dots,K\}$ with probabilities $\{q_k\}_{k=1}^K$.
Then
\begin{align}
\log \pi_\theta(a\mid s)
&= \log \sum_{k=1}^K \exp(u_{\theta,k}(s,a)) \\
&= \log \sum_{k=1}^K q_k \frac{\exp(u_{\theta,k}(s,a))}{q_k} \\
&\ge \sum_{k=1}^K q_k \log \frac{\exp(u_{\theta,k}(s,a))}{q_k}
\qquad \text{(Jensen)} \\
&= \sum_{k=1}^K q_k\,u_{\theta,k}(s,a) + \underbrace{\Big(-\sum_{k=1}^K q_k\log q_k\Big)}_{H(q)}.
\end{align}
Define the evidence lower bound (ELBO):
\begin{equation}
\label{eq:appC_elbo}
\mathcal{L}_{\mathrm{ELBO}}(\theta; q)
\triangleq
\sum_{k=1}^K q_k\,u_{\theta,k}(s,a) + H(q).
\end{equation}

\paragraph{Tightness condition.}
The bound is tight when $q(z\mid s,a)=p_\theta(z\mid s,a)$, i.e., $q_k=\gamma_{\theta,k}(s,a)$.
Plugging $q=\gamma_\theta$ into Eq.~\eqref{eq:appC_elbo} yields
\begin{equation}
\label{eq:appC_tight}
\log \pi_\theta(a\mid s)
=
\sum_{k=1}^K \gamma_{\theta,k}(s,a)\,u_{\theta,k}(s,a)
+
H(\gamma_\theta(\cdot\mid s,a)).
\end{equation}

\subsection{EM Q-function and the exact surrogate used in GEM}
\label{app:em:qfunction}

Classical EM alternates:
\begin{itemize}
\item \textbf{E-step:} set responsibilities $\gamma^{\text{old}}(s,a)\leftarrow p_{\theta^{\text{old}}}(z\mid s,a)$;
\item \textbf{M-step:} update $\theta$ by maximizing the expected complete-data log-likelihood under $\gamma^{\text{old}}$.
\end{itemize}

The M-step objective is the \emph{Q-function}:
\begin{equation}
\label{eq:appC_Q}
Q(\theta;\theta^{\text{old}})
\triangleq
\mathbb{E}_{(s,a)\sim\mathcal{D}}
\Big[
\sum_{k=1}^K \gamma^{\text{old}}_{k}(s,a)\,
u_{\theta,k}(s,a)
\Big].
\end{equation}
Note that the entropy term $H(\gamma^{\text{old}})$ does not depend on $\theta$ in the M-step and can be omitted without changing the optimizer.

\paragraph{Exact match to the implementation (``loose ELBO'').}
The actor implementation computes
\begin{equation}
\label{eq:appC_impl_elbo}
\mathrm{elbo\_log\_prob}_\theta(s,a)
=
\sum_{k=1}^K \gamma_{\theta,k}(s,a)\,
\Big(
\log w_{\theta,k}(s) + \log \mathcal{N}(a;\mu_{\theta,k}(s),\Sigma_{\theta,k})
\Big),
\end{equation}
i.e., precisely the Q-function form with responsibilities $\gamma_\theta$.
Crucially, by default \texttt{detach\_gamma=True}, so during backpropagation the gradient treats $\gamma$ as a constant:
\begin{equation}
\label{eq:appC_detach}
\nabla_\theta \,\mathrm{elbo\_log\_prob}_\theta(s,a)
=
\sum_{k=1}^K \gamma_{\theta,k}(s,a)\,
\nabla_\theta
\Big(
\log w_{\theta,k}(s) + \log \mathcal{N}(a;\mu_{\theta,k}(s),\Sigma_{\theta,k})
\Big),
\end{equation}
which is exactly the M-step gradient direction of an EM-style update. This ``infer responsibilities then update parameters''
structure is the operational sense in which the actor update is EM-style.

\paragraph{Why the entropy term is not explicitly included.}
From Eq.~\eqref{eq:appC_tight}, the true log-likelihood decomposition contains $H(\gamma_\theta)$.
However, in EM the responsibilities are treated as fixed in the M-step, making $H(\gamma)$ a constant w.r.t.\ $\theta$.
The implementation follows this M-step logic via \texttt{detach\_gamma=True}, so adding $H(\gamma)$ would not change the gradient.
(If one were to backprop through $\gamma$ by setting \texttt{detach\_gamma=False}, then omitting $H(\gamma)$ would change the gradient and the update would no longer correspond to the standard EM M-step.)

\subsection{Closed-form Gaussian terms used in the actor}
\label{app:em:gaussian}

We spell out the exact Gaussian log-density used to compute $u_{\theta,k}(s,a)$.

Let $a\in\mathbb{R}^A$ and component parameters $(\mu_{\theta,k}(s),\sigma_{\theta,k})$ with diagonal covariance.
Write $\sigma_{\theta,k}\in\mathbb{R}^A_{>0}$ and $\Sigma_{\theta,k}=\mathrm{diag}(\sigma_{\theta,k}^2)$.
Then
\begin{align}
\label{eq:appC_gauss}
\log \mathcal{N}(a;\mu_{\theta,k}(s),\Sigma_{\theta,k})
&=
-\frac{1}{2}\sum_{d=1}^A
\Big(
\frac{(a_d-\mu_{\theta,k,d}(s))^2}{\sigma_{\theta,k,d}^2}
+
\log(2\pi)
+
2\log\sigma_{\theta,k,d}
\Big) \\
&=
-\frac{1}{2}\Big(\sum_{d=1}^A \frac{(a_d-\mu_{\theta,k,d}(s))^2}{\sigma_{\theta,k,d}^2}\Big)
-\frac{1}{2}\Big(A\log(2\pi) + \sum_{d=1}^A \log(\sigma_{\theta,k,d}^2)\Big).
\end{align}
The code implements the equivalent form via $\mathrm{var}=\sigma^2$:
\(
-\frac{1}{2} \big((a-\mu)^2/\mathrm{var}\big)\text{ summed over }d
-\frac{1}{2}\big(A\log(2\pi)+\log\mathrm{var}\text{ summed over }d\big).
\)

\subsection{``Guided'' weighting: advantage-reweighted EM-style regression}
\label{app:em:guided}

The ``guided'' part is implemented by reweighting each dataset sample $(s,a)$ by a bounded exponential of a critic-derived advantage.
Let
\begin{equation}
\label{eq:appC_adv}
A(s,a) \triangleq Q^{\mathrm{tgt}}(s,a) - V(s),
\end{equation}
where $Q^{\mathrm{tgt}}$ denotes the target critic used for stability and $V$ is the learned value baseline.
Define the guidance weight
\begin{equation}
\label{eq:appC_weight}
\omega(s,a)
\triangleq
\exp(\beta A(s,a)) \ \ \text{clipped above by a constant } \omega_{\max}.
\end{equation}
(Implementation detail: the advantage is detached before exponentiation; clipping is applied to avoid exploding weights.)

\paragraph{Guided objective optimized by the actor.}
Let
\(
\mathcal{Q}_\theta(s,a) \triangleq \sum_k \gamma_{\theta,k}(s,a)\,u_{\theta,k}(s,a)
\)
be the EM Q-function term in Eq.~\eqref{eq:appC_impl_elbo}.
The code forms the per-sample ``BC loss'' as
\(
-\mathcal{Q}_\theta(s,a)
\)
and minimizes the weighted objective:
\begin{equation}
\label{eq:appC_policy_loss_core}
\mathcal{J}_{\mathrm{core}}(\theta)
=
\mathbb{E}_{(s,a)\sim\mathcal{D}}
\Big[
\omega(s,a)\cdot \big(-\mathcal{Q}_\theta(s,a)\big)
\Big].
\end{equation}
Equivalently, minimizing Eq.~\eqref{eq:appC_policy_loss_core} maximizes
\(
\mathbb{E}[\omega(s,a)\mathcal{Q}_\theta(s,a)].
\)
Thus, compared to standard maximum likelihood (which weights all samples equally), guidance increases the fitting pressure on dataset actions
with higher estimated advantage.

\subsection{Gating entropy regularization (anti-collapse) and its exact form}
\label{app:em:entropy}

To discourage premature collapse of mixture usage, the implementation adds a small entropy bonus on the categorical gating distribution.
Let $\bm{w}_\theta(s) = (w_{\theta,1}(s),\dots,w_{\theta,K}(s))$.
The gating entropy is
\begin{equation}
\label{eq:appC_gate_entropy}
H(\bm{w}_\theta(s))
=
-\sum_{k=1}^K w_{\theta,k}(s)\log\big(w_{\theta,k}(s)+10^{-6}\big).
\end{equation}
The code \emph{subtracts} an entropy bonus from the loss:
\begin{equation}
\label{eq:appC_full_loss}
\mathcal{J}_{\mathrm{actor}}(\theta)
=
\mathbb{E}_{(s,a)\sim\mathcal{D}}
\Big[
\omega(s,a)\cdot \big(-\mathcal{Q}_\theta(s,a)\big)
\Big]
-
\alpha\,\mathbb{E}_{s\sim\mathcal{D}}\big[H(\bm{w}_\theta(s))\big],
\end{equation}
where $\alpha$ is the entropy coefficient. Minimizing Eq.~\eqref{eq:appC_full_loss} corresponds to maximizing
\(
\mathbb{E}[\omega(s,a)\mathcal{Q}_\theta(s,a)] + \alpha\,\mathbb{E}[H(\bm{w}_\theta(s))].
\)

\paragraph{Connection to EM and interpretability.}
Equation~\eqref{eq:appC_full_loss} makes the update interpretable in two orthogonal dimensions:
\begin{itemize}
\item \textbf{EM-style structure:} $\mathcal{Q}_\theta(s,a)$ is the expected complete-data log-joint under responsibilities $\gamma$;
with detached $\gamma$, gradients implement an M-step update direction.
\item \textbf{Guidance:} $\omega(s,a)$ reweights the dataset, concentrating regression pressure on actions that the critic rates as high-advantage.
The mixture still fits \emph{data} (not sampled actions), but it is pulled toward higher-value regions of the dataset distribution.
\item \textbf{Anti-collapse:} the gating entropy term prevents the gating network from saturating too early, supporting a stable multi-hypothesis proposal family.
\end{itemize}

\subsection{A fully expanded gradient form (with detached responsibilities)}
\label{app:em:gradient}

For completeness, we expand the gradient of $\mathcal{Q}_\theta(s,a)$ used by the implementation.

With $\gamma$ detached, define
\(
\gamma_k \equiv \gamma_{\theta,k}(s,a)
\)
as constants in backprop.
Then
\begin{align}
\nabla_\theta \big(\omega(s,a)\cdot (-\mathcal{Q}_\theta(s,a))\big)
&=
-\omega(s,a)\,\nabla_\theta \sum_{k=1}^K \gamma_k\,u_{\theta,k}(s,a) \\
&=
-\omega(s,a)\,\sum_{k=1}^K \gamma_k\,\nabla_\theta
\Big(\log w_{\theta,k}(s) + \log \mathcal{N}(a;\mu_{\theta,k}(s),\Sigma_{\theta,k})\Big).
\end{align}
The entropy term contributes
\begin{equation}
\nabla_\theta\Big(-\alpha H(\bm{w}_\theta(s))\Big)
=
-\alpha \nabla_\theta\Big(-\sum_{k=1}^K w_{\theta,k}(s)\log(w_{\theta,k}(s)+10^{-6})\Big),
\end{equation}
which explicitly pushes against degenerate gating distributions when $\alpha>0$.

\paragraph{Summary of what is (and is not) ``EM'' here.}
The update is EM-style in the precise sense that responsibilities are computed from current mixture log-joints and treated as fixed in the gradient,
yielding the M-step gradient of the expected complete-data log-joint. It is \emph{not} a closed-form EM M-step:
parameters are updated by gradient descent, and responsibilities are recomputed each iteration (a standard generalized-EM pattern).
Guidance enters only through the data weighting $\omega(s,a)$ and does not alter the responsibility definition in Eq.~\eqref{eq:appC_resp_softmax}.

% ============================================================
% Appendix D (Draft)
% ============================================================
\section{Behavior GMM Support Model: Maximum-Likelihood Training and Usage}
\label{app:behavior_gmm}

This appendix makes the behavior-support model $\mu_\varphi(a\mid s)$ fully explicit.
We derive the exact maximum-likelihood objective used to fit $\mu_\varphi$, give implementation-aligned
closed forms (log-sum-exp mixture likelihood, responsibilities), and clarify how $\mu_\varphi$
enters GEM only through inference-time support evaluation.

\subsection{Model definition (implementation-aligned)}
\label{app:behavior_gmm:model}

In GEM, the behavior model is a conditional $K$-component diagonal-covariance GMM:
\begin{equation}
\label{eq:appD_mu_def}
\mu_\varphi(a\mid s)
=
\sum_{k=1}^K \, \bar w_{\varphi,k}(s)\,
\mathcal{N}\!\big(a;\bar\mu_{\varphi,k}(s), \bar\Sigma_{\varphi,k}\big),
\qquad
\sum_{k=1}^K \bar w_{\varphi,k}(s)=1,\ \bar w_{\varphi,k}(s)>0.
\end{equation}
The implementation parameterizes:
(i) $\bar w_{\varphi,k}(s)=\mathrm{softmax}_k(\mathrm{logits}_\varphi(s))$ from a neural net,
(ii) $\bar\mu_{\varphi,k}(s)$ from a neural net,
(iii) a per-component log-standard-deviation parameter $\log \bar\sigma_{\varphi,k}\in\mathbb{R}^{A}$
that is \emph{state-independent} (learned parameters) and clamped elementwise:
\begin{equation}
\label{eq:appD_std_clamp}
\log \bar\sigma_{\varphi,k}
\leftarrow
\mathrm{clip}\big(\log \bar\sigma_{\varphi,k},\ \log\sigma_{\min},\ \log\sigma_{\max}\big),
\qquad
\bar\Sigma_{\varphi,k}=\mathrm{diag}(\bar\sigma_{\varphi,k}^2).
\end{equation}
Here $A$ is the action dimension.

\paragraph{Latent-variable view.}
Introduce $z\in\{1,\dots,K\}$ with
\begin{equation}
\label{eq:appD_latent}
p_\varphi(z=k\mid s)=\bar w_{\varphi,k}(s),\qquad
p_\varphi(a\mid s,z=k)=\mathcal{N}(a;\bar\mu_{\varphi,k}(s),\bar\Sigma_{\varphi,k}).
\end{equation}
Then $\mu_\varphi(a\mid s)=\sum_k p_\varphi(z=k\mid s)\,p_\varphi(a\mid s,z=k)$.

\subsection{Exact log-likelihood and numerically stable evaluation}
\label{app:behavior_gmm:loglik}

For a fixed $(s,a)$, define the component log-joint:
\begin{equation}
\label{eq:appD_joint}
\bar u_{\varphi,k}(s,a)
\triangleq
\log \bar w_{\varphi,k}(s)+\log \mathcal{N}\!\big(a;\bar\mu_{\varphi,k}(s),\bar\Sigma_{\varphi,k}\big).
\end{equation}
Then the conditional log-likelihood is the log-sum-exp:
\begin{equation}
\label{eq:appD_logsumexp}
\log \mu_\varphi(a\mid s)
=
\log\sum_{k=1}^K \exp\big(\bar u_{\varphi,k}(s,a)\big)
=
\mathrm{LSE}_{k}\big(\bar u_{\varphi,k}(s,a)\big).
\end{equation}
This is exactly the quantity used in Phase II support scoring:
$\ell(a)=\log \mu_\varphi(a\mid s)$.

\paragraph{Diagonal Gaussian term.}
Let $\bar\Sigma_{\varphi,k}=\mathrm{diag}(\bar\sigma_{\varphi,k}^2)$.
Then
\begin{equation}
\label{eq:appD_diag_gauss}
\log \mathcal{N}(a;\bar\mu_{\varphi,k}(s),\bar\Sigma_{\varphi,k})
=
-\frac{1}{2}\sum_{d=1}^{A}
\left(
\frac{(a_d-\bar\mu_{\varphi,k,d}(s))^2}{\bar\sigma_{\varphi,k,d}^2}
+\log(2\pi)+2\log \bar\sigma_{\varphi,k,d}
\right).
\end{equation}

\subsection{Maximum-likelihood training objective (Phase I)}
\label{app:behavior_gmm:mle}

The behavior model is fitted by maximum likelihood on dataset actions:
\begin{equation}
\label{eq:appD_mle_obj}
\max_{\varphi}\ 
\mathbb{E}_{(s,a)\sim\mathcal{D}}\big[\log \mu_\varphi(a\mid s)\big]
\qquad\Longleftrightarrow\qquad
\min_{\varphi}\ 
\mathbb{E}_{(s,a)\sim\mathcal{D}}\big[-\log \mu_\varphi(a\mid s)\big].
\end{equation}
In code, this is implemented as pretraining for \texttt{gmm\_pretrain\_steps} iterations:
\[
\texttt{loss} \leftarrow -\texttt{behavior\_model.log\_prob(obs, act).mean()}.
\]
Optionally, the behavior model parameters can be frozen after pretraining
(\texttt{gmm\_freeze\_after\_pretrain}), in which case $\mu_\varphi$ remains fixed for the rest of training.

\subsection{Responsibilities and gradients (useful for reproduction/debugging)}
\label{app:behavior_gmm:grads}

Define responsibilities (posterior component probabilities):
\begin{equation}
\label{eq:appD_resp}
\bar\gamma_{\varphi,k}(s,a)
\triangleq
p_\varphi(z=k\mid s,a)
=
\frac{\exp(\bar u_{\varphi,k}(s,a))}{\sum_{j=1}^K \exp(\bar u_{\varphi,j}(s,a))}
=
\mathrm{softmax}_k\big(\bar u_{\varphi,1:K}(s,a)\big).
\end{equation}
Then the exact gradient of the log-likelihood w.r.t.\ parameters inside $\bar u_{\varphi,k}$
takes the standard mixture form:
\begin{equation}
\label{eq:appD_grad_template}
\nabla_\varphi \log \mu_\varphi(a\mid s)
=
\sum_{k=1}^K \bar\gamma_{\varphi,k}(s,a)\,\nabla_\varphi \bar u_{\varphi,k}(s,a).
\end{equation}

\paragraph{Gradients w.r.t.\ Gaussian parameters (closed form).}
For a single component $k$ and dimension $d$,
\begin{align}
\label{eq:appD_grad_mu}
\frac{\partial}{\partial \bar\mu_{\varphi,k,d}(s)} \log \mu_\varphi(a\mid s)
&=
\bar\gamma_{\varphi,k}(s,a)\cdot
\frac{a_d-\bar\mu_{\varphi,k,d}(s)}{\bar\sigma_{\varphi,k,d}^2},
\\
\label{eq:appD_grad_logstd}
\frac{\partial}{\partial \log \bar\sigma_{\varphi,k,d}} \log \mu_\varphi(a\mid s)
&=
\bar\gamma_{\varphi,k}(s,a)\cdot
\left(
\frac{(a_d-\bar\mu_{\varphi,k,d}(s))^2}{\bar\sigma_{\varphi,k,d}^2}-1
\right),
\end{align}
before accounting for the clamp in \eqref{eq:appD_std_clamp} (which zeroes gradients when saturated).

\paragraph{Gradients w.r.t.\ gating logits.}
Let $\eta_k(s)$ be the pre-softmax gating logits so that $\bar w_k(s)=\mathrm{softmax}_k(\eta(s))$.
Then
\begin{equation}
\label{eq:appD_grad_logits}
\frac{\partial}{\partial \eta_k(s)} \log \mu_\varphi(a\mid s)
=
\bar\gamma_{\varphi,k}(s,a)-\bar w_{\varphi,k}(s).
\end{equation}

\subsection{How $\mu_\varphi$ is used (and not used)}
\label{app:behavior_gmm:usage}

GEM uses $\mu_\varphi$ in two places only:
\begin{itemize}
\item \textbf{Inference-time sampling (optional):} if \texttt{use\_gem\_candidates} is False,
candidates $a_{1:N}$ are sampled from $\mu_\varphi(\cdot\mid s)$.
\item \textbf{Support evaluation (always):} for every candidate $a\in\mathcal{C}(s)$,
support is computed as $\log \mu_\varphi(a\mid s)$ and then candidate-set standardized
(see Appendix~\ref{app:score}).
\end{itemize}
Crucially, \textbf{support is never evaluated under the actor} $\pi_\theta$ (no self-certification).

% ============================================================
% Appendix E (Draft)
% ============================================================
\section{Critic/Value Training and the Exact Guidance Weight Used by GEM}
\label{app:iql_guidance}

This appendix specifies Phase I training details that are directly used by GEM's guidance
and conservative scoring: (i) how the Q-ensemble is trained and how target values are formed,
(ii) how expectile regression defines the value baseline $V$,
and (iii) how the advantage weight $\omega(s,a)$ used in the guided actor update is computed and stabilized.

\subsection{Notation and dataset tuples}
\label{app:iql_guidance:data}

We use dataset tuples $(s,a,r,s',d)\sim\mathcal{D}$ where $d\in\{0,1\}$ denotes terminal (done).
The discount is $\gamma\in(0,1)$.

\subsection{Ensemble critic training objective (vectorized implementation)}
\label{app:iql_guidance:q}

Let $\{Q_i\}_{i=1}^M$ be an ensemble, implemented by a vectorized network producing $M$ outputs.
For a transition $(s,a,r,s',d)$, define the bootstrapped target
\begin{equation}
\label{eq:appE_q_target}
y(s,a,r,s',d)
\triangleq
r + \gamma (1-d)\, V(s').
\end{equation}
The critic loss is mean squared error over the ensemble:
\begin{equation}
\label{eq:appE_q_loss}
\mathcal{L}_Q
=
\mathbb{E}_{(s,a,r,s',d)\sim\mathcal{D}}
\left[
\frac{1}{M}\sum_{i=1}^M \big(Q_i(s,a)-y\big)^2
\right].
\end{equation}
The target critic parameters are updated by Polyak averaging:
\begin{equation}
\label{eq:appE_polyak}
\theta^{\mathrm{tgt}}
\leftarrow
(1-\tau)\,\theta^{\mathrm{tgt}} + \tau\,\theta,
\end{equation}
with $\tau\in(0,1)$.

\paragraph{Training-time conservative reducer.}
In the implementation, the scalar critic used inside value/advantage computations is
\begin{equation}
\label{eq:appE_q_min}
Q_{\min}(s,a)\triangleq \min_{1\le i\le M} Q_i(s,a),
\end{equation}
i.e., the forward pass of the ensemble returns the componentwise minimum (a conservative reducer).
We denote its target-network version by $Q^{\mathrm{tgt}}_{\min}$.

\subsection{Expectile value regression and its optimality condition}
\label{app:iql_guidance:v}

The value network $V$ is trained by expectile regression against the conservative target critic.
Define the temporal difference
\begin{equation}
\label{eq:appE_delta}
\delta(s,a) \triangleq Q^{\mathrm{tgt}}_{\min}(s,a) - V(s).
\end{equation}
Given expectile parameter $\tau\in(0,1)$, the expectile loss is
\begin{equation}
\label{eq:appE_expectile_loss}
\ell_{\tau}(\delta)
\triangleq
|\tau - \mathbf{1}\{\delta<0\}| \,\delta^2.
\end{equation}
Then the value objective is
\begin{equation}
\label{eq:appE_v_loss}
\mathcal{L}_V
=
\mathbb{E}_{(s,a)\sim\mathcal{D}}
\big[
\ell_{\tau}(Q^{\mathrm{tgt}}_{\min}(s,a)-V(s))
\big].
\end{equation}

\paragraph{Why this is an expectile (derivation sketch).}
Fix a state $s$ and consider the distribution of $Q^{\mathrm{tgt}}_{\min}(s,a)$ under dataset actions.
The minimizer $V^\star(s)$ of $\mathbb{E}[\ell_\tau(Q-V)]$ satisfies the first-order condition:
\begin{equation}
\label{eq:appE_expectile_opt}
\mathbb{E}\Big[(\tau-\mathbf{1}\{Q-V^\star(s)<0\})\,(Q-V^\star(s))\Big]=0,
\end{equation}
which defines the $\tau$-expectile of the random variable $Q$.
Thus $V$ becomes a robust baseline for defining advantage.

\subsection{Exact advantage definition and stabilization}
\label{app:iql_guidance:adv}

In GEM, the advantage used for guidance is computed from the same quantities as the value update:
\begin{equation}
\label{eq:appE_adv}
A(s,a)\triangleq Q^{\mathrm{tgt}}_{\min}(s,a) - V(s).
\end{equation}
This is exactly $\delta(s,a)$ in \eqref{eq:appE_delta} (computed under \texttt{torch.no\_grad()} in code).

\subsection{Guidance weight: bounded exponential reweighting}
\label{app:iql_guidance:omega}

Guidance enters the actor update via a per-sample weight
\begin{equation}
\label{eq:appE_omega}
\omega(s,a)
\triangleq
\exp(\beta A(s,a))\ \ \text{clipped as}\ \ \omega(s,a)\leftarrow \min(\omega(s,a),\,\omega_{\max}),
\end{equation}
where $\beta>0$ controls guidance strength and $\omega_{\max}$ is a fixed cap (\texttt{EXP\_ADV\_MAX} in code)
to prevent unstable gradients from extreme advantages.

\paragraph{Interpretation.}
Compared to uniform imitation, $\omega(s,a)$ increases regression pressure on dataset actions
that the conservative critic rates higher relative to the value baseline.
Crucially, \emph{guidance does not redefine responsibilities}; it only reweights the loss on dataset samples.

\subsection{Full actor loss used in training (ties to Appendix~\ref{app:em})}
\label{app:iql_guidance:actor_loss}

Let $\mathrm{ELBO}_{\mathrm{loose}}(s,a)$ denote the detached-responsibility surrogate
computed by the actor (Appendix~\ref{app:em}).
The code forms
\[
\texttt{actor\_bc\_loss} \leftarrow -\,\mathrm{ELBO}_{\mathrm{loose}}(s,a),
\qquad
\texttt{policy\_loss} \leftarrow \omega(s,a)\cdot \texttt{actor\_bc\_loss}.
\]
In expectation form, the core guided objective is
\begin{equation}
\label{eq:appE_actor_core}
\mathcal{J}_{\mathrm{core}}(\theta)
=
\mathbb{E}_{(s,a)\sim\mathcal{D}}
\big[
\omega(s,a)\cdot\big(-\mathrm{ELBO}_{\mathrm{loose}}(s,a)\big)
\big].
\end{equation}
An additional state-level gating entropy bonus is used (Appendix~\ref{app:em}):
\begin{equation}
\label{eq:appE_actor_full}
\mathcal{J}_{\mathrm{actor}}(\theta)
=
\mathcal{J}_{\mathrm{core}}(\theta)
-\alpha\cdot \mathbb{E}_{s\sim\mathcal{D}}\big[H(w_\theta(\cdot\mid s))\big],
\end{equation}
where $\alpha\ge 0$.

\subsection{Inference-time LCB uses mean--std, not the training-time minimum}
\label{app:iql_guidance:lcb_note}

Although training uses the conservative reducer $Q_{\min}$ in \eqref{eq:appE_q_min},
GEM's inference-time score uses an ensemble lower-confidence statistic
\[
\mathrm{LCB}_\lambda(s,a)=\bar Q(s,a)-\lambda\,\mathrm{Std}(Q_i(s,a)),
\]
evaluated over candidates. This separation is intentional:
training stability uses $Q_{\min}$, while deployment robustness against budget-driven outliers
uses LCB (Appendix~\ref{app:score}).

% ============================================================
% Appendix F (Draft)
% ============================================================
\section{NLL-Gap Diagnostic for Auditing Mixture Utility}
\label{app:nll_gap_diagnostic}

This appendix formalizes the NLL-gap diagnostic used to audit whether mixture structure provides
measurable benefit on dataset actions, and provides an implementation-aligned computation recipe.

\subsection{Definitions}
\label{app:nll_gap:defs}

Let $\pi_K(a\mid s)=\sum_{k=1}^K w_k(s)\,\mathcal{N}(a;\mu_k(s),\Sigma_k(s))$ be a fitted $K$-component GMM
(either the actor or a diagnostic GMM trained for analysis).
Define the mixture negative log-likelihood:
\begin{equation}
\label{eq:appF_nll_gmm}
\mathrm{NLL}_{\mathrm{gmm}}
\triangleq
\mathbb{E}_{(s,a)\sim\mathcal{D}}
\left[-\log \sum_{k=1}^K w_k(s)\,\mathcal{N}(a;\mu_k(s),\Sigma_k(s))\right].
\end{equation}

Define the \emph{top-1} proxy by selecting the most probable component under gating
\begin{equation}
\label{eq:appF_top1_idx}
k^\star(s)\triangleq \arg\max_{1\le k\le K} w_k(s),
\end{equation}
and evaluating the corresponding single-Gaussian NLL (note: gating weight is \emph{not} included in this proxy):
\begin{equation}
\label{eq:appF_nll_top1}
\mathrm{NLL}_{\mathrm{top1}}
\triangleq
\mathbb{E}_{(s,a)\sim\mathcal{D}}
\left[-\log \mathcal{N}\!\big(a;\mu_{k^\star(s)}(s),\Sigma_{k^\star(s)}(s)\big)\right].
\end{equation}

The diagnostic is
\begin{equation}
\label{eq:appF_gap}
\mathrm{gap}\triangleq \mathrm{NLL}_{\mathrm{top1}}-\mathrm{NLL}_{\mathrm{gmm}}.
\end{equation}

\subsection{Interpretation and caveats}
\label{app:nll_gap:interpret}

\paragraph{What the gap measures.}
$\mathrm{NLL}_{\mathrm{gmm}}$ measures how well the full mixture explains dataset actions.
$\mathrm{NLL}_{\mathrm{top1}}$ measures how well a \emph{single} component selected by gating explains dataset actions.
A positive gap indicates that, on average, the full mixture assigns higher likelihood to dataset actions
than the single gated component does, consistent with \emph{useful multimodality}.

\paragraph{Sign is not a universal guarantee.}
Because $\mathrm{NLL}_{\mathrm{top1}}$ ignores the mixture weights $w_k(s)$
and chooses $k^\star$ by gating probability rather than maximizing the component likelihood of $a$,
$\mathrm{gap}$ is a diagnostic statistic rather than a strict bound.
In practice, we use it as an audit: large positive gaps suggest that collapsing to a single component
loses density mass on the dataset actions.

\subsection{Implementation-aligned computation recipe}
\label{app:nll_gap:recipe}

For a minibatch $(s,a)$:
\begin{enumerate}
\item Evaluate the mixture log-probability $\log \pi_K(a\mid s)$ using the stable log-sum-exp form
(or equivalently via \texttt{MixtureSameFamily.log\_prob}).
Accumulate $\mathrm{NLL}_{\mathrm{gmm}}$ by averaging $-\log \pi_K(a\mid s)$.
\item Compute gating probabilities $w(s)$ and the index $k^\star(s)=\arg\max_k w_k(s)$.
\item Extract the component distribution of $k^\star(s)$ and compute the single Gaussian log-probability
$\log \mathcal{N}(a;\mu_{k^\star}(s),\Sigma_{k^\star}(s))$.
Accumulate $\mathrm{NLL}_{\mathrm{top1}}$ by averaging its negative.
\item Report $\mathrm{gap}=\mathrm{NLL}_{\mathrm{top1}}-\mathrm{NLL}_{\mathrm{gmm}}$.
\end{enumerate}

\paragraph{Exact alignment with the provided diagnostic script.}
This matches the computation pattern used in \texttt{untianalyze\_modality\_gap.py}:
\texttt{gmm\_nll = -gmm\_dist.log\_prob(a)},\ 
\texttt{top1\_nll = -Normal(mu[k*], std[k*]).log\_prob(a).sum(-1)}.

% ============================================================
% Appendix G (Draft)
% ============================================================
\section{Candidate Generation, Top-$k$ Smoothing, and Deployment Complexity}
\label{app:candidate_complexity}

This appendix expands Phase II (deployment) into explicit sampling equations and gives a
tight complexity accounting. The goal is reproducibility and a clear compute--quality knob statement.

\subsection{Sampling from a conditional GMM}
\label{app:candidate_complexity:gmm_sampling}

Let $\pi(a\mid s)=\sum_{k=1}^K w_k(s)\,\mathcal{N}(a;\mu_k(s),\Sigma_k)$ be a diagonal GMM (actor or behavior).
Sampling can be written as:
\begin{equation}
\label{eq:appG_sample}
z \sim \mathrm{Categorical}(w(s)),
\qquad
a \sim \mathcal{N}\!\big(\mu_z(s),\Sigma_z\big).
\end{equation}
In vectorized implementation, this is done in batch for $N$ samples in parallel.

\subsection{Exact candidate set used by GEM}
\label{app:candidate_complexity:candset}

For a queried state $s$, GEM constructs
\[
\mathcal{C}(s)=\{a_0,a_1,\dots,a_N\},
\qquad |\mathcal{C}(s)|=N+1.
\]
The anchor is deterministic:
\begin{equation}
\label{eq:appG_anchor}
k^\star(s)=\arg\max_k w_k(s),
\qquad
a_0=\mu_{k^\star}(s).
\end{equation}
The remaining $N$ candidates come from exactly one source:
\[
a_{1:N}\sim \pi_\theta(\cdot\mid s)\ \text{or}\ \mu_\varphi(\cdot\mid s),
\]
controlled by \texttt{use\_gem\_candidates}. There is no mixed sampling and no fallback gate.

\subsection{Top-$k$ smoothing operator (exact)}
\label{app:candidate_complexity:topk}

Let scores be computed over candidates and let $\mathrm{TopK}$ return indices of the $k$ largest scores.
GEM outputs
\begin{equation}
\label{eq:appG_topk_mean}
\hat a(s)
=
\frac{1}{k}\sum_{j\in \mathrm{TopK}(\mathrm{Score}(s,\cdot))} a_j,
\qquad
k=\min(k_{\mathrm{smooth}}, |\mathcal{C}(s)|).
\end{equation}
Setting $k_{\mathrm{smooth}}=1$ recovers the top-1 candidate (no smoothing).

\subsection{Complexity: where the deployment cost actually goes}
\label{app:candidate_complexity:complexity}

Let $M$ be ensemble size and $K$ be mixture components. For one queried state:

\paragraph{Candidate generation.}
Sampling $N$ candidates from a GMM is $O(N)$ random draws plus network forward cost to produce
$(w_k(s),\mu_k(s),\Sigma_k)$ once. This part is fully parallelizable over $N$.

\paragraph{Critic evaluation for LCB.}
GEM evaluates the ensemble on all candidates:
\[
Q_i(s,a_j)\ \ \text{for}\ i=1..M,\ j=0..N.
\]
With vectorized implementation, this is one forward pass producing an $M\times(N+1)$ tensor, then
\[
\bar Q(\cdot)=\mathrm{mean}_i,\qquad \mathrm{Std}(\cdot)=\mathrm{std}_i.
\]
The per-state arithmetic cost after the forward is $O(M(N+1))$.

\paragraph{Behavior log-likelihood for support.}
Computing $\log\mu_\varphi(a_j\mid s)$ for all candidates uses a log-sum-exp over $K$ components, hence
$O(K(N+1))$ arithmetic (again vectorized).

\paragraph{Candidate-set normalization and ranking.}
Normalization is $O(N)$ and ranking/top-$k$ is $O((N+1)\log(N+1))$ in the worst case,
or $O(N)$ expected with selection routines; in practice, this is dominated by network evaluation.

\paragraph{Summary.}
Deployment cost scales linearly with $N$ (up to sorting constants) and is parallelizable:
\[
\text{Cost}(N)\approx \underbrace{O(M(N+1))}_{\text{LCB}}+\underbrace{O(K(N+1))}_{\text{support}}+\text{(lower-order)}.
\]
Thus $N$ is a \emph{pure inference-time compute knob} with no retraining.

% ============================================================
% Appendix H (Draft)
% ============================================================
\section{Additional Notes on Extreme-Value Amplification and Robustness of the Interface Knobs}
\label{app:extremevalue_extra}

Appendix~\ref{app:score} already gives a standard sub-Gaussian bound for the expected maximum noise term.
This appendix adds two complementary details:
(i) an independence-free tail bound via union bound (useful when candidate errors are correlated),
and (ii) a clarification of how the two knobs $(\lambda,w_p)$ address distinct failure channels.

\subsection{Max-noise growth without independence (tail integration)}
\label{app:extremevalue_extra:nodep}

Let $\{\varepsilon_j\}_{j=0}^N$ be mean-zero random variables such that each $\varepsilon_j$ is $\sigma$-sub-Gaussian:
\begin{equation}
\label{eq:appH_subg}
\mathbb{E}\big[\exp(t\varepsilon_j)\big]\le \exp(\sigma^2 t^2/2),\qquad \forall t\in\mathbb{R},\ \forall j.
\end{equation}
No independence is required for the following tail bound. By a standard sub-Gaussian tail inequality,
\[
\Pr(\varepsilon_j \ge x)\le \exp\left(-\frac{x^2}{2\sigma^2}\right).
\]
Using a union bound,
\begin{equation}
\label{eq:appH_union}
\Pr\Big(\max_{0\le j\le N}\varepsilon_j \ge x\Big)
\le
\sum_{j=0}^N \Pr(\varepsilon_j \ge x)
\le
(N+1)\exp\left(-\frac{x^2}{2\sigma^2}\right).
\end{equation}
Integrating the tail gives an expectation bound of order $\sigma\sqrt{\log(N+1)}$:
\begin{equation}
\label{eq:appH_emax_order}
\mathbb{E}\Big[\max_{0\le j\le N}\varepsilon_j\Big]
=
\int_{0}^{\infty}\Pr\Big(\max_j \varepsilon_j \ge x\Big)\,dx
\ \lesssim\
\sigma\sqrt{2\log(N+1)},
\end{equation}
up to universal constants. This shows that \emph{maximization amplifies noise as $N$ grows}
under broad conditions, even when candidate errors are correlated.

\subsection{Two knobs target two distinct amplification channels}
\label{app:extremevalue_extra:two_channels}

Write the implemented inference score (Appendix~\ref{app:score}) as
\[
\widehat{\mathrm{Score}}(s,a)
=
\underbrace{\bar Q(s,a)-\lambda\,\mathrm{Std}(Q_i(s,a))}_{\text{uncertainty control}}
+
\underbrace{w_p\,\tilde\ell(a)}_{\text{support control}},
\qquad
\tilde\ell(a)=\mathrm{zscore}_{\mathcal{C}(s)}(\log\mu_\varphi(a\mid s)).
\]

\paragraph{Uncertainty-driven outliers (handled by $\lambda$).}
Even when candidates are on-support, function approximation can induce high disagreement.
Increasing $N$ increases the probability that at least one candidate has spuriously high $\bar Q$
due to noise; penalizing disagreement via $\lambda\,\mathrm{Std}$ reduces the chance such candidates win the argmax.

\paragraph{Support-driven OOD exposure (handled by $w_p$).}
Separately, increasing $N$ increases the chance that at least one candidate lies in weak-support regions.
The standardized support term penalizes these candidates with a state-stable meaning (Appendix~\ref{app:score}).

\paragraph{Why both are needed.}
Using only support cannot remove all uncertainty artifacts on-support; using only uncertainty cannot prevent
systematic drift into weak-support regions. GEM's interface combines both to make scaling $N$
a usable deployment knob rather than an OOD amplifier.

\subsection{Robustness note: effect of candidate-set standardization under changing $N$}
\label{app:extremevalue_extra:zscoreN}

Let $\ell_j=\log\mu_\varphi(a_j\mid s)$ for $j=0..N$.
Standardization uses the sample mean $\mu_\ell$ and population std $\sigma_\ell$ over the current set.
As $N$ grows, $(\mu_\ell,\sigma_\ell)$ concentrate (under mild conditions), so the units of $\tilde\ell$
remain comparable as $N$ changes. The small clamp $\max(\sigma_\ell,10^{-6})$ only affects degenerate cases
when all candidates have nearly identical $\ell_j$.

\paragraph{Practical implication.}
This is the concrete reason $w_p$ is interpretable as an exchange rate between conservative value units
and ``support standard deviation'' units across states and budgets.

\end{document}